# ChemSpaceAL: An Efficient Active Learning Methodology Applied to Protein-Specific Molecular Generation

**Gregory W. Kyro[†], Anton Morgunov[†], Rafael I. Brent[†], Victor S. Batista[*]**

Yale University
{gregory.kyro, anton.morgunov, rafi.brent, victor.batista}@yale.edu

## Abstract

The incredible capabilities of generative artificial intelligence models have inevitably led to their application in the domain of drug discovery. Within this domain, the vastness of chemical space motivates the development of more efficient methods for identifying regions with molecules that exhibit desired characteristics. In this work, we present a computationally efficient active learning methodology that requires evaluation of only a subset of the generated data in the constructed sample space to successfully align a generative model with respect to a specified objective. We demonstrate the applicability of this methodology to targeted molecular generation by fine-tuning a GPT-based molecular generator toward a protein with FDA-approved small-molecule inhibitors, c-Abl kinase. Remarkably, the model learns to generate molecules similar to the inhibitors without prior knowledge of their existence, and even reproduces two of them exactly. We also show that the methodology is effective for a protein without any commercially available small-molecule inhibitors, the HNH domain of the CRISPR-associated protein 9 (Cas9) enzyme. We believe that the inherent generality of this method ensures that it will remain applicable as the exciting field of in silico molecular generation evolves. To facilitate implementation and reproducibility, we have made all of our software available through the open-source ChemSpaceAL Python package.

# 1. Introduction

The vast majority of pharmaceutical drugs function by targeting a specific protein.[1] Virtual screening and de novo drug design are popular methods for developing effective drugs.[2] Molecular generation methods powered by generative artificial intelligence (AI) can benefit both of these strategies, and there have been numerous reports of recurrent neural networks (RNNs),[3-25] generative adversarial networks (GANs),[26-39] autoencoders,[40-63] and transformers[64-71] demonstrating remarkable capabilities.

Active Learning (AL) methods can be used to fine-tune an AI model with selectively chosen data points, ensuring that the model retains its broad domain knowledge while narrowing its focus toward a more precise objective. In its basic form, AL can be applied by exclusively using data points that have been directly evaluated and satisfy specific criteria. However, within the AL framework, it is feasible to extend traditional methods by not only including directly evaluated data points, but also incorporating a mechanism that utilizes unevaluated data points similar to the evaluated ones deemed satisfactory. This approach facilitates the use of resource-intensive scoring functions that otherwise would be too expensive by scoring only a strategically selected subset of data points and extending the insights obtained from the scores to data that have not been evaluated. In this context, the total computational cost is largely dependent on the number of sampled molecules necessary to sufficiently represent the ideal search space.

Although there are many notable examples of AL methods for supervised learning tasks related to virtual screening for drug discovery,[72-77] the application of AL in generative AI for molecular generation is comparatively unexplored. In this emerging field, Filella-Merce et al. recently presented a two-tiered AL strategy using a variational autoencoder, where an inner loop filters molecules based on molecular properties, and an outer loop docks the molecules that pass the inner loop's filters to a protein target to identify molecules with satisfactory in silico binding affinities.[52] While this method is successful in generating chemically viable molecules with improved predicted affinity toward the target, it relies on docking each molecule in the outer loop, which is computationally expensive. It is therefore of interest to develop a computationally efficient approach for fine-tuning a molecular generator toward a protein target that does not require docking each molecule, and instead exploits a strategic method for estimating binding ability of molecules that have not been directly evaluated. This would significantly enhance the computational efficiency of AL for aligning molecular generators toward specified targets.

In this work, we present an efficient AL methodology that employs a strategic sampling algorithm and requires evaluation of only a subset of the generated data to successfully align the generated molecular ensemble toward a specified protein target. Specifically, we demonstrate the effectiveness of our methodology by independently aligning a Generative Pretrained Transformer (GPT)-based model to c-Abl kinase and the HNH domain of the CRISPR-associated protein 9 (Cas9) enzyme.[78,79]

## 2. Overview of the ChemSpaceAL Methodology

Our demonstration of the ChemSpaceAL methodology applied to molecular generation (Figure 1) proceeds as follows:

1) Pretrain the GPT-based model on millions of SMILES (Simplified Molecular Input Line Entry System) strings
2) Use the trained model to generate 100,000 unique molecules (determined by SMILES-string canonicalization)
3) Calculate molecular descriptors for each generated molecule
4) Project the descriptor vectors of the generated molecules into a Principal Component Analysis (PCA)-reduced space constructed from the descriptors of all molecules in the pretraining set
5) Use k-means clustering on the generated molecules within the space to group those with similar properties
6) Sample about 1% of molecules from each cluster and dock each of them to a protein target (e.g., c-Abl kinase or the HNH domain of Cas9)
7) Evaluate the top-ranked pose of each protein-ligand complex with an attractive interaction-based scoring function
8) Construct an AL training set by sampling from the clusters proportionally to the mean scores of the evaluated molecules within each respective cluster, and combining the sampled molecules with replicas of the evaluated molecules whose scores meet a specified threshold
9) Fine-tune the model with the AL training set

*) Repeat steps (2) – (9) for multiple iterations

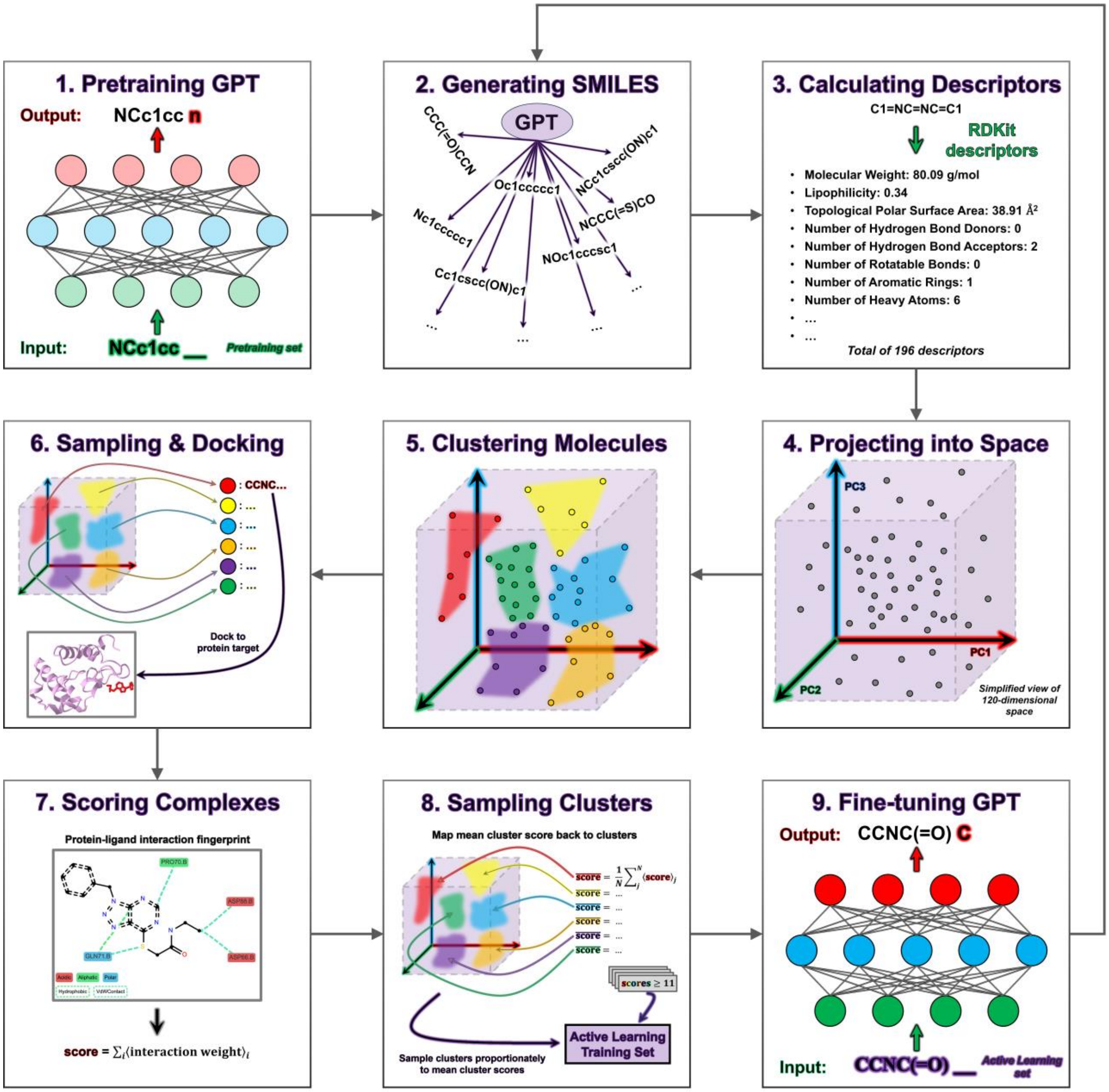

**Figure 1.** Process flow diagram depicting the complete ChemSpaceAL active learning methodology applied to molecular generation.

# 3. Aligning the Generative Model to Specified Protein Targets

Utilizing a transformer decoder-based GPT model (more details in Section 7),[80] our initial goal is to pretrain the model on data that span as much of true chemical space as possible. This approach allows the pretrained model to develop a rich internal representation of SMILES strings, enabling it to generate a diverse array of molecules. To curate an extensive dataset for pretraining the model, we combine SMILES strings from four datasets: ChEMBL 33 (about 2.4 million bioactive molecules with drug-like properties),[81] GuacaMol v1 (about 1.6 million molecules derived from ChEMBL 24 that have been synthesized and tested against biological targets),[82] MOSES (about 1.8 million molecules selected from Zinc 15 to maximize internal diversity and suitability for medicinal chemistry),[83,84] and BindingDB 08-2023 (about 1.2 million unique small molecules bound to proteins).[85] After processing, the resulting dataset contains about 5.6 million unique and valid SMILES strings, and will be referred to as the *combined dataset*. More details regarding the data that we use and the preprocessing methods that we employ are discussed in Section 6. To assess the dependence of our methodology on the nature of the pretraining set, we compare two independent models: one pretrained on the combined dataset (C model), and one pretrained on the MOSES dataset (M model).

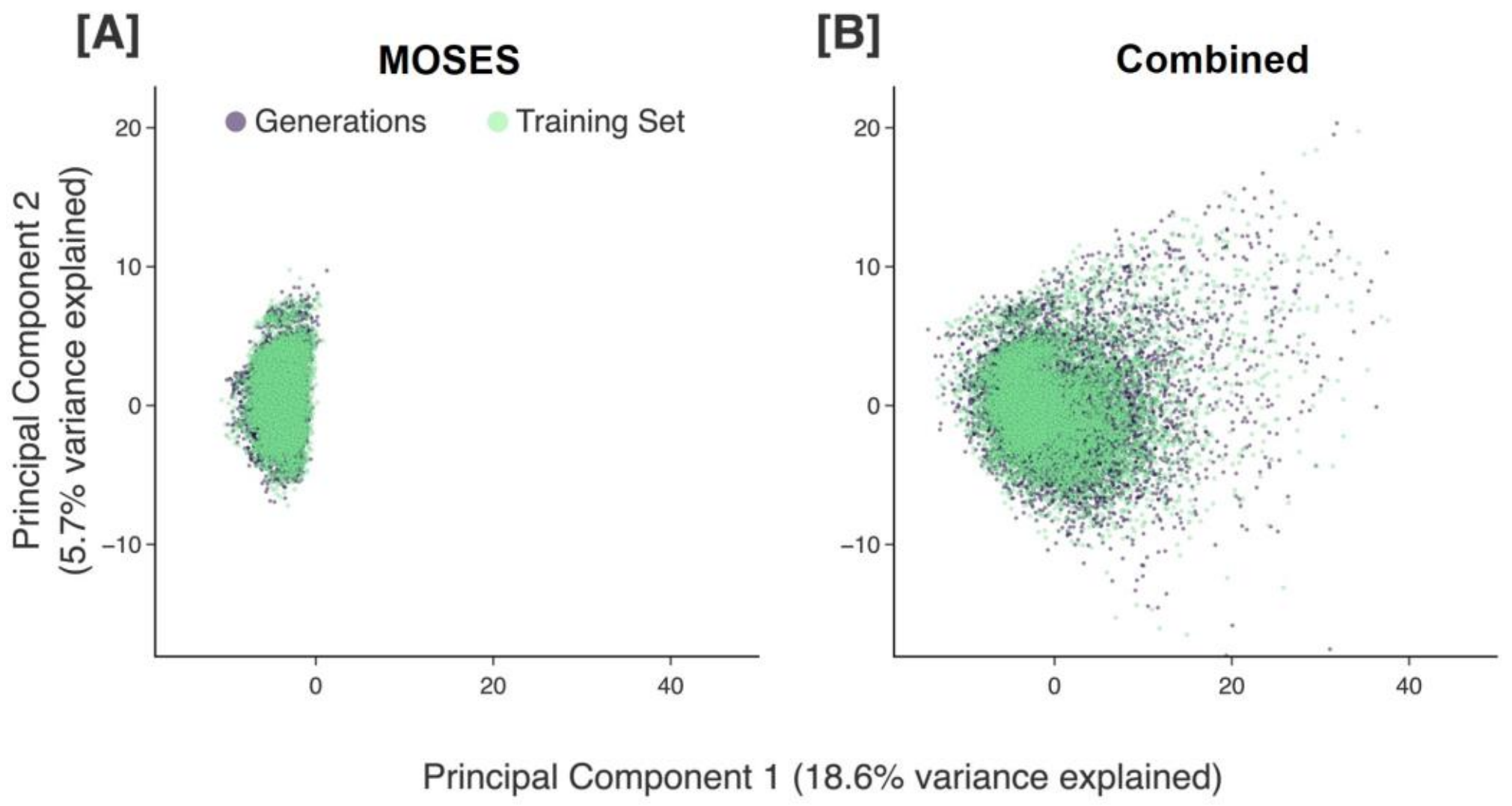

**Figure 2.** Different pretraining sets (green) plotted with the molecules generated (purple) by the corresponding pretrained model that is trained only on the respective pretraining set. 100,000 data points are randomly sampled from each pretraining set, and 100,000 are generated in each case. The descriptor vectors of the data points are projected into our chemical space proxy and the first two principal components are shown. Results are displayed for the MOSES **(A)** and combined **(B)** pretraining sets.

In Figure 2, we show 100,000 generated molecules from each model trained solely on either the MOSES or combined dataset along the first two principal components of our chemical space proxy. It should be noted that the PCA reduction is performed only once on the molecular descriptors of all molecules in the combined dataset and the obtained principal components are used for all visualizations throughout this work, ensuring fair comparison between different sets of data points (more details in Section 8.1). We see that the pretrained models are able to generate molecules that roughly cover the area spanned by the corresponding pretraining sets (Figure 2).

Using both pretrained models, we independently assess the ChemSpaceAL methodology with c-Abl kinase and the HNH domain of Cas9. In the first case, we aim to validate our methodology by showing that the generated molecular ensemble evolves toward the FDA-approved small-molecule inhibitors of c-Abl kinase. In the latter case, we investigate the applicability of the methodology to a protein without any commercially available small-molecule inhibitors.

In both cases, the generated molecules are filtered based on ADMET (Absorption, Distribution, Metabolism, Excretion, and Toxicity) metrics and functional group restrictions.[86] ADMET filters are employed to ensure that the molecules possess drug-like properties, and functional group restrictions are used to discard chemical moieties that are less favorable for biological applications. More details regarding the ADMET and functional group filters that we use are reported in Tables S1.1 and S1.2 in the *Supporting Information*.

## 3.1. Aligning to C-Abl Kinase

C-Abl kinase (PDB ID: 1IEP)[72] is of significant scientific interest because its dysfunction is associated with the development of chronic myeloid leukemia, making it a vital target for anticancer drugs designed to inhibit its activity and thereby control the proliferation of cancer cells. There are multiple FDA-approved small-molecule inhibitors of c-Abl kinase that have similar structures, including imatinib, nilotinib, dasatinib, bosutinib, ponatinib, bafetinib, and asciminib.[78,87,88] We dock and score each of the inhibitors using our scoring function, and choose the lowest score among them (37) to be the score threshold for our methodology (more details in Section 8.3).

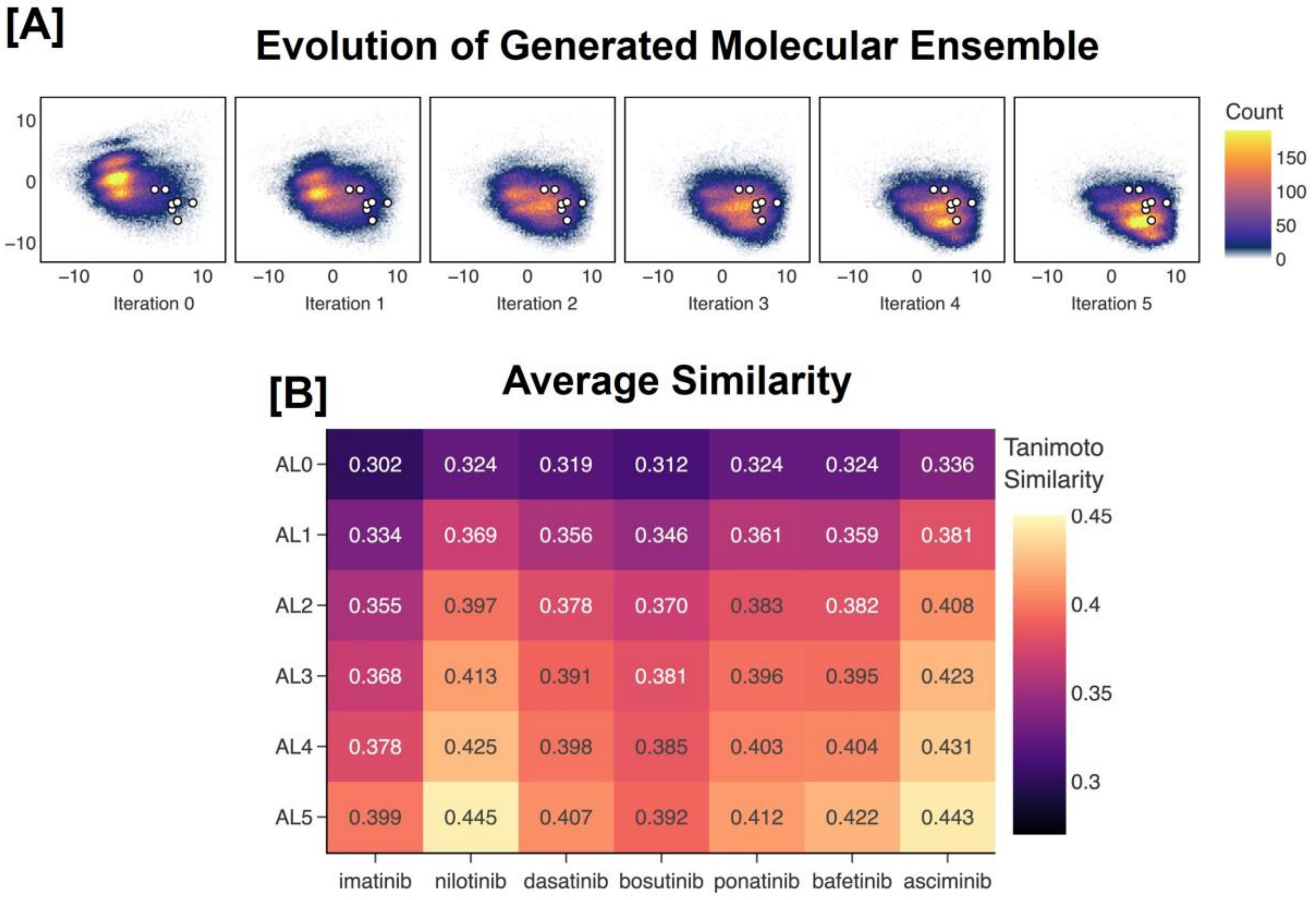

**Figure 3.** Comparing the evolution of the generated molecular ensemble from the model pretrained on the combined dataset to the FDA-approved small-molecule inhibitors of c-Abl kinase. In **(A)**, the descriptor vectors of the generated molecules across each iteration of our methodology are projected into our chemical space proxy and visualized along the first two principal components. The inhibitor descriptor vectors are also projected into the space and are represented by white dots with black outline. In **(B)**, the average Tanimoto similarities between the RDKit fingerprints of all generated molecules at each iteration and that of each inhibitor are shown. Tanimoto similarities between the inhibitors are reported in Figure S2.1 of the *Supporting Information*. Iteration 0 refers to the pretraining phase, while later iterations refer to the active learning phases.

For the C model, the mean Tanimoto similarities between the generated molecular ensemble and each of the seven inhibitors increase at each iteration, indicating a constant evolution toward the inhibitors (Figure 3B). This shift of the distribution toward the region of space that contains the FDA-approved inhibitors can be visualized by projecting the descriptor vectors of the generated ensemble at each iteration of the methodology and those of the inhibitors into the chemical space proxy (Figure 3A). Moreover, the set of generated molecules after five iterations contains imatinib and bosutinib (Figure 4).

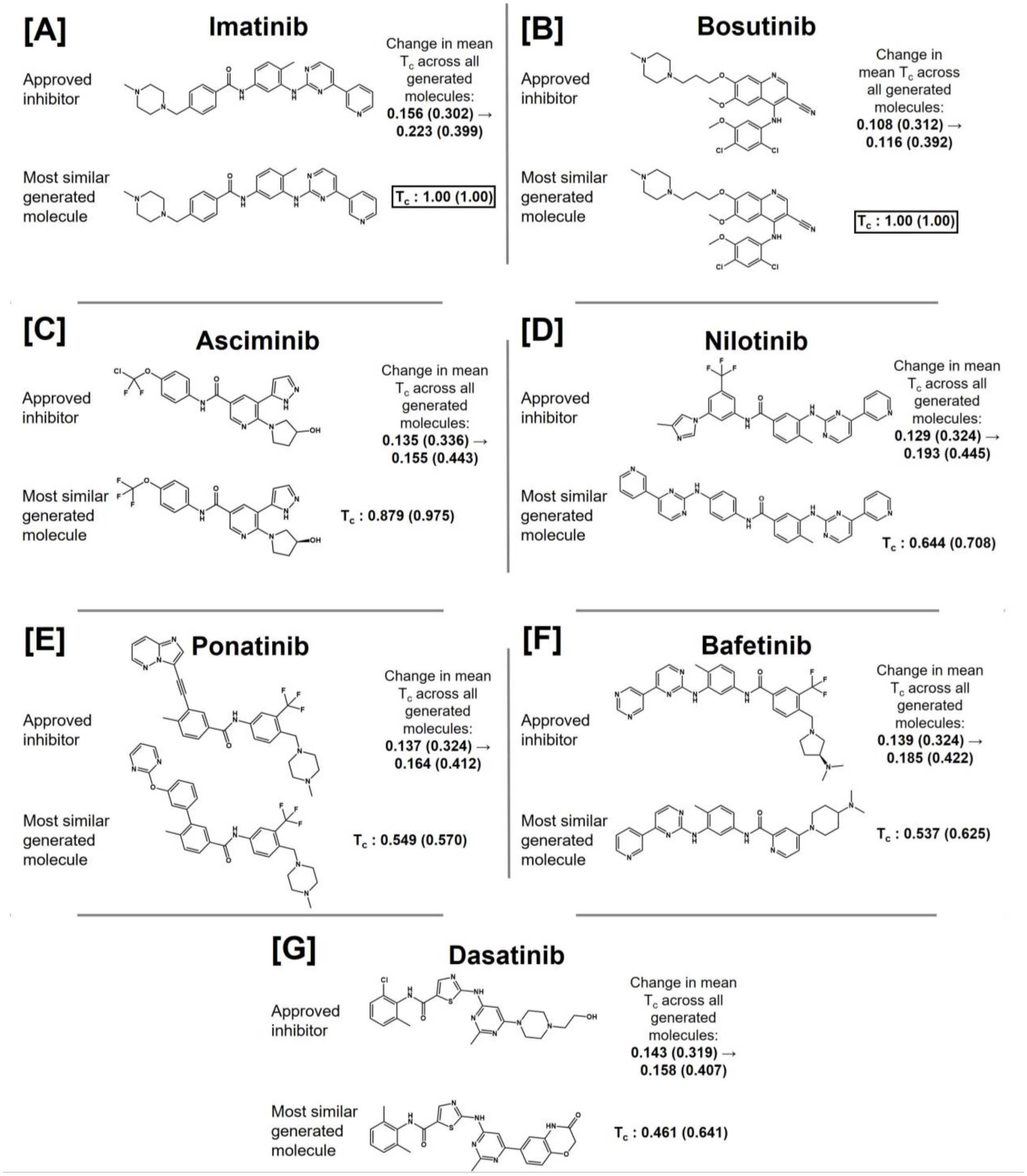

**Figure 4.** Comparison of the generated molecular ensemble from the model pretrained on the combined dataset to the FDA-approved small-molecule inhibitors of c-Abl kinase. For each inhibitor, the most similar generated molecule, after five iterations, is shown, as well as the Tanimoto similarity ($T_C$) between the two. The change in the mean similarity between each inhibitor and all generated molecules from iteration 0 (pretrained model) to iteration 5 is shown. For all comparisons in this figure, the $T_C$ between Extended-Connectivity Fingerprint 4s is shown along with the $T_C$ between RDKit fingerprints in parentheses.[89,90] Results are shown for imatinib **(A)**, bosutinib **(B)**, asciminib **(C)**, nilotinib **(D)**, ponatinib **(E)**, bafetinib **(F)**, and dasatinib **(G)**.

We also assess the performance of the methodology by analyzing the distribution of scores of generated molecules throughout AL iterations. For both the C and M models, the percentage of molecules that reaches the scoring threshold is significantly increased after five iterations of AL, further validating the applicability of our method to c-Abl kinase; the percentage is increased from 38.8% to 91.6% for the C model, and from 21.7% to 80.3% for the M model (Table 1). The evolutions of these distributions can be seen in Figure 5.

**Table 1.** Evolution of protein-ligand attractive interaction scores between molecules in the generated ensemble and c-Abl kinase across our complete active learning methodology.

| Iter | C %>37 | C Mean | C Max | M %>37 | M Mean | M Max |
|---|---|---|---|---|---|---|
| 0 | 38.8 | 32.8 | 70.0 | 21.7 | 30.3 | 55.5 |
| 1 | 59.3 | 38.4 | 74.5 | 42.1 | 35.2 | 57.0 |
| 2 | 70.1 | 41.4 | 68.0 | 59.2 | 38.0 | 60.5 |
| 3 | 81.2 | 44.0 | 73.5 | 68.8 | 39.9 | 60.0 |
| 4 | 86.6 | 46.0 | 77.5 | 76.2 | 41.0 | 59.0 |
| 5 | 91.6 | 48.5 | 77.0 | 80.3 | 41.8 | 61.0 |

[a] The percentage of generated molecules with attractive interaction scores equal to or above our score threshold (% > 37), the mean score, and the maximum score are shown for the model pretrained on the combined dataset (C), and the model pretrained on the MOSES dataset (M) for five iterations of the methodology.
[b] Iteration 0 refers to the pretraining phase, while later iterations refer to the active learning phases.

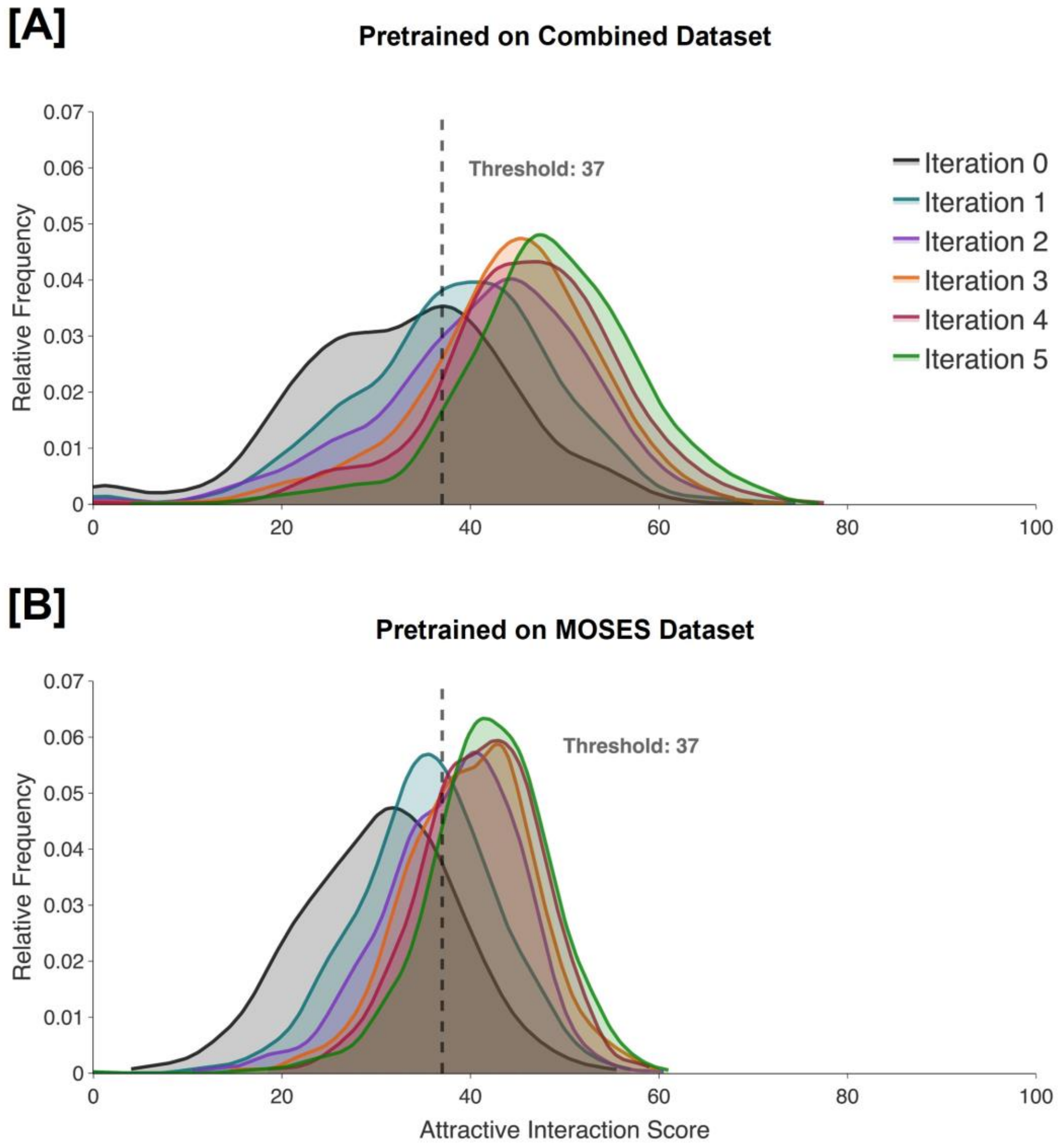

**Figure 5**. Attractive interaction scores of evaluated molecules across five iterations of active learning for c-Abl kinase. The distributions for the model pretrained on the combined dataset are shown in **(A)**, and the distributions for the model pretrained on the MOSES dataset are shown in **(B)**. Iteration 0 refers to the pretraining phase, while later iterations refer to the active learning phases.

It is worth nothing that 38.8% of the molecules generated by the C model reach the score threshold immediately after pretraining, while only 21.7% of the molecules generated by the M model reach the threshold, indicating that our combined pretraining set covers regions of chemical space not spanned by the MOSES dataset that contain higher-scoring molecules (Table 1). Moreover, after applying the methodology, the molecular ensemble generated by the C model is more similar to the FDA-approved inhibitors than that generated by the M model (Figure S3.1 in the *Supporting Information*), and is comprised exclusively of molecules with satisfactory ADMET profiles (Figure S4.1 in the *Supporting Information*). These results support the notion that our methodology is more effective at generating drug-like molecules specific to a protein target by pretraining on the combined dataset and applying filters to the generation stage, rather than pretraining on a refined dataset such as the MOSES dataset.

## 3.2. Aligning to the HNH Domain of Cas9

To further evaluate our methodology, we apply it to a protein without any commercially available small-molecule inhibitors, the HNH domain of Cas9 (PDB ID: 6O56).[79] This protein is a nuclease component critical to the function of the CRISPR/Cas9 system, and is responsible for cleaving the target DNA strand complementary to the guide RNA, which directs the Cas9 enzyme to the correct sequence for gene modification. The HNH domain of Cas9 is therefore particularly interesting because understanding its structure and dynamics can lead to enhancements in the precision and efficiency of CRISPR-based gene editing tools.[91] Furthermore, the ability to develop binders for HNH could offer a direct way to modulate its behavior.

Our methodology requires a score threshold in order to select molecules to be included in the AL training set. In the absence of known small-molecule binders for HNH, we refer to a large database of experimentally determined protein-ligand complexes, the PDBbind v2020 refined set,[92] and select this threshold to be 11 (more details in Section 8.3). This lack of known binders also leads us to use the change in the distribution of scores as the primary metric for evaluation. After five iterations of AL, the percentage of generated molecules that reaches the score threshold increases from 21.3% to 52.1% for the C model, and from 14.3% to 28.2% for the M model (Table 2); the performance differential between the C and M models is commensurate with that observed for c-Abl kinase. The evolutions of these distributions can be seen in Figure S5.1 in the *Supporting Information*.

**Table 2.** Evolution of protein-ligand attractive interaction scores between molecules in the generated ensemble and HNH across our complete active learning methodology.

| Iter | C %>11 | C Mean | C Max | M %>11 | M Mean | M Max |
|---|---|---|---|---|---|---|
| 0 | 21.3 | 7.9 | 32.5 | 14.3 | 7.3 | 22.5 |
| 1 | 31.9 | 9.1 | 26.5 | 18.9 | 7.8 | 21.0 |
| 2 | 39.1 | 9.8 | 25.0 | 22.5 | 8.2 | 22.0 |
| 3 | 43.9 | 10.4 | 23.0 | 24.5 | 8.6 | 23.0 |
| 4 | 50.1 | 11.1 | 33.5 | 28.7 | 8.9 | 21.0 |
| 5 | 52.1 | 11.5 | 34.0 | 28.2 | 9.0 | 23.0 |

[a] The percentage of generated molecules with attractive interaction scores equal to or above our score threshold (% > 11), the mean score, and the maximum score are shown for the model pretrained on the combined dataset (C), and the model pretrained on the MOSES dataset (M) for five iterations of the methodology.
[b] Iteration 0 refers to the pretraining phase, while later iterations refer to the active learning phases.

# 4. Evaluating Individual Components of the Methodology

The goal of this section is to isolate and analyze the effectiveness of individual components of our methodology: the chemical space proxy, clustering algorithm, scoring method, and sampling algorithm for constructing AL training sets. For all results presented here, the methodology is applied to the model pretrained on our combined dataset, aligned to HNH, and without any filters during generation stages. It is imperative to note that the function of the proposed methodology is simply to align the model toward achieving high scores as determined by a scoring function, through a constructed space that correlates with this function. In this section, we therefore seek to evaluate how the model responds with respect to the scoring function, and do not consider any filters on the generated molecules. However, to ensure a rigorous assessment, we additionally perform analogous analyses of the methodology applied to c-Abl kinase with ADMET and functional group filters applied to the generated molecules, and observe similar results as those included in this section (Figures S3.2 and S3.4 in the *Supporting Information*).

## 4.1. Naïve Active Learning Control

In order to establish a baseline for comparison to our methodology, we perform a naïve version of AL where we generate 100,000 unique molecules, randomly select 1,000 of them, dock and score each of the selected molecules, and then fine-tune the model with the scored molecules that reach the score threshold. The purpose of this approach is to demonstrate how the fine-tuning would occur if we did not sample from clusters in the chemical space proxy to construct an AL training set. In this case, we construct the AL training set from $N$ replicas of each molecule that scores equal to or above the score threshold (11), where $N$ is the smallest integer that achieves a total number of molecules of at least 5,000. The model is then further trained on this AL set, and the fine-tuned model is used to generate another 100,000 unique molecules which are subsequently used for another iteration of the methodology. We repeat this procedure for a total of five iterations, and observe that the percentage of generated molecules that reaches the score threshold increases from 26.2% to 44.2% (Figure 6A).

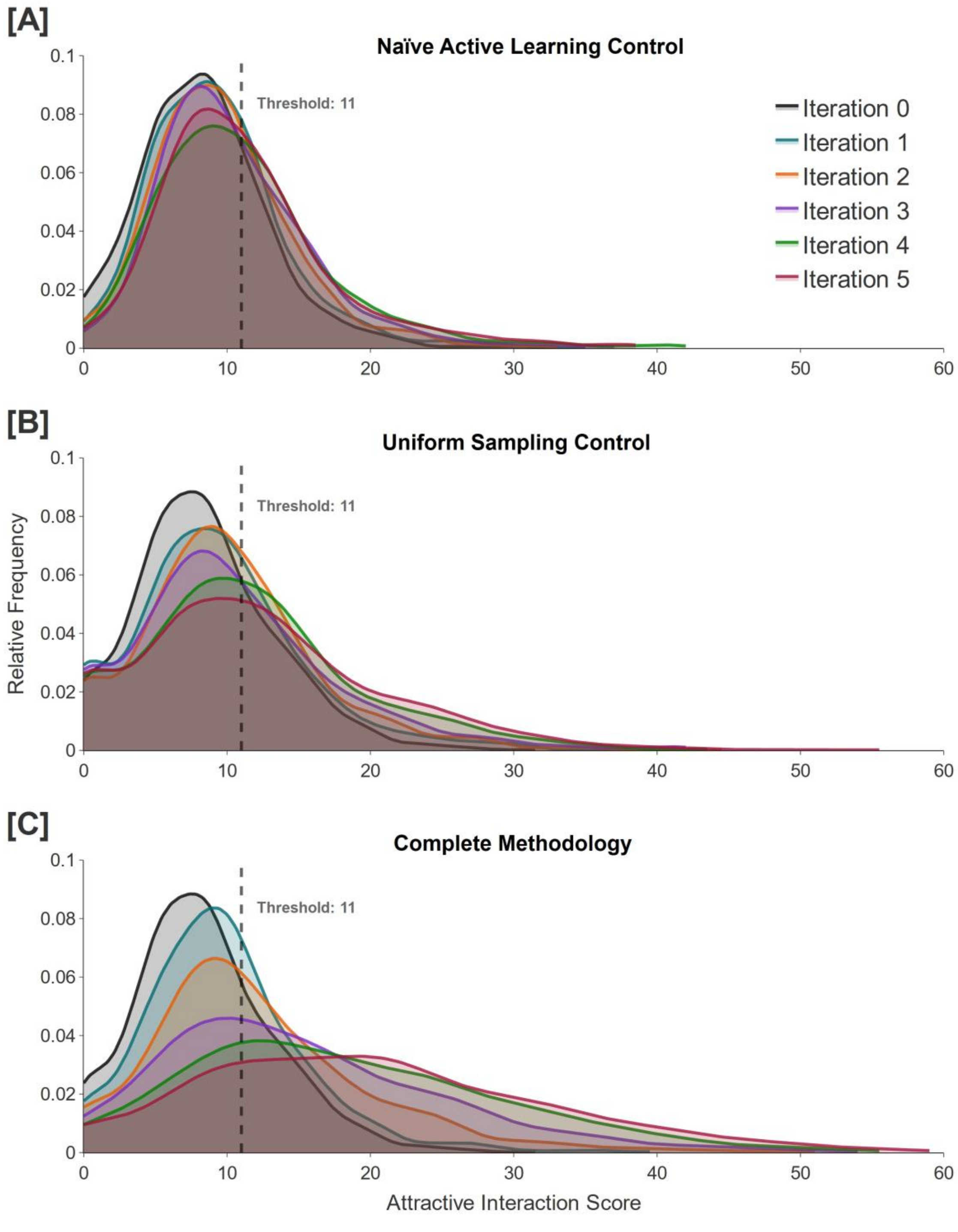

**Figure 6.** Attractive interaction scores of evaluated molecules across five iterations of active learning. Results for the naïve active learning control are shown in **(A)**, which utilizes random selection of molecules and fine-tuning with only replicas of those that score equal to or above the score threshold of 11. Results for the uniform sampling control are shown in **(B)**, which uses cluster-based sampling where each cluster is assigned a sampling fraction $f = 0.01$ during the construction of the active learning set. Results for our complete methodology are shown in **(C)**. Iteration 0 refers to the pretraining phase, while later iterations refer to the active learning phases.

## 4.2. Chemical Space Proxy and Clustering Algorithm

In order to improve upon naïve AL, we propose to strategically select molecules to be in the AL training set that have not been evaluated. This requires a method for relating molecules that have been scored to those that have not. To achieve this goal, we construct a proxy for chemical space that is predicated on molecular properties, allowing us to operate within a space where nearby molecules share similar chemical features. More details regarding the construction of our chemical space proxy are discussed in Section 8.1.

A correlation must exist between position in the chemical space proxy and values produced by the scoring function in order to successfully estimate the scores of molecules that have not been evaluated. Visualizing all of the scored molecules from all iterations of the complete methodology (6,000 molecules) along the first two principal components of our chemical space proxy, we observe a continuous gradient of scores (Figure 7A), illustrating the relation between position in our chemical space proxy and values produced by our scoring function. Moreover, when the positions of the scored molecules in the chemical space proxy are reduced to two dimensions using t-distributed stochastic neighbor embedding (t-SNE), a technique that captures nonlinear structures, we also see that the regions containing molecules with higher scores are easily identifiable (Figure 7B; more details in Section 6 of the *Supporting Information*).

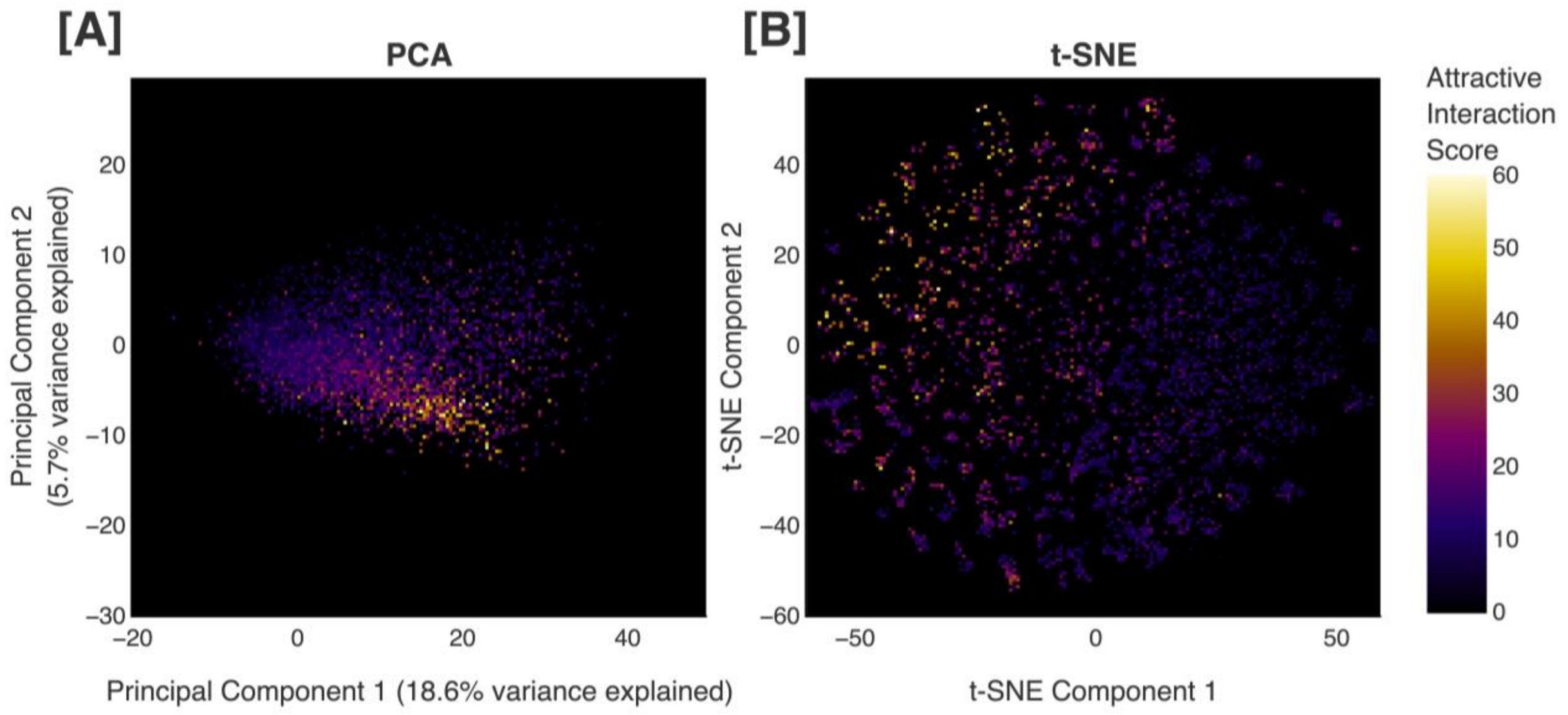

**Figure 7.** Visualization of scored molecules in the chemical space proxy. All of the scored molecules from all iterations of the complete methodology applied to the model pretrained on our combined dataset, aligned to HNH, and with no filters on the generated molecules (6,000 molecules) are displayed. **(A)** Descriptor vectors of the generated molecules projected into the chemical space proxy and shown along the first two principal components, and **(B)** two-dimensional t-distributed stochastic neighbor embedding (t-SNE) plot of the generated molecules are shown. Plots are colored by score obtained with the scoring function, where black/purple corresponds to lower scores and white/yellow corresponds to higher scores.

Within our chemical space proxy, we utilize k-means clustering (with k = 100) to group molecules that exhibit similar chemical properties. We also report results for k = 10, which proves to be less effective (see Figures S7.1 – S7.4 in the *Supporting Information*). This is likely because much of the diversity in the chemical space is homogenized into clusters which, in the case of k = 10, are very large compared to k = 100, and valuable information is lost. In short, we generate 100 clusters and then randomly sample up to 10 molecules from each cluster, selecting all molecules in cases where a cluster contains fewer than 10 molecules. More details of our clustering method are discussed in Section 8.2.

## 4.3. Docking and Scoring

After strategically selecting 1,000 molecules, we dock each of them to a protein target using DiffDock (more details in Section 8 of the *Supporting Information*),[93] and evaluate it using our scoring function, which is essentially a sum of attractive protein-ligand contact points, each weighted by its interaction type. More details regarding the scoring function we use are discussed in Section 8.3. We calculate a score for each of the 1,000 docking poses to serve as an estimate for the ligand's potential to bind the protein target.

## 4.4. Uniform Sampling Control

Because the generated molecules are not evenly distributed in the chemical space proxy, cluster-based sampling introduces a bias in which molecules from less dense regions are sampled more frequently than they would be with random selection. This leads to a score-independent shift in the distribution throughout AL iterations, which we will refer to as the *diffusion effect*. To assess this bias, we construct AL training sets by randomly selecting 10 molecules from each cluster, scoring each of them, selecting the molecules with scores that reach the score threshold (at least 5,000 molecules including replicas), and sampling from each cluster with the same sampling fraction $f$ = 0.01 (about 50 from each cluster for a total of 5,000 molecules) for a total of approximately 10,000 molecules. This approach serves as a control for isolating the effectiveness of our algorithm for sampling unscored molecules to be in the AL training set. For this uniform sampling-based approach, the increase in the scores of the molecules in the generated ensemble after five iterations (28.1% to 51.1%) is slightly more pronounced than that achieved via naïve AL (26.2% to 44.2%), as shown in Figure 6B. Although these results mark a slight improvement over those obtained with naïve AL, they are significantly worse than those achieved with our complete methodology (28.1% to 76.0%), indicating that our score-based sampling method is necessary for high performance and aligns the model with the scoring function much more effectively than does uniform sampling.

## 4.5. Sampling from Clusters Proportionally to Their Scores

In order to improve upon uniform sampling, we propose a way to intelligently weight the importance of each cluster when sampling molecules from the chemical space proxy to be in the AL training set. After scoring each of the 1,000 protein-ligand pairs, we sample from the clusters proportionally to the mean scores calculated from the evaluated molecules within each respective cluster. These sampled molecules are then combined with replicas of the evaluated molecules whose scores meet the score threshold, forming the AL training set. More details regarding our sampling algorithm are discussed in Section 8.4. Our sampling procedure allows us to enrich the AL training set with unscored molecules that would likely obtain high scores, exploiting the fact that position in the chemical space proxy correlates with the scoring function (Figure 7).

Our complete methodology shifts the percentage of generated molecules that reaches the score threshold from 28.1% to 76.0% (Figure 6C). This increase is attributable to the shift of the generated molecular ensemble toward the region of the chemical space proxy associated with higher scores. Figure 8 illustrates this progression, depicting the evolution of the generated ensemble in a constant direction through the chemical space proxy.

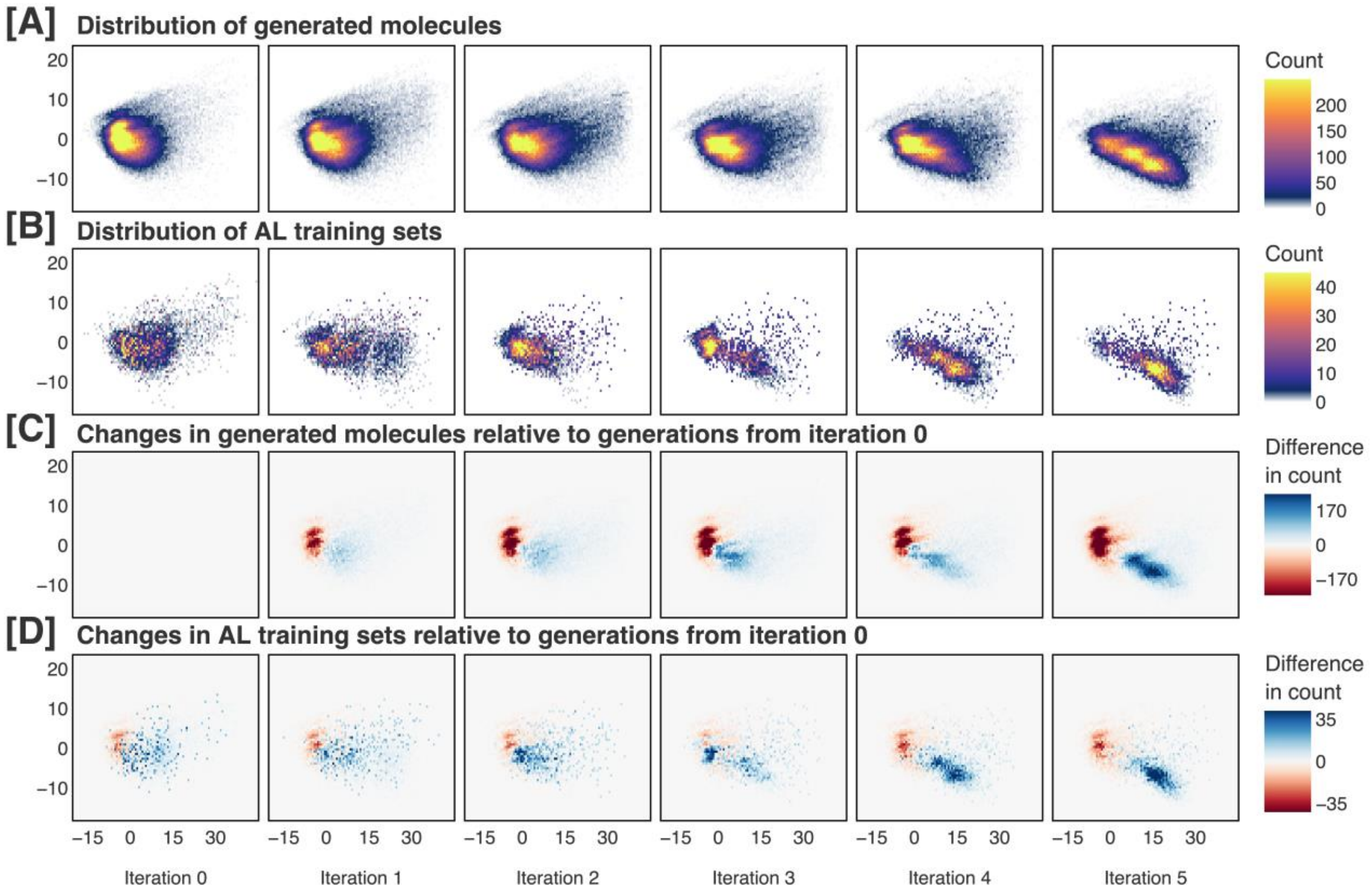

**Figure 8.** Generated molecules and active learning training sets across five iterations of our complete methodology, visualized along the first two principal components of our chemical space proxy. The generated molecular ensembles and active learning training sets at each iteration are shown in **(A)** and **(B)**, respectively. Changes in the generated molecular ensembles and active learning training sets relative to the molecules generated at iteration 0 are shown in **(C)** and **(D)**, respectively. In **(A)** and **(C)**, the 100,000 unique generated molecules from each iteration are used. In **(B)**, the full active learning training sets, each containing approximately 10,000 molecules, are used. In **(D)**, for proper comparison between the generated molecules at iteration 0 and the active learning training sets, 5,000 molecules are randomly sampled from the generated ensemble at iteration 0, and 5,000 molecules are randomly sampled from the active learning training set at each iteration. Iteration 0 refers to the pretraining phase, while later iterations refer to the active learning phases. More details of this analysis are reported in Figure S9.1 in the *Supporting Information*.

# 5. Summary and Future Outlook

In this work, we present an efficient AL methodology that requires evaluation of only a subset of the generated data to successfully align a generative AI model with respect to a specified objective. We demonstrate its applicability in the context of targeted molecular generation by independently enhancing attractive interactions between the molecules in the generated ensemble and two protein targets, namely, c-Abl kinase and the HNH domain of Cas9. When aligning toward c-Abl kinase, we are able to shift the distribution of generated molecules toward the region of the chemical space proxy corresponding to several FDA-approved inhibitors for this target. We also show that our methodology is effective for a protein without any commercially available small-molecule inhibitors, the HNH domain of Cas9. Moreover, we analyze the effectiveness of individual components of our methodology, and show that the integration of these components in our complete approach aligns the model with the scoring function much more effectively than more naïve AL methods.

The generative model, constructed sample space, and scoring function are all highly substitutable within the framework of our methodology, and we therefore envision that it will be adaptable to future innovations. For instance, the GPT-based model could be replaced by a more capable architecture as soon as one is developed. In addition, rather than constructing a sample space from molecular descriptors, any quantifiable features that are correlated with the scoring function can be used. In the context of molecular generation, the list of descriptors used to construct our chemical space proxy could be substituted as better molecular descriptors are developed (i.e., ones that correlate better with the scoring function). Moreover, the scoring function that we use can be replaced by a better metric to achieve closer correspondence with experimental results. The generality of our approach facilitates the applicability and utility of the ChemSpaceAL methodology both at present and as the state of the field inevitably improves.

# 6. Dataset Collection and Preprocessing

## 6.1. Data Collection

We combine all of the SMILES strings from ChEMBL 33, GuacaMol v1, MOSES, and BindingDB, filter out the strings that are identified as invalid by the RDKit molecular parser, and remove any duplicate strings. The resulting combined dataset contains 5,622,772 unique and valid SMILES strings.

## 6.2. Tokenization

Our combined dataset initially has a vocabulary of 196 unique tokens. We find that 148 tokens are represented in the dataset fewer than 1000 times; to reduce the size of our vocabulary (from 196 to 48), we remove all SMILES strings containing at least one token that appears less than 1000 times in the combined dataset (details in Tables S10.1 and S10.2 in the *Supporting Information*). Most of the SMILES strings excluded contain rare transition metals or isotopes.

## 6.3. Data Preprocessing

The longest SMILES string in the combined dataset contains 1,503 tokens, while 99.99% of the strings in the dataset have 133 or fewer tokens (details in Figures S11.1 – S11.2 in the *Supporting Information*). We impose a SMILES string length cutoff of 133, and remove any string from the dataset whose length is greater than this cutoff. All remaining SMILES strings are then extended, if necessary, to the length of the longest SMILES string in the dataset (133) using a padding token “<”, and are augmented with a start token “!” and an end token “~”. The resulting dataset contains 5,539,765 SMILES strings, which are randomly split into training (5,262,776 entries; 95.0%) and validation (276,989 entries; 5.0%) sets for pretraining.

# 7. Details of the Generative Model

We utilize a GPT-based model (details of the model architecture can be found in Section 12 of the *Supporting Information*). Our model embeds inputs into a 256-dimensional space, and is composed of eight transformer decoder blocks, each of which contains eight attention heads. Dropout with a probability of 10% is applied after each feed-forward network except for the output layer to mitigate overfitting, and gradient clipping with a maximum norm of 1.0 is used in conjunction with layer normalization to stabilize the optimization process and prevent exploding gradients. All weights are initialized according to a Gaussian distribution with a mean of 0 and a standard deviation of 0.02 except for weights involved in layer normalization, which are initialized to 1, and bias parameters, which are initialized to 0. The training process utilizes cross-entropy loss with L2 regularization applied to the linear layers using $\lambda$=0.1, and the SophiaG optimizer with $\beta_1$=0.965, $\beta_2$=0.99 and $\rho$=0.04.[94]

## 7.1. Pretraining

During pretraining, the learning rate warms up to $3\times10^{-4}$ until the model has been trained on 10% of the total number of tokens in the dataset, then decays to $3\times10^{-5}$ using cosine decay. The model is trained with a batch size of 512 for 30 epochs. Learning curves are reported in Figures S13.1 and S13.2 in the *Supporting Information*.

## 7.2. Benchmarking

Many generative AI models for molecular discovery have been evaluated with the MOSES benchmark,[77] which constitutes an important standard for the field, with the objective of assessing models' abilities to generate diverse collections of novel and valid molecules. We show that our pretrained model performs among the best in the field (details in Table S14.1 and S14.2 in the *Supporting Information*), establishing its merit as a starting point for AL.

## 7.3. Fine-tuning

After compiling the AL training set, the model is further trained with a batch size of 512 for 10 epochs using a learning rate of $3\times10^{-5}$, with no warmup and a cosine decay to $3\times10^{-6}$.

# 8. Details of the ChemSpaceAL Methodology

## 8.1. Chemical Space Proxy

We first calculate the full set of molecular descriptors available through RDKit's CalcMolDescriptors function for each molecule in the combined pretraining set, encompassing a wide range of molecular properties including structural, topological, geometrical, electronic and thermodynamic characteristics. Among these 209 descriptors, 13 return NaN (not a number) or infinity for at least one SMILES string in the dataset and consequently are discarded (details in Table S15.1 in the *Supporting Information*), resulting in 196 descriptors (details in Table S15.2 in the *Supporting Information*). We use as many RDKit descriptors as possible because this step in the methodology is very fast (see Figure S16.1 in the *Supporting Information*), enabling us to generate maximally descriptive molecular representations. We also independently investigate the performance of the methodology using only the 42 RDKit molecular quantum numbers (MQNs), which are not included in the CalcMolDescriptors function, and find this representation to yield worse results than those obtained using the PCA-reduced 120-dimensional representation of the 196 descriptors (Figure S17.1 in the *Supporting Information*). We note that for the proposed methodology to work, the set of descriptors used must satisfy two criteria: 1) position in the chemical space proxy correlates with the scoring function, and 2) nearby molecules in the chemical space proxy have similar scores. It is evident that there could exist many sets of descriptors satisfying these criteria; a thorough investigation into the choice of descriptors is outside the scope of this work. After performing PCA using the 196 RDKit descriptors for all molecules in the combined pretraining set, we find that 99% of the variance is explained by the first 113 principal components (details in Figure S15.3 of the *Supporting Information*), and use the first 120 principal components throughout the methodology as our chemical space proxy. Our methodology might attain similar results with fewer principal components retained, but this reduction is not necessary since this step is computationally inexpensive.

## 8.2. Clustering Algorithm

Within our chemical space proxy, we utilize k-means clustering to group molecules that exhibit similar chemical properties, with k = 100. Given that running k-means is incredibly fast, we perform k-means 100 times to mitigate the potential for poor initialization, seeking to minimize k-means loss and cluster size variance. Initially, we take the five clusterings with the lowest loss, thereby preserving those with more compact clusters. Of these five, we select the clustering with the lowest variance in cluster size for use in the following stages of the methodology.

After clustering the generated molecules in our chemical space proxy, we randomly select 10 molecules from each cluster that contains at least 10 molecules, and select all of the molecules from any cluster that contains less than 10 molecules. For AL iterations 1-5, when applying the methodology to the C model for aligning to HNH with no filters on the generated molecules, the number of clusters containing fewer than 10 molecules out of 100 clusters are 4, 3, 5, 2, and 3 for each respective iteration (see Figures S18.1 – S18.4 in the *Supporting Information*). We then randomly sample from the clusters with more than 10 molecules until we achieve a set of 1,000 molecules.

## 8.3. Scoring Function

Our scoring function considers attractive protein-ligand contact points using the prolif software package,[95] and assigns handpicked weights for each interaction type: hydrophobic interactions are scored at 2.5; hydrogen-bond interactions at 3.5; ionic interactions at 7.5; interactions between aromatic rings and cations at 2.5; Van der Waals interactions at 1.0; halogen-bond interactions at 3.0; face-to-face pi-stacking interactions at 3.0; edge-to-face pi-stacking interactions at 1.0; and metallic complexation interactions at 3.0.

We assess our scoring function with the PDBbind v2020 refined set, which contains 5,316 unique experimentally determined protein-ligand binding complexes with high-quality labels and structures.[92] We find that there is a positive Pearson correlation of 0.32 between the scores derived from our scoring function and the experimentally determined binding affinities (Figure S19.1A in the *Supporting Information*), supporting our scoring function as an approximate yet meaningful estimate of binding ability. Furthermore, we find that 99.6% of the complexes achieve the score threshold of 11 (Figure S19.1B in the *Supporting Information*).

The optimal weights for the interaction types may vary significantly depending on the specific target, and therefore the scoring function employed in this work should be considered a crude estimation. However, the positive correlation with experimentally determined binding affinities supports its utility as a heuristic approximation to binding ability. Moreover, it can be replaced with a more precise metric, as long as the replacement metric correlates with the descriptors used to construct the chemical space proxy.

## 8.4. Sampling Algorithm

After scoring each of the 1,000 protein-ligand pairs, we select $N$ replicas of each molecule that scores equal to or above the score threshold, where $N$ is the smallest integer that achieves a total number of molecules of at least 5,000. We then calculate mean cluster scores from the scored molecules, which are converted to sampling fractions with the softmax function. We also consider other methods for converting cluster scores to sampling fractions and report the results for each method attempted (Figures S20.1 – S20.17 in the *Supporting Information*). We then convert $f_i \times 5{,}000$ to an integer (where $f_i$ is the calculated fraction for sampling from cluster $i$), and sample the corresponding number of molecules randomly from each respective cluster. If a given cluster has fewer molecules than would satisfy the calculated fraction, we distribute the surplus among the other clusters relative to their sampling fractions. We combine these 5,000 molecules with the replicas of molecules that meet the scoring threshold to generate an AL training set of approximately 10,000 molecules.

## Supporting Information

The Supporting Information contains figures and tables regarding the following aspects: ADMET and functional group filters; similarity between FDA-approved inhibitors of c-Abl kinase; comparing the generations to c-Abl kinase inhibitors for different methods; radar plots showing evolution of ADMET metrics; scores of molecules across five iterations of active learning for HNH; implementation details of t-distributed stochastic neighbor embedding (t-SNE); choosing the number of clusters to use for k-means; details and parameters used for running DiffDock; t-SNE visualization of the evolution of the generated molecular ensembles; vocabulary composition of the combined dataset; frequencies of block sizes, molecular weights and tokens in pretraining sets; details of the GPT architecture; training the GPT model; pretrained GPT model performance on the MOSES benchmark; RDKit descriptors used to construct the chemical space proxy; wall times of each step in the complete pipeline; evaluating the methodology with lower-dimensional MQN filters; frequency as a function of cluster size for alignment to c-Abl kinase; evaluation of scoring function compared to PDBbind v2020 refined set; alternative methods for converting mean cluster scores to sampling fractions; distributions of mean and median cluster scores; additional evaluation of generations across active learning iterations.

The following files are available free of charge:
DOC and PDF files of Supporting Information

## Data and Software Availability

All of our software is available as open source at https://github.com/batistagroup/ChemSpaceAL. Additionally, the ChemSpaceAL Python package is available via PyPI at https://pypi.org/project/ChemSpaceAL/.

## Author Information

**Corresponding Author**: Victor S. Batista
Phone: (203) 432-6672
Email: victor.batista@yale.edu

**Present Address**: Department of Chemistry, Yale University, New Haven, Connecticut 06511-8499

**Author Contributions**
GWK, AM, RIB designed research; GWK, AM, RIB developed software; GWK, AM, RIB published the Python package; GWK, AM, RIB performed research; GWK, AM, RIB, VSB analyzed data; and GWK, AM, RIB wrote the paper. All authors have given approval to the final version of the manuscript.

[‡]GWK, AM, RIB contributed to this work equally

[*]VSB is corresponding author

**Funding Sources**
National Institutes of Health: Grants R01GM136815
National Science Foundation: Grant DGE-2139841

## Acknowledgments

We acknowledge financial support from the National Institutes of Health under Grant R01GM136815, as well as from the National Science Foundation under Grant DGE-2139841. VSB also acknowledges high-performance computer time from the National Energy Research Scientific Computing Center and from the Yale University Faculty of Arts and Sciences High Performance Computing Center.

## Abbreviations

AI, artificial intelligence; RNNs, recurrent neural networks ; GNNs, generative adversarial networks; AL, active learning; GPT, Generative Pretrained Transformer; Cas9, CRISPR-associated protein 9; SMILES, Simplified Molecular Input Line Entry System; PCA, Principal Component Analysis; ADMET, Absorption, Distribution, Metabolism, Excretion, and Toxicity; $T_C$, Tanimoto similarity; LogP, logarithm of the partition coefficient; t-SNE, t-distributed stochastic neighbor embedding; NaN, not a number; MQNs, molecular quantum numbers.

# Supporting Information

ChemSpaceAL: An Efficient Active Learning Methodology Applied to Protein-Specific Molecular Generation

Gregory W. Kyro, Anton Morgunov, Rafael I. Brent, Victor S. Batista

Department of Chemistry, Yale University, New Haven, Connecticut 06511-8499

# List of Sections

## **Section 1:** ADMET and Functional Group Filters.

**Table S1.1.** Upper and lower bounds applied to each ADMET metric used for generation filter. We use upper bound for logP of 6.5 because one of the FDA-approved inhibitors of c-Abl kinase, nilotinib, has a value of 6.356 as calculated by RDKit. All other bounds are taken from ADMETlab 2.0.[1]

| ADMET Property | Lower Bound | Upper Bound |
|---|---|---|
| Molecular Weight | 100 | 600 |
| Number of Hydrogen Bond Acceptors | 0 | 12 |
| Number of Hydrogen Bond Donors | 0 | 7 |
| Number of Rotatable Bonds | 0 | 11 |
| Number of Rings | 0 | 6 |
| Number of Heteroatoms | 1 | 15 |
| Formal Charge | -4 | 4 |
| Topological Polar Surface Area | 0 | 140 |
| LogP | -0.4 | 6.5 |

**Table S1.2.** List of functional groups excluded by generation filter.

- ‘fr_azide’
- ‘fr_isocyan’
- ‘fr_isothiocyan’
- ‘fr_nitro’
- ‘fr_nitro_arom’
- ‘fr_nitro_arom_nonortho’
- ‘fr_nitroso’
- ‘fr_phos_acid’
- ‘fr_phos_ester’
- ‘fr_sulfonamd’
- ‘fr_sulfone’
- ‘fr_term_acetylene’
- ‘fr_thiocyan’
- ‘fr_prisulfonamd’
- ‘fr_C_S’
- ‘fr_azo’
- ‘fr_diazo’
- ‘fr_epoxide’
- ‘fr_ester’
- ‘fr_COO2’
- ‘fr_Imine’
- ‘fr_N_O’
- ‘fr_SH’
- ‘fr_aldehyde’
- ‘fr_dihydropyridine’
- ‘fr_hdrzine’
- ‘fr_hdrzone’
- ‘fr_ketone’
- ‘fr_thiophene’
- ‘fr_phenol’

**<u>Section 2</u>:** Similarity between FDA-Approved Inhibitors of C-Abl Kinase.

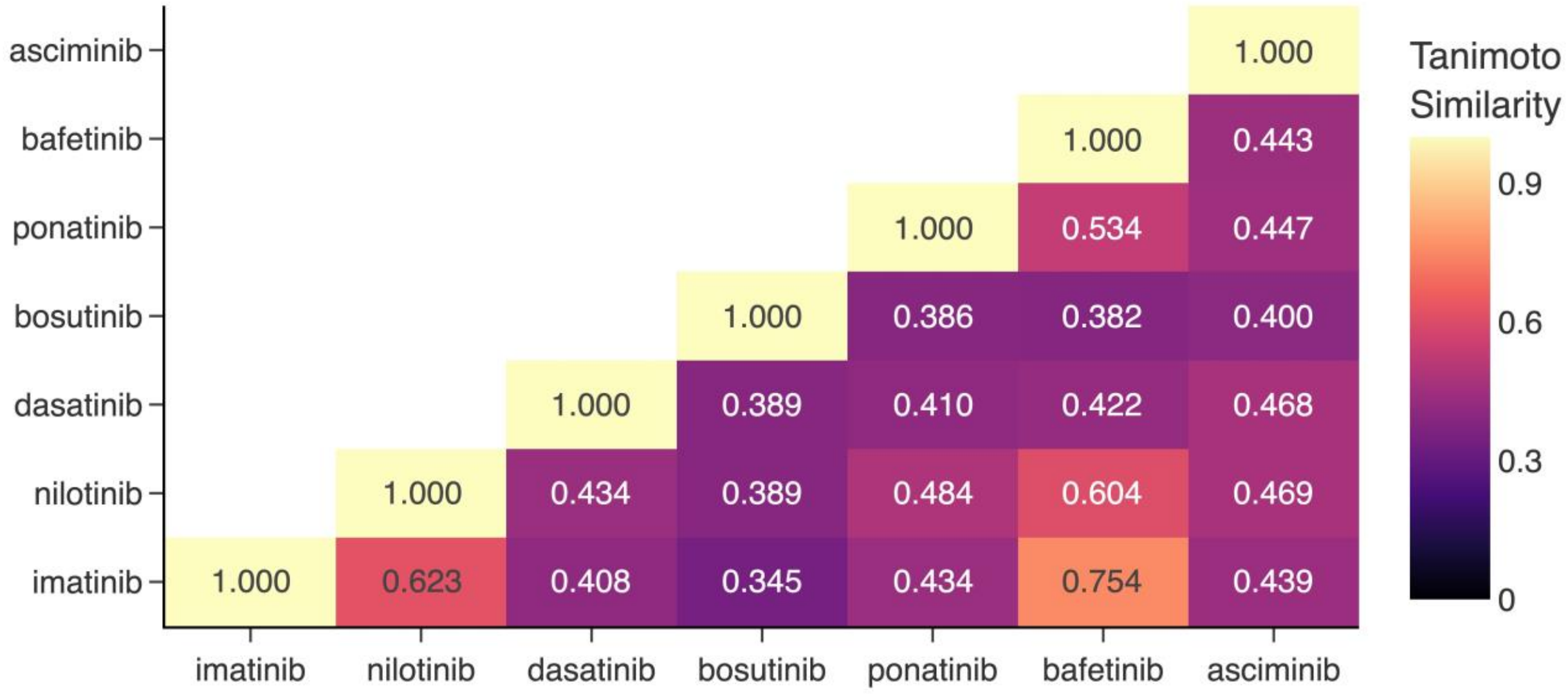

**Figure S2.1.** Tanimoto Similarity between RDKit fingerprints of the FDA-approved inhibitors of c-Abl kinase: imatinib, nilotinib, dasatinib, bosutinib, ponatinib, bafetinib, and asciminib.

**Section 3:** Comparing the Generations to c-Abl Kinase Inhibitors for Different Methods.

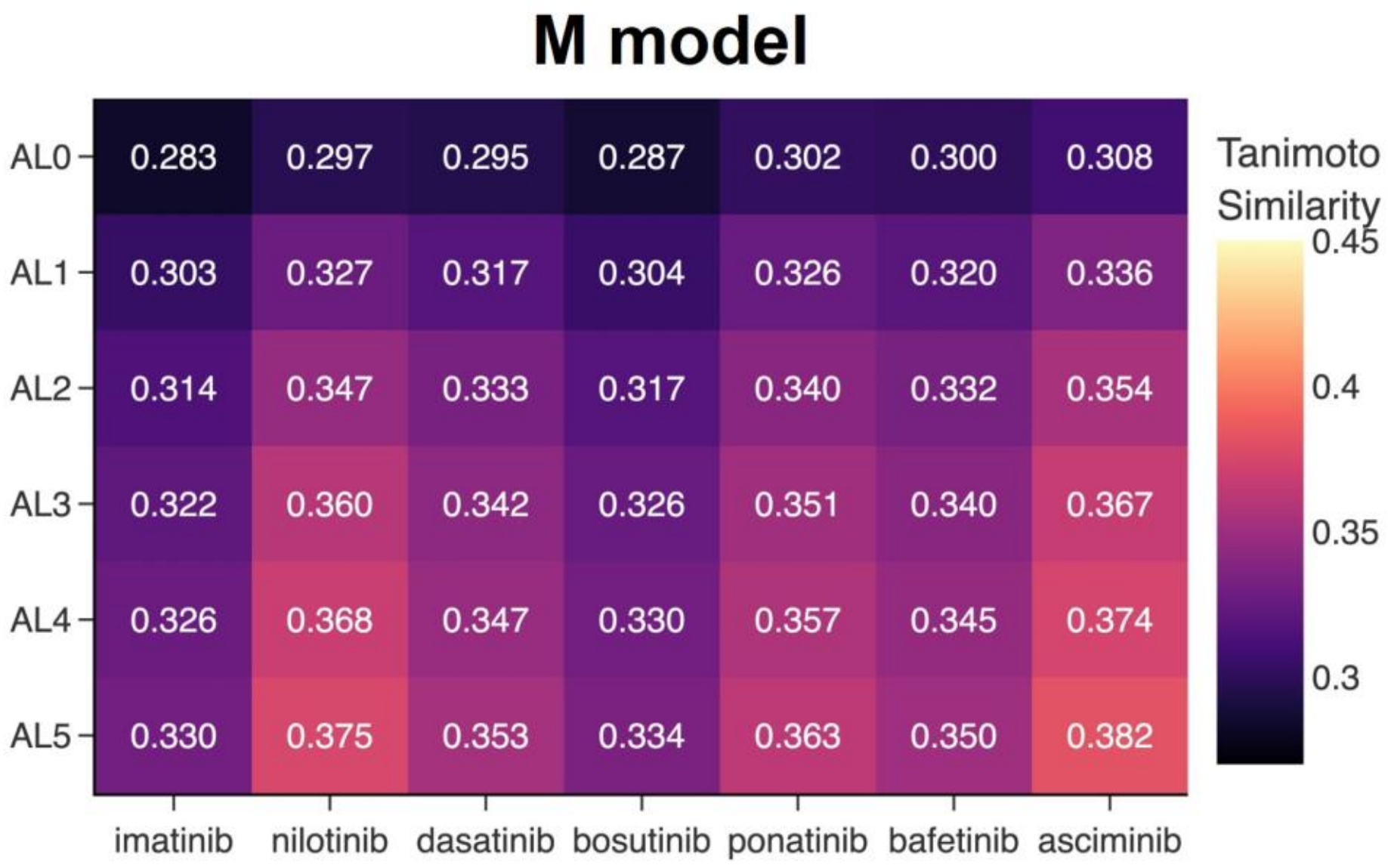

**Figure S3.1.** Visualizing the evolution of the generated molecular ensemble from the model pretrained on the MOSES dataset with the generation filtered based on ADMET metrics and functional group restrictions, and comparing it to the FDA-approved small-molecule inhibitors of c-Abl kinase. The average Tanimoto similarities between the RDKit fingerprints of all generated molecules at each iteration of the pipeline and each inhibitor are shown. Iteration 0 refers to the pretraining phase, while later iterations refer to the active learning phases

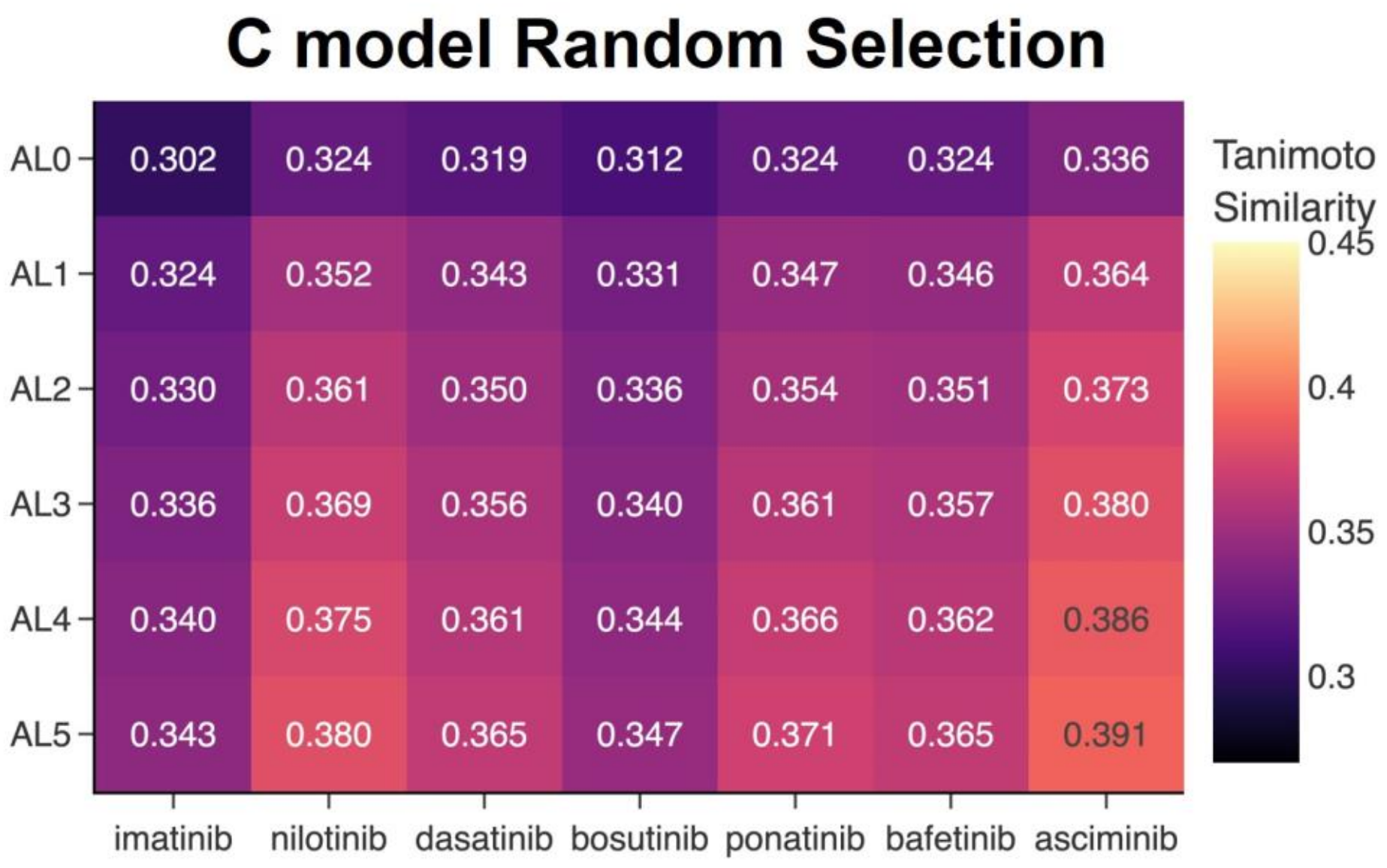

**Figure S3.2.** Visualizing the evolution of the generated molecular ensemble from the model utilizing random selection (i.e., 1,000 molecules are randomly selected from the generated ensemble and scored, and those that satisfy the score threshold and replicated $N$ times where $N$ is the smallest integer to achieve a total of 5,000 datapoints to be in the active learning set), pretrained on the combined dataset with the generation filtered based on ADMET metrics and functional group restrictions, and comparing it to the FDA-approved small-molecule inhibitors of c-Abl kinase. The average Tanimoto similarities between the RDKit fingerprints of all generated molecules at each iteration of the pipeline and each inhibitor are shown. Iteration 0 refers to the pretraining phase, while later iterations refer to the active learning phases

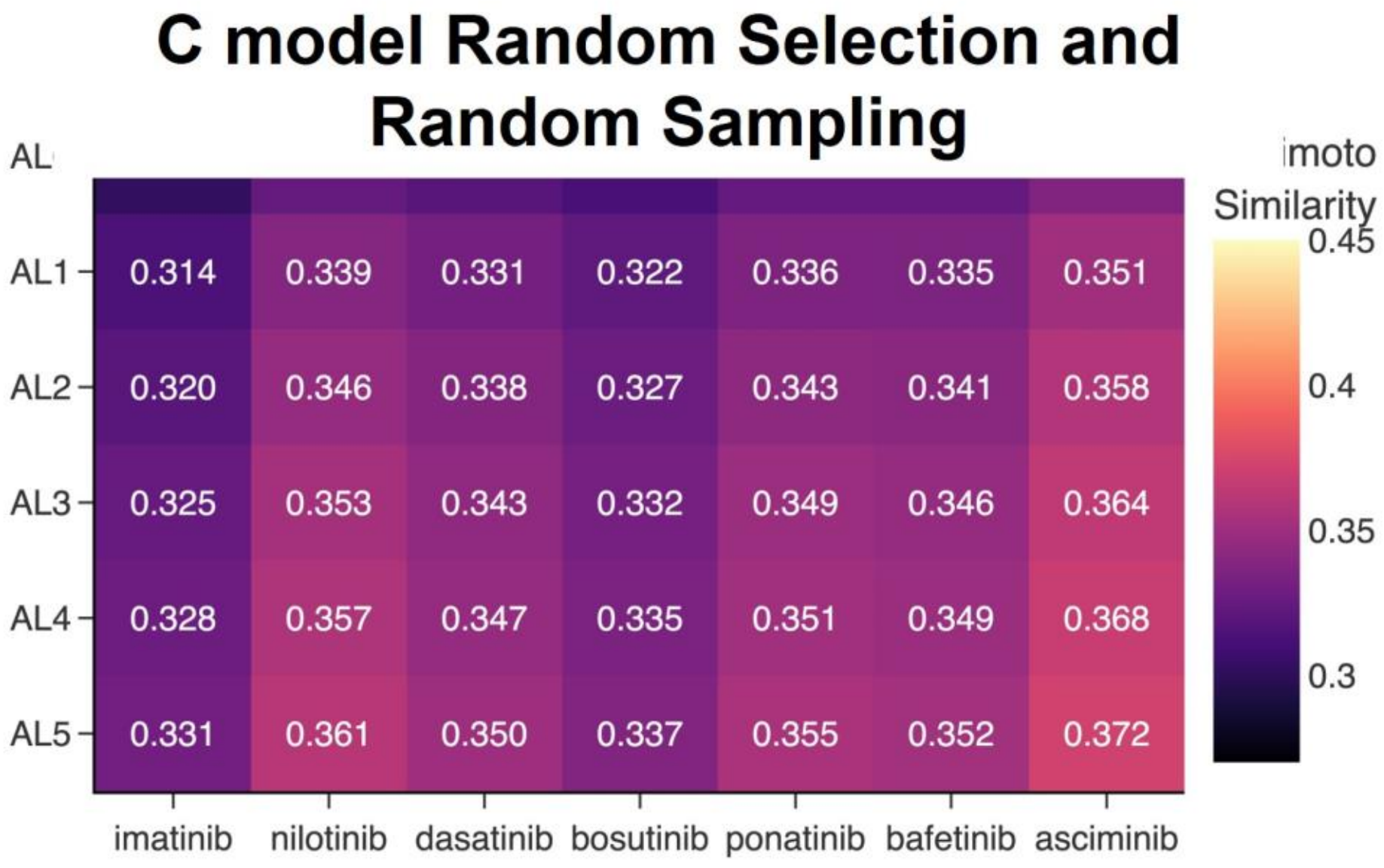

**Figure S3.3.** Visualizing the evolution of the generated molecular ensemble from the model utilizing random selection with random sampling (i.e., random selection with the addition of randomly sampling 5,000 molecules from the generated ensemble that have not been scored to be in the active learning training set), pretrained on the combined dataset with the generation filtered based on ADMET metrics and functional group restrictions, and comparing it to the FDA-approved small-molecule inhibitors of c-Abl kinase. The average Tanimoto similarities between the RDKit fingerprints of all generated molecules at each iteration of the pipeline and each inhibitor are shown. Iteration 0 refers to the pretraining phase, while later iterations refer to the active learning phases.

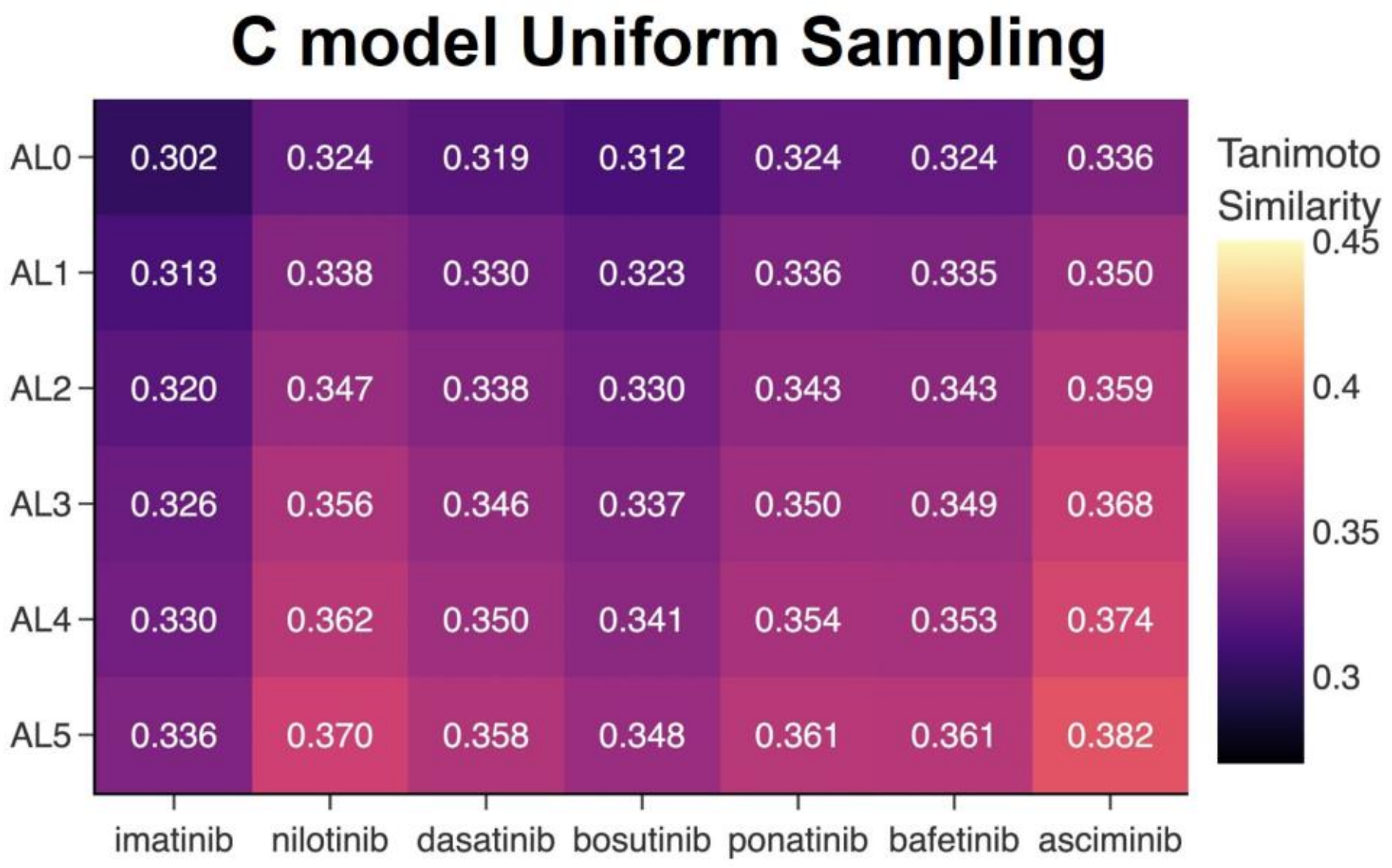

**Figure S3.4.** Visualizing the evolution of the generated molecular ensemble from the model utilizing uniform sampling (i.e., cluster-based sampling where each cluster is assigned a sampling fraction $f$ = 0.01 to generate the active learning set), pretrained on the combined dataset with the generation filtered based on ADMET metrics and functional group restrictions, and comparing it to the FDA-approved small-molecule inhibitors of c-Abl kinase. The average Tanimoto similarities between the RDKit fingerprints of all generated molecules at each iteration of the pipeline and each inhibitor are shown. Iteration 0 refers to the pretraining phase, while later iterations refer to the active learning phases

**Section 4:** Radar Plots Showing Evolution of ADMET Metrics.

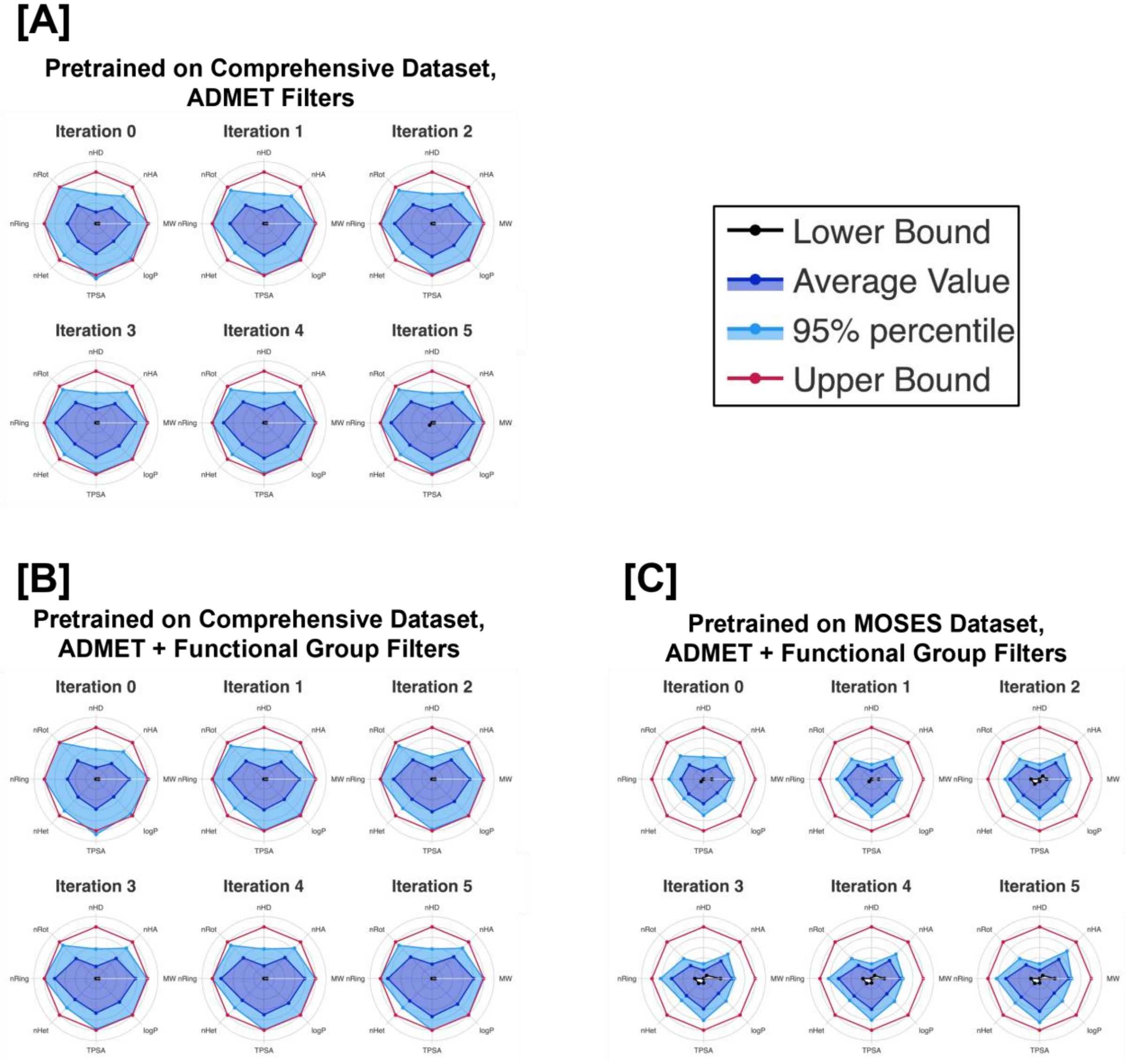

**Figure S4.1.** Radar charts for c-Abl kinase depicting the mean and 95$^{th}$ percentile values for each ADMET metric with respect to the lower and upper bounds enforced for the generated molecular ensembles from the model pretrained on the combined dataset with the generations filtered based on ADMET metrics are shown in (A), the charts for the ensembles from the model pretrained on the combined dataset with the generations filtered based on ADMET metrics and functional group restrictions are shown in (B), and the ensemble for the model pretrained on the MOSES dataset with the generations filtered based on ADMET metrics and functional group restrictions are shown in (C). Iteration 0 refers to the pretraining phase, while later iterations refer to the active learning phases.

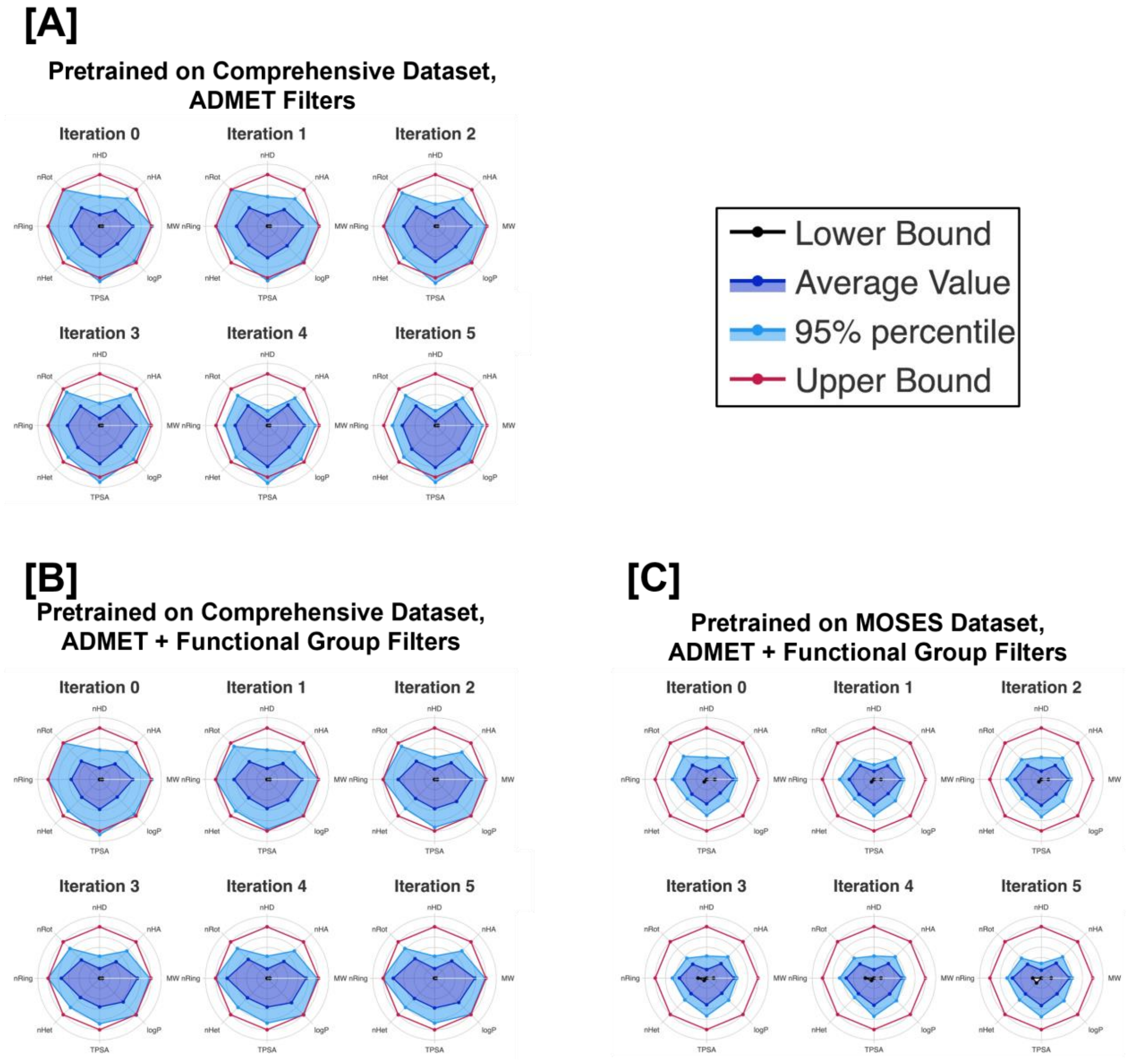

**Figure S4.2.** Radar charts for the HNH domain of Cas9 depicting the mean and 95$^{th}$ percentile values for each ADMET metric with respect to the lower and upper bounds enforced for the distribution for the model pretrained on the combined dataset with generation conditioned on ADMET filters are shown in (A), the distribution for the model pretrained on the combined dataset with generation conditioned on ADMET and functional group filters are shown in (B), and the distribution for the model pretrained on the MOSES dataset with generation conditioned on ADMET and functional group filters are shown in (C). Iteration 0 refers to the pretraining phase, while later iterations refer to the active learning phases.

**Section 5:** Scores of Molecules across Five Iterations of Active Learning for HNH.

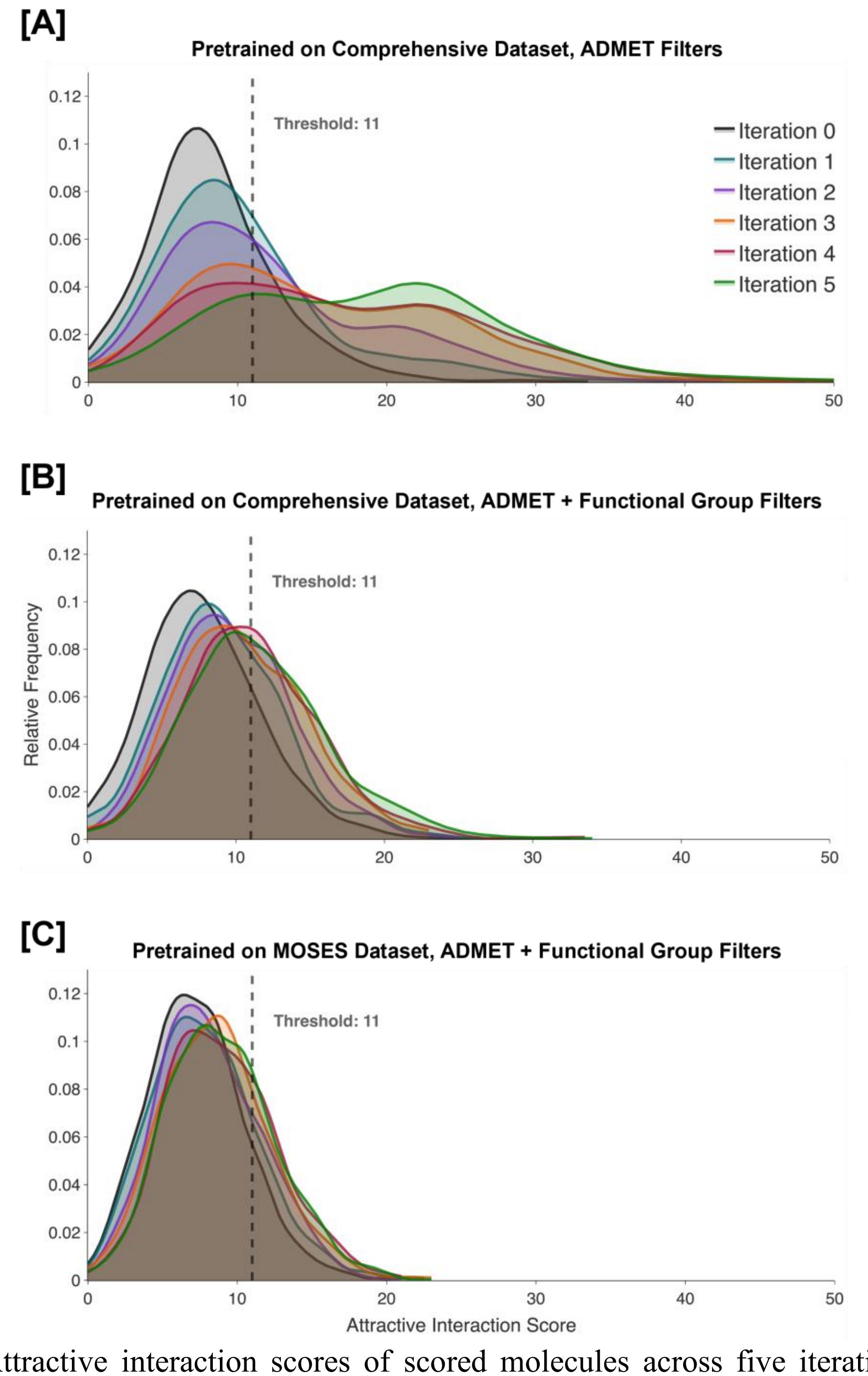

**Figure S5.1**. Attractive interaction scores of scored molecules across five iterations of active learning for the HNH domain of Cas9. The distribution for the model pretrained on the combined dataset with the generations filtered based on ADMET metrics are shown in (A). The distributions for the model pretrained on the combined dataset with the generations filtered based on ADMET metrics and functional group restrictions are shown in (B). The distributions for the model pretrained on the MOSES dataset with the generations filtered based on ADMET metrics and functional group restrictions are shown in (C). Iteration 0 refers to the pretraining phase, while later iterations refer to the active learning phases.

**<u>Section 6</u>:** Implementation Details of t-Distributed Stochastic Neighbor Embedding (t-SNE).

To create a standard t-SNE space which involves a constant coordinate system, we proceed as follows. Firstly, we collect scored molecules from all iterations (6,000 molecules). Secondly, we add the molecules from all active learning training sets that employed either softmax or uniform selection methods. Thirdly, we add a random sample of 10,000 molecules from the set of generations at each iteration. We perform this sampling to have the same number of molecules from the active learning training sets and generated sets, which enables us to fairly compute the difference in distributions. Note that our training sets usually contain slightly more than 10,000 molecules, so we sample exactly 10,000 for consistency. After combining all molecules and dropping all duplicates, we perform a t-SNE reduction.

**<u>Section 7</u>:** Choosing the Number of Clusters to Use for k-means.

For each implementation of k-means, we utilize the scikit-learn Python package,[21] which employs the k-means++ initialization algorithm, where the first centroid is selected randomly and subsequent centroids are iteratively chosen with a probability proportional to their squared distance from the nearest existing centroid

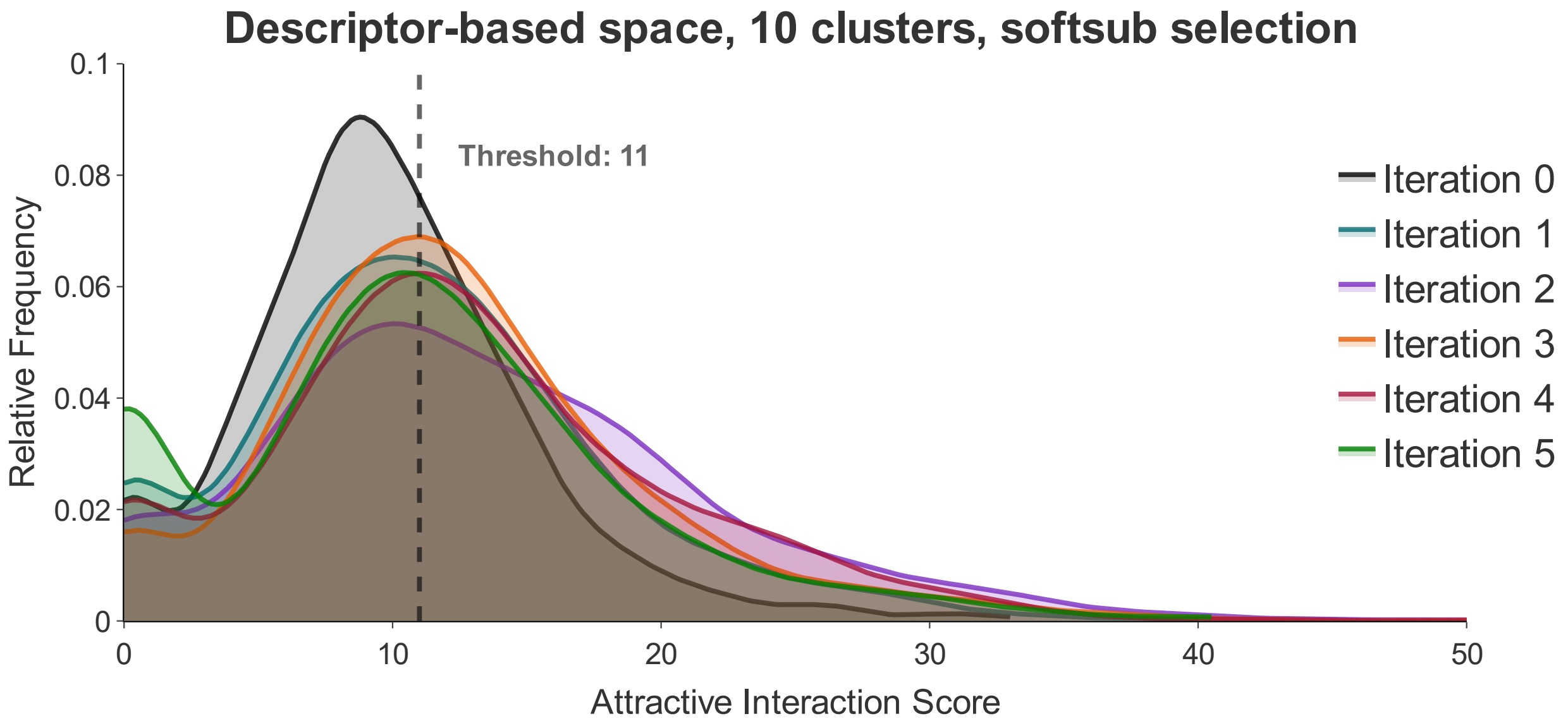

**Figure S7.1.** Attractive interaction scores for molecules generated by the pretrained model (iteration 0) and by the model after each of the five iterations of active learning where, prior to sampling for docking, molecules in the chemical space are grouped into 10 clusters. Cluster scores are converted into sampling fractions using the *softsub* approach.

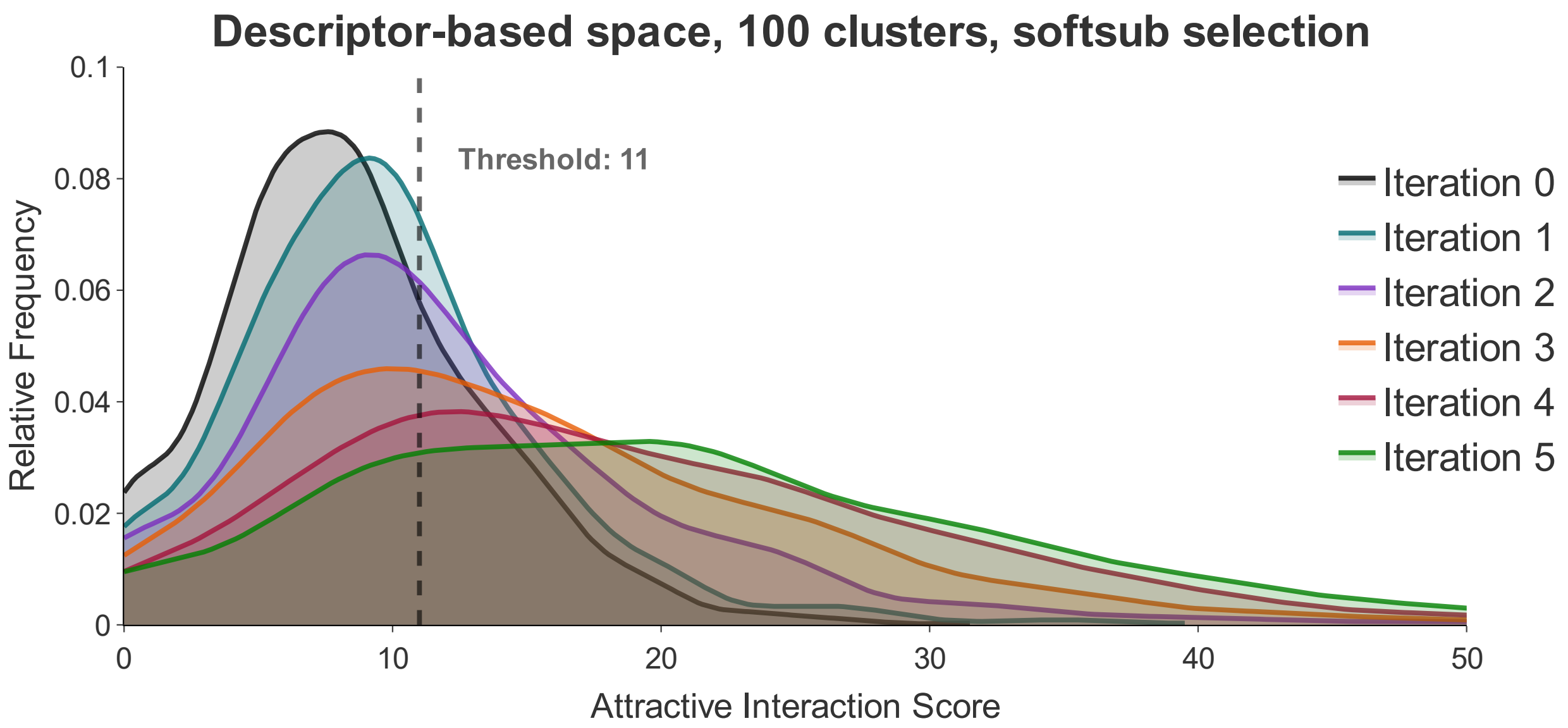

**Figure S7.2.** Attractive interaction scores for molecules generated by the pretrained model (iteration 0) and by the model after each of the five iterations of active learning where, prior to sampling for docking, molecules in the chemical space are grouped into 100 clusters. Cluster scores are converted into sampling fractions using the *softsub* approach. This figure occurs in the main text (Figure 4C), but is also shown here for comparison.

**Table S7.3.** Statistics of the distribution of attractive interaction scores, when molecules are clustered into 10 groups and cluster scores are converted into sampling fractions using the *softsub* method.

| Iteration | Percent > 11 | Q1 | Q2 | Mean | Q3 | Max | Std |
|---|---|---|---|---|---|---|---|
| 0 | 14.29 | 8.26 | 8.91 | 8.75 | 10.28 | 11.25 | 2.25 |
| 1 | 42.86 | 8.96 | 10.27 | 10.67 | 11.49 | 16.09 | 2.62 |
| 2 | 57.14 | 10.32 | 11.51 | 13.31 | 15.34 | 21.00 | 3.91 |
| 3 | 37.50 | 6.80 | 10.45 | 8.88 | 12.51 | 14.94 | 5.44 |
| 4 | 37.50 | 7.33 | 9.10 | 9.03 | 11.74 | 14.12 | 3.83 |
| 5 | 25.00 | 7.50 | 8.75 | 8.67 | 10.99 | 12.90 | 2.96 |

[a] The percentage of generated molecules with attractive interaction scores equal to or above our score threshold is shown (Percent > 11), as well as the score at the first quartile (Q1), second quartile (Q2), Mean, third quartile (Q3), maximum (Max), and standard deviation (Std) of the distribution.

[b] Iteration 0 refers to the pretraining phase, while later iterations refer to the active learning phases.

**Table S7.4.** Statistics of the distribution of attractive interaction scores, when molecules are clustered into 100 groups and cluster scores are converted into sampling fractions using the *softsub* method.

| Iteration | Percent > 11 | Q1 | Q2 | Mean | Q3 | Max | Std |
|---|---|---|---|---|---|---|---|
| 0 | 28.10 | 5.50 | 8.00 | 8.46 | 11.50 | 31.50 | 4.89 |
| 1 | 37.00 | 6.00 | 9.00 | 9.76 | 12.50 | 39.50 | 5.63 |
| 2 | 49.70 | 7.50 | 10.50 | 12.22 | 16.00 | 51.00 | 7.93 |
| 3 | 62.60 | 8.00 | 13.50 | 15.14 | 20.63 | 54.00 | 9.70 |
| 4 | 72.90 | 10.00 | 16.50 | 18.25 | 25.00 | 55.50 | 10.90 |
| 5 | 76.00 | 11.00 | 19.00 | 20.08 | 27.63 | 59.00 | 11.90 |

[a] The percentage of generated molecules with attractive interaction scores equal to or above our score threshold is shown (Percent > 11), as well as the score at the first quartile (Q1), second quartile (Q2), Mean, third quartile (Q3), maximum (Max), and standard deviation (Std) of the distribution.

[b] Iteration 0 refers to the pretraining phase, while later iterations refer to the active learning phases.

<u>**Section 8**</u>**:** Details and Parameters Used for Running DiffDock.

DiffDock handles all of the ligand preparation; we simply provide it with a protein structure and ligand SMILES string. It should be noted that since DiffDock is a diffusion generative model, it is inherently stochastic in nature During the docking inference stage, we utilize 20 inference steps, 10 samples for each complex, and a batch size of 6. Utilizing RDKit, DiffDock uses the MolFromSmiles module to process ligands, adds hydrogen atoms, and retrieves the 3D atomic coordinates with the AllChem.EmbedMolecule module employing the ETKDGv2 methodology.

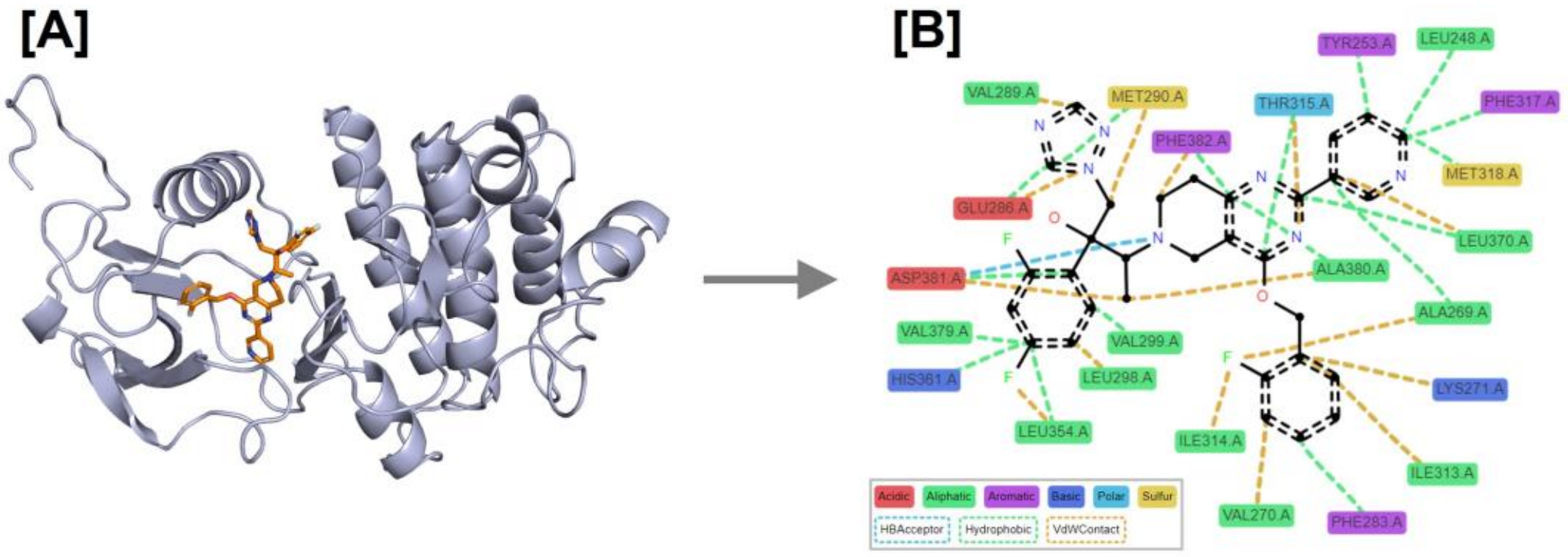

**Figure S8.1.** Generated molecule docked to the c-Abl kinase **(A)** with the corresponding protein-ligand fingerprint **(B)**.

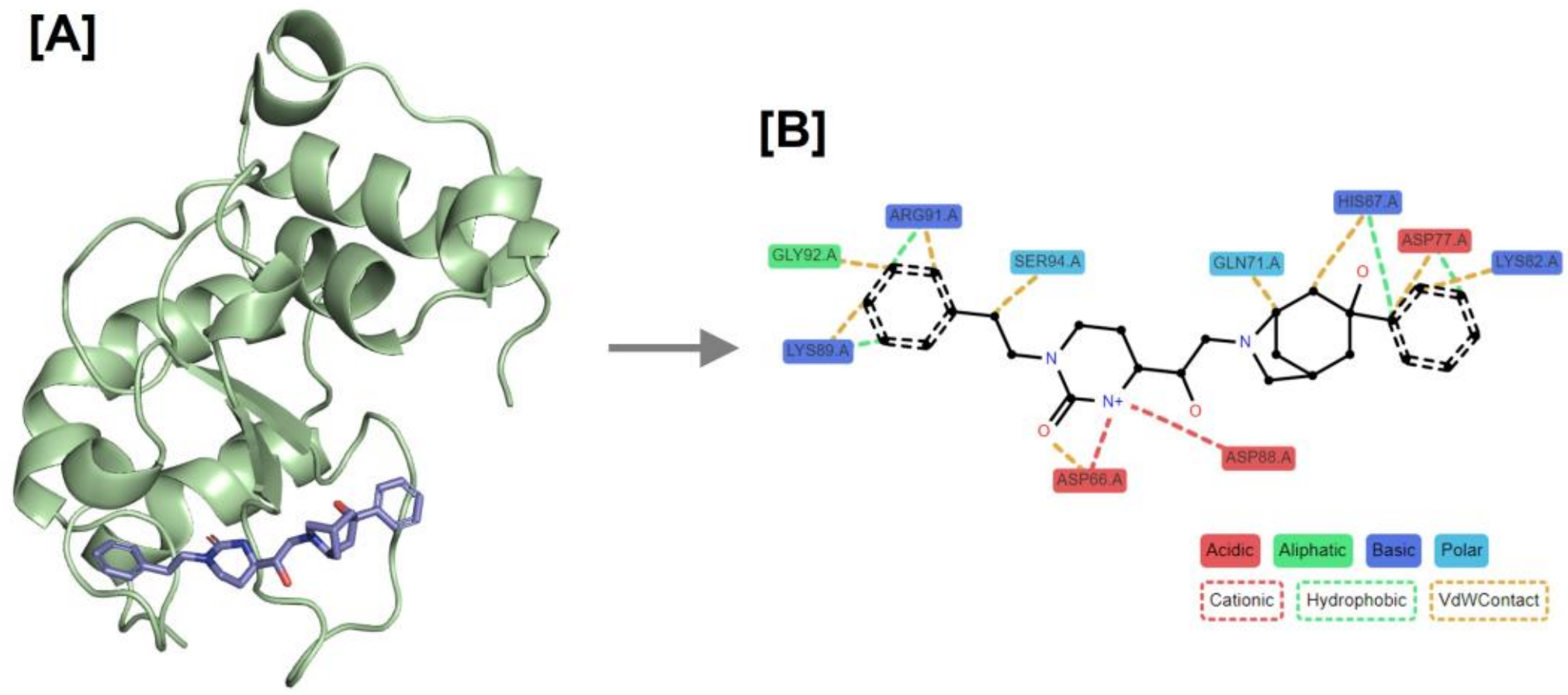

**Figure S8.2.** Generated molecule docked to the HNH domain of Cas9 **(A)** with the corresponding protein-ligand fingerprint **(B)**.

<u>**Section 9**</u>**:** t-SNE Visualization of the Evolution of the Generated Molecular Ensembles.

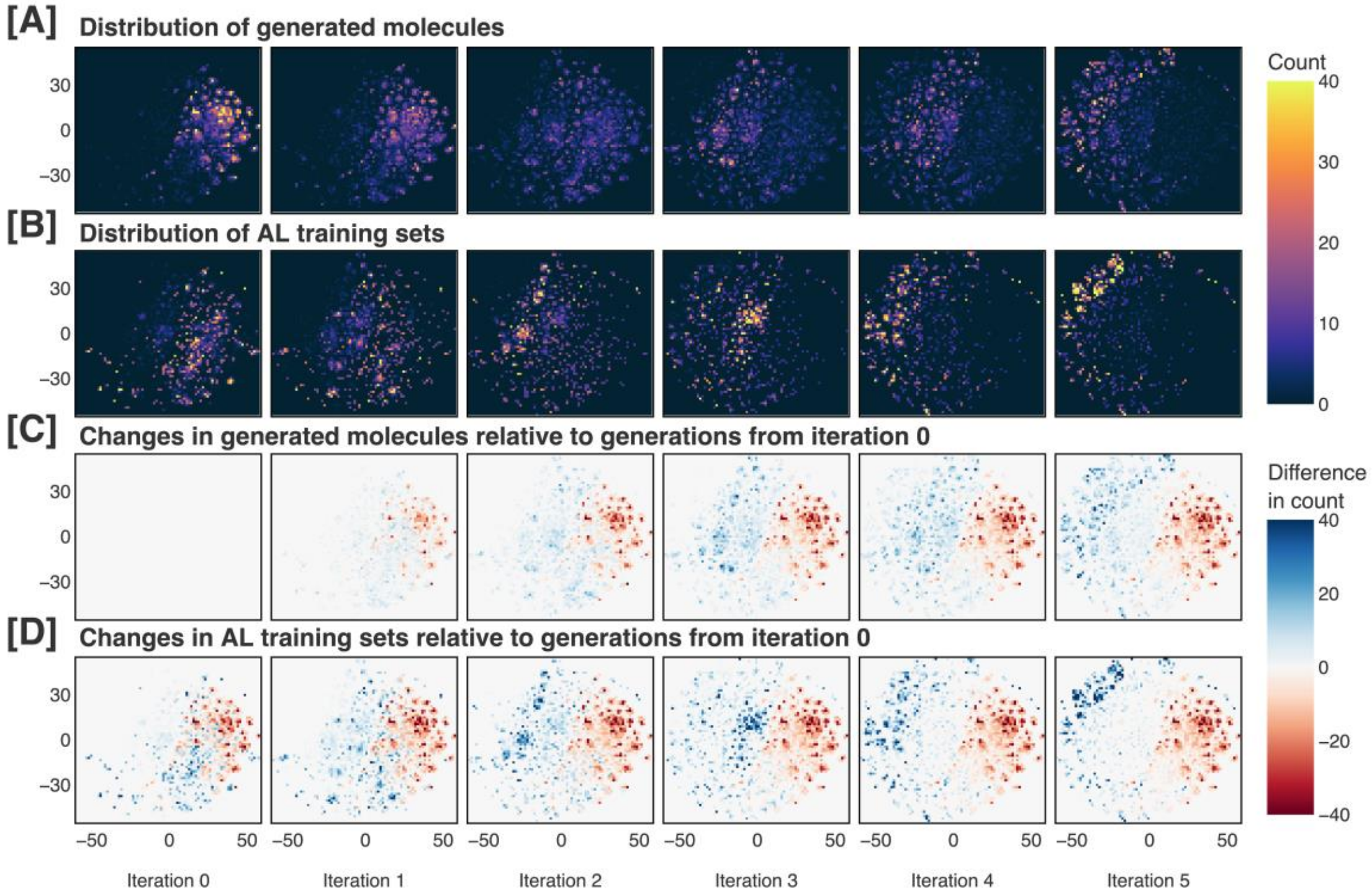

**Figure S9.1.** Generated molecules and active learning training sets across each iteration of our pipeline, visualized in two dimensions after performing t-distributed stochastic neighbor embedding (t-SNE). The generated molecules and active learning training sets are shown in (A) and (B), respectively. Changes in the generated molecules and active learning training sets relative to the molecules generated at iteration 0 are shown in (C) and (D), respectively. Iteration 0 refers to the pretraining phase, while later iterations refer to the active learning phases.

**Section 10:** Vocabulary Composition of the Combined Dataset.

**Table S10.1.** List of unique tokens that occur in the unfiltered combined dataset less than 1,000 times.

- '%10'
- '%11'
- '%12'
- '%13'
- '%14'
- '%15'
- '%16'
- '%17'
- '%18'
- '%19'
- '%20'
- '%21'
- '%22'
- '%23'
- '%24'
- '%25'
- '%26'
- '%27'
- '%28'
- '%29'
- '%30'
- '%31'
- '%32'
- '*'
- ':'
- '[*]'
- '[10B]'
- '[11C-]'
- '[11C@@H]'
- '[11CH2]'
- '[11CH3]'
- '[11CH]'
- '[11C]'
- '[11c]'
- '[123I-]'
- '[123IH]'
- '[123I]'
- '[123Te]'
- '[124I-]'
- '[124I]'
- '[125I-]'
- '[125IH]'
- '[125I]'
- '[127I]'
- '[127Xe]'
- '[129Xe]'
- '[131Cs]'
- '[131I-]'
- '[131I]'
- '[133Xe]'
- '[135I]'
- '[13CH2]'
- '[13CH3]'
- '[13CH]'
- '[13C]'
- '[13NH3]'
- '[13cH]'
- '[13c]'
- '[14C@@H]'
- '[14C@@]'
- '[14C@H]'
- '[14CH2]'
- '[14CH3]'
- '[14CH]'
- '[14C]'
- '[14cH]'
- '[14c]'
- '[15NH]'
- '[15OH2]'
- '[15nH]'
- '[15n]'
- '[17F]'
- '[18F-]'
- '[18FH]'
- '[18F]'
- '[18OH]'
- '[18O]'
- '[19F]'
- '[211At]'
- '[223Ra]'
- '[22Na+]'
- '[32PH]'
- '[32P]'
- '[35S]'
- '[3H]'
- '[42K+]'
- '[45Ca+2]'
- '[47Ca+2]'
- '[4H]'
- '[73Se]'
- '[75Se]'
- '[76BrH]'
- '[76Br]'
- '[81Kr]'
- '[82Rb+]'
- '[82Rb]'
- '[85Sr+2]'
- '[85SrH2]'
- '[89Sr+2]'
- '[Ag+]'
- '[Ag-4]'
- '[Ag-]'
- '[Ag]'
- '[Al+3]'
- '[Al-3]'
- '[Al]'
- '[Ar]'
- '[As+]'
- '[As-]'
- '[AsH3]'
- '[AsH]'
- '[As]'
- '[At]'
- '[Au-]'
- '[Au]'
- '[B@-]'
- '[B@@-]'
- '[BH-]'
- '[BH2-]'
- '[BH3-]'
- '[B]'
- '[Ba+2]'
- '[Ba]'
- '[Be+2]'
- '[Bi+3]'
- '[BiH3]'
- '[Bi]'
- '[Br+2]'
- '[Br]'
- '[C+]'
- '[CH+]'
- '[CH-]'
- '[CH2+]'
- '[CH2-]'
- '[CH2]'
- '[CH]'
- '[C]'
- '[Ca++]'
- '[Ca+2]'
- '[CaH2]'
- '[Ca]'
- '[Cl+2]'
- '[Cl+3]'
- '[Cl+]'
- '[Cl]'
- '[Co]'
- '[Cs+]'
- '[Cs]'
- '[Cu-]'
- '[Cu]'
- '[F+]'
- '[F-]'
- '[Fe++]'
- '[Fe--]'
- '[Fe-3]'
- '[Fe]'
- '[Gd-4]'
- '[Gd-5]'
- '[H+]'
- '[H-]'
- '[HH]'
- '[He]'
- '[Hg]'
- '[I+2]'

- '[I+3]'
- '[I+]'
- '[IH2]'
- '[IH]'
- '[I]'
- '[KH]'
- '[K]'
- '[Kr]'
- '[Li+]'
- '[LiH]'
- '[Li]'
- '[Mg+2]'
- '[Mg+]'
- '[MgH2]'
- '[Mg]'
- '[Mn]'
- '[Mo]'
- '[N@+]'
- '[N@@+]'
- '[N@@H+]'
- '[N@@]'
- '[N@H+]'
- '[N@]'
- '[NH-]'
- '[NH2+]'
- '[NH4+]'
- '[NH]'
- '[N]'
- '[NaH]'
- '[Na]'
- '[Nb--]'
- '[Ni++]'
- '[Ni]'
- '[O+]'
- '[O-2]'
- '[OH+]'
- '[OH-]'
- '[OH]'
- '[O]'
- '[Os]'
- '[P-]'
- '[P@+]'
- '[P@@+]'
- '[P@@]'
- '[P@]'
- '[PH+]'
- '[PH2+]'
- '[PH2]'
- '[PH]'
- '[P]'
- '[Pd--]'
- '[Pd]'
- '[Pt--]'
- '[Pt]'
- '[Ra]'
- '[Rb+]'
- '[Rb]'
- '[Re-]'
- '[Re]'
- '[Ru-]'
- '[Ru]'
- '[S-2]'
- '[S-]'
- '[S@+]'
- '[S@@+]'
- '[S@]'
- '[SH+]'
- '[SH-]'
- '[SH2]'
- '[SH]'
- '[S]'
- '[Sb]'
- '[Se+]'
- '[Se-2]'
- '[Se-]'
- '[SeH2]'
- '[SeH]'
- '[Si-]'
- '[Si@]'
- '[SiH-]'
- '[SiH2]'
- '[SiH3-]'
- '[SiH3]'
- '[SiH4]'
- '[SiH]'
- '[Sn]'
- '[Sr++]'
- '[Sr+2]'
- '[SrH2]'
- '[Tc]'
- '[Te+]'
- '[Te-]'
- '[TeH2]'
- '[TeH]'
- '[Te]'
- '[V]'
- '[W]'
- '[Xe]'
- '[Zn++]'
- '[Zn+2]'
- '[Zn+]'
- '[Zn-2]'
- '[Zn]'
- '[b-]'
- '[c+]'
- '[c-]'
- '[cH+]'
- '[cH-]'
- '[c]'
- '[n-]'
- '[nH+]'
- '[n]'
- '[o+]'
- '[o]'
- '[s+]'
- '[s]'
- '[se+]'
- '[te+]'
- '[te]'
- 'b'
- 'p'

**Table S10.2.** List of unique tokens that occur in the filtered combined dataset.

- '!'
- '#'
- '('
- ') '
- '-'
- '.'
- '/'
- '1'
- '2'
- '3'
- '4'
- '5'
- '6'
- '7'
- '8'
- '9'
- '<'
- '='
- 'B'
- 'Br'
- 'C'
- 'Cl'
- 'F'
- 'I'
- 'N'
- 'O'
- 'P'
- 'S'
- '[2H]'
- '[B-]'
- '[Br-]'
- '[C-]'
- '[C@@H]'
- '[C@@]'
- '[C@H]'
- '[C@]'
- '[Cl-]'
- '[H]'
- '[I-]'
- '[K+]'
- '[N+]'
- '[N-]'
- '[NH+]'
- '[NH3+]'
- '[Na+]'
- '[O-]'
- '[P+]'
- '[S+]'
- '[S@@]'
- '[Se]'
- '[Si]'
- '[n+]'
- '[nH]'
- '[se]'
- '\'
- 'c'
- 'n'
- 'o'
- 's'
- '~'

**Section 11:** Frequencies of Block Sizes, Molecular Weights and Tokens in Pretraining Sets.

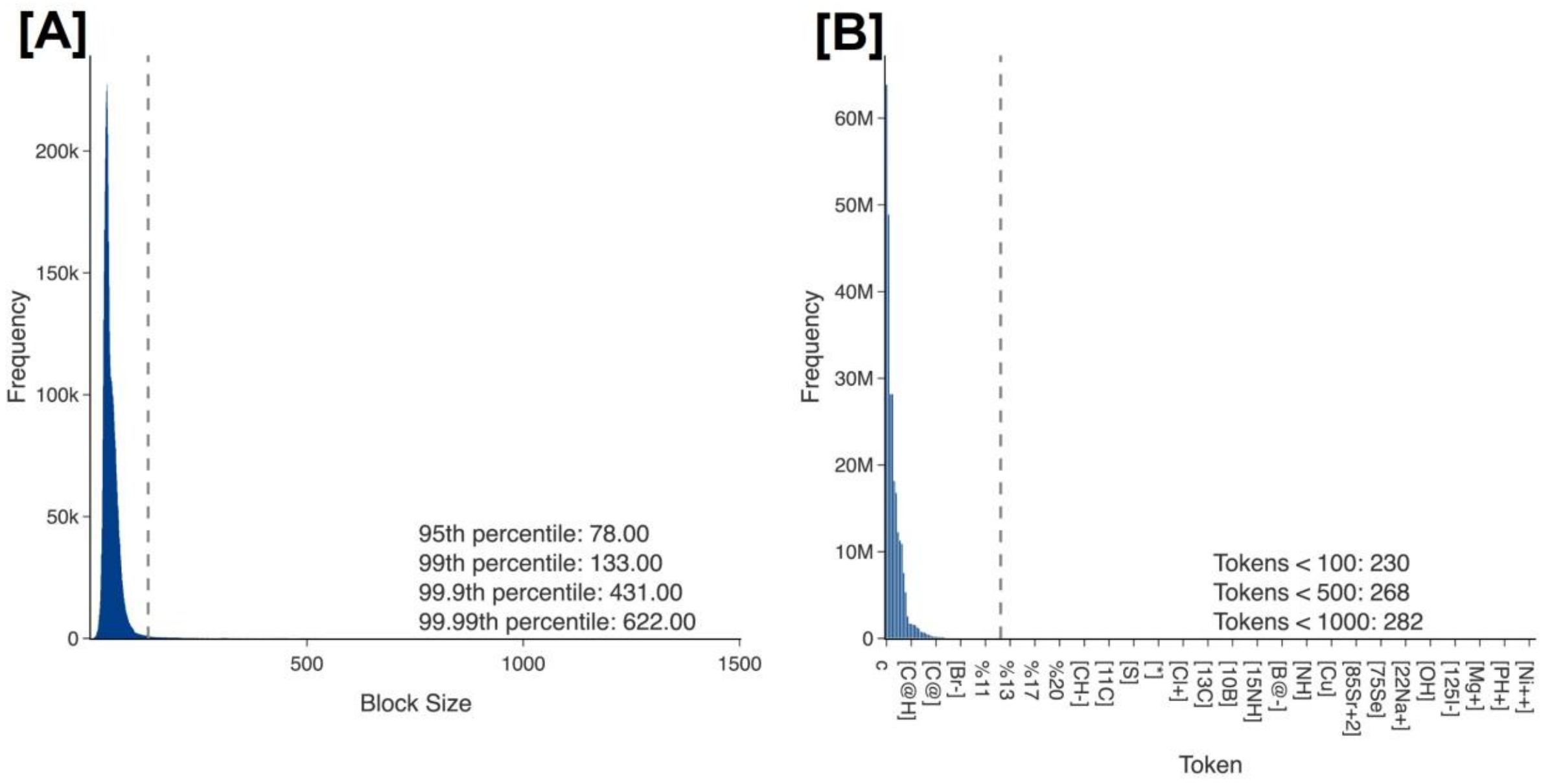

**Figure S11.1.** Frequency as a function of block size (A) and token (B). Vertical dotted lines are positioned at 133 in (A) and serves as our block size cutoff. In (B), our cutoff, illustrated with the vertical dotted lines, is positioned at the first token where the frequency is less than 1,000 times.

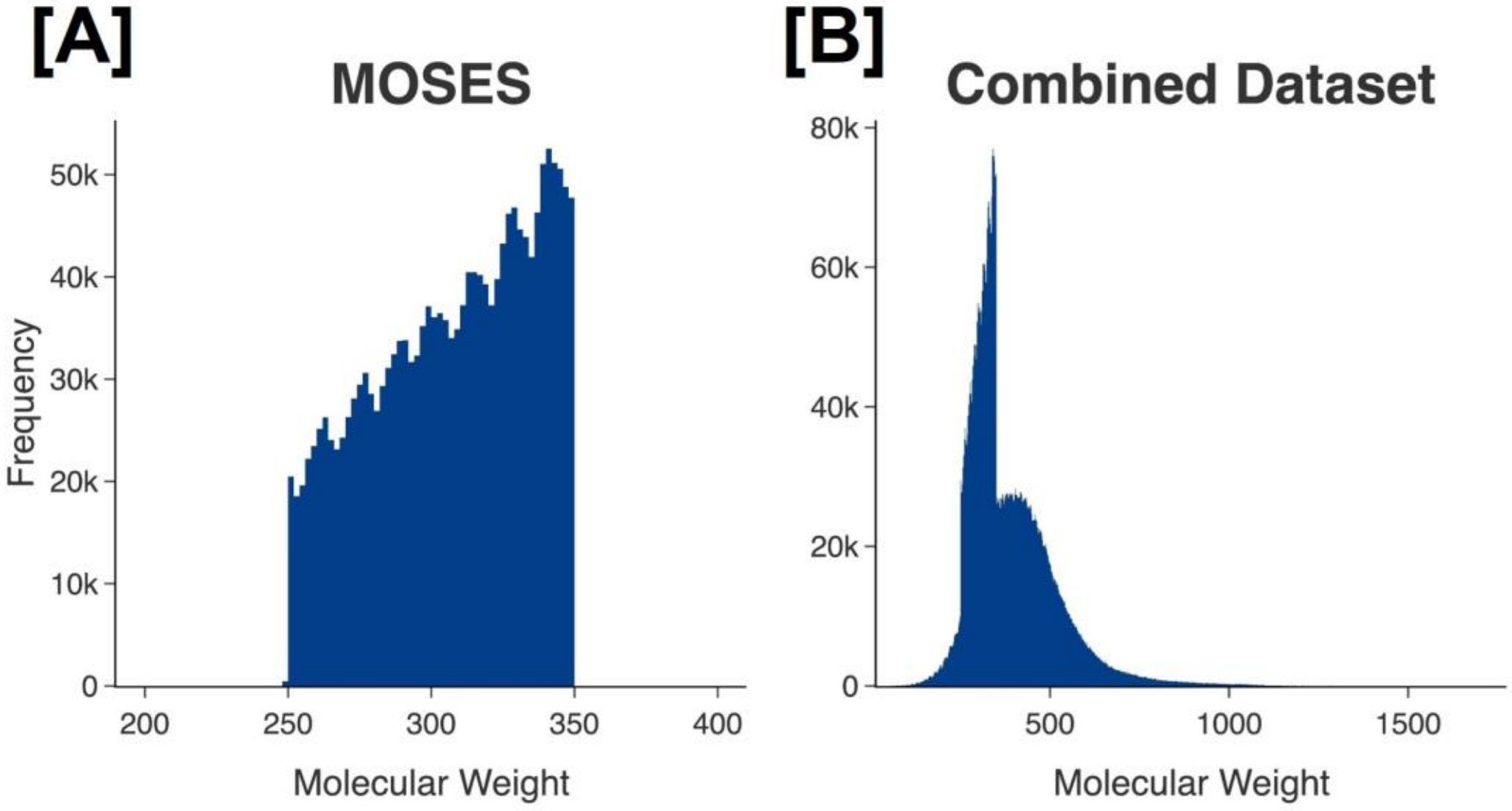

**Figure S11.2.** Frequency as a function of molecular weight of molecule in the (A) MOSES and (B) combined datasets.

**<u>Section 12</u>:** Details of the GPT Architecture.

The GPT model that we employ is based on the transformer architecture introduced in the revolutionary paper, "Attention is All You Need".**Error! Bookmark not defined.** Within the framework of the transformer architecture, the encoder processes input data into a sequence of context-rich vectors, while the decoder utilizes this contextual information to generate output data. Both of these
components utilize a self-attention mechanism, which enables the model to selectively focus on distinct parts of the input sequence at each computational step. The technical difference between the encoder and decoder parts of the transformer model is that the decoder ensures that the prediction for a particular token only depends on the preceding tokens, while each token in the encoder can attend to all other tokens in the sequence. Our GPT model is constructed as a series of transformer decoder blocks. This approach is appropriate for tasks that require generating novel sequential data such as SMILES strings.

The forward pass of our GPT model begins by dividing each SMILES string into distinct units known as tokens, processing each token with embedding layers, and combining these embeddings to form a vector representation of each token. These embedded vectors are then sequentially passed through a series of transformer decoder blocks, each comprised of a self-attention layer and a feed-forward network, with additional structural elements to enhance learning. The final result is a sequence of vectors, each corresponding to a position in the output SMILES string, where the elements of each vector represent probabilities for each token in the vocabulary. This high-level overview sets the stage for a more detailed discussion of the individual components.

*Embeddings*: Initially, a vocabulary comprising all of the unique tokens in the training data is constructed. For any given SMILES string in the input data, the input tokens undergo three distinct processing methods: token, positional, and type embeddings. The token embedding maps each token in the input sequence to a learnable vector representation, allowing the model to learn an optimal high-dimensional characterization for each token. Similarly, the positional embedding maps each input token to a learnable vector based on its position in the sequence. The type embedding layer uniformly assigns a constant bias to all embeddings of each input sequence. The sum of these three embeddings is passed through a dropout layer, setting 10% of its scalar components to 0. This embedding process transforms the input tokens into a form more suitable for the downstream modeling process.

*Transformer Decoder Stack*: For each token in the input sequence, the resulting embedding is passed to the first transformer decoder block, which begins with layer normalization, a process that adjusts and scales each embedding to have a mean of 0 and a standard deviation of 1. A self-attention mechanism is then applied to the normalized embedding, using learned matrices to linearly transform the embedding into three different vectors known as the query, key, and value vectors:

$$\mathbf{q}_i = \mathbf{W}_q \times \mathbf{e}_i \quad (1)$$

$$\mathbf{k}_i = \mathbf{W}_k \times \mathbf{e}_i \quad (2)$$

$$\mathbf{v}_i = \mathbf{W}_v \times \mathbf{e}_i \quad (3)$$

where $\mathbf{W}_q$, $\mathbf{W}_k$, and $\mathbf{W}_v$ are learned weight matrices that transform each input embedding, represented by $\mathbf{e}_i$, into the corresponding query, key, and value vectors. The dot products of the query and each key vector are then scaled according to the dimensionality of the key vectors and passed through a softmax function, transforming them into a probability distribution to serve as attention weights. Finally, the attention scores are used to generate a weighted sum of the value vectors, as shown in the following equation:

$$\mathbf{e}'_i = \mathbf{V} \times \text{softmax}\begin{pmatrix} \frac{\mathbf{q}_i \cdot \mathbf{k}_1}{\sqrt{\mathrm{d}_k}} \\ \cdots \\ \frac{\mathbf{q}_i \cdot \mathbf{k}_L}{\sqrt{\mathrm{d}_k}} \end{pmatrix} \tag{4}$$

Here, $\mathbf{e}'_i$ represents the output of the attention mechanism at position $i$ in the sequence, $\mathbf{V}$ is the value matrix whose $j^{\text{th}}$ column is the value vector corresponding to the embedding at position $j$ in the sequence, $\mathrm{d}_k$ denotes the dimensionality of the key vectors, and $L$ represents the length of the entire sequence. This operation amplifies the information from value vectors corresponding to higher attention weights (i.e., tokens that are more relevant to the current query), while suppressing the information from less relevant value vectors.

In practice, the self-attention mechanism is executed multiple times in parallel through what is known as *multi-head* attention. Each head (i.e., execution) uses its own set of learned linear transformations to generate query, key, and value vectors for all tokens in the sequence for each item in the batch, allowing the model to simultaneously focus on different aspects of the input across the various heads. The outputs from all attention heads are then concatenated and passed through a learned linear transformation to generate the final output of the multi-head attention mechanism.

A residual connection is a shortcut that skips one or more layers and allows the original input to be added directly to the output of those layers. This technique aids in training deeper networks by mitigating the vanishing gradient problem, where the gradients become too small for the network to learn effectively. In the context of GPT models, a residual connection is made by adding the input of the attention mechanism to the output. This sum is then processed using layer normalization, and the transformed embeddings are passed through a feed-forward network using the equation:

$$\mathbf{H} = \text{Dropout}(\mathbf{W_2} \times \text{GELU}(\mathbf{W_1} \times \mathbf{E}' + \mathbf{b_1}) + \mathbf{b_2}) \tag{5}$$

where $\mathbf{H}$ is the output of the feed-forward network, $\mathbf{E}'$ represents the matrix whose columns are the transformed embeddings, and $\mathbf{W}_1$ (shape: 1024×256), $\mathbf{b}_1$ (shape: 1024), $\mathbf{W}_2$ (shape: 256×1024), and $\mathbf{b}_2$ (shape: 256) represent the weight matrices and bias vectors of the two linear layers. GELU, or Gaussian Error Linear Unit, is an activation function used to introduce non-linearity into the model. A residual connection is established by summing the input to this feed-forward network with the output.

This entire process is repeated for additional decoder blocks, and the output of the final decoder block is processed with layer normalization. The normalized output is then passed through a learned linear transformation with bias to map the embeddings to the output vocabulary size, and the resulting vectors are processed with softmax to generate the output probabilities at each position in the sequence.

**<u>Section 13</u>:** Training the GPT Model.

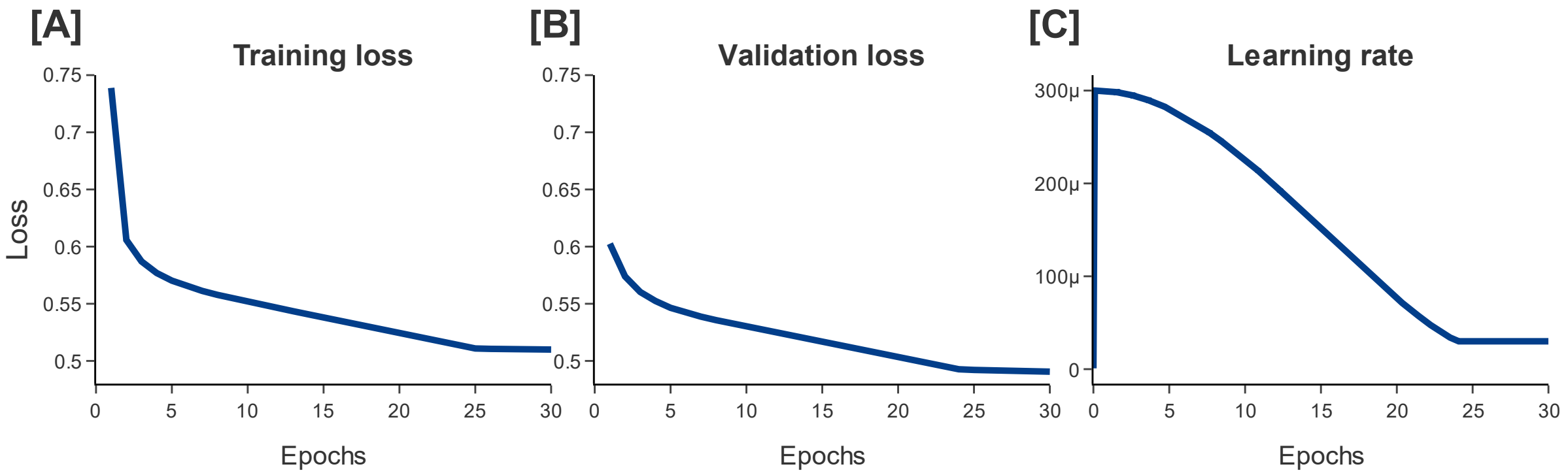

**Figure S13.1.** Training loss (A), validation loss (B), and learning rate (C) during the 30 epochs of pretraining of our model on the combined dataset.

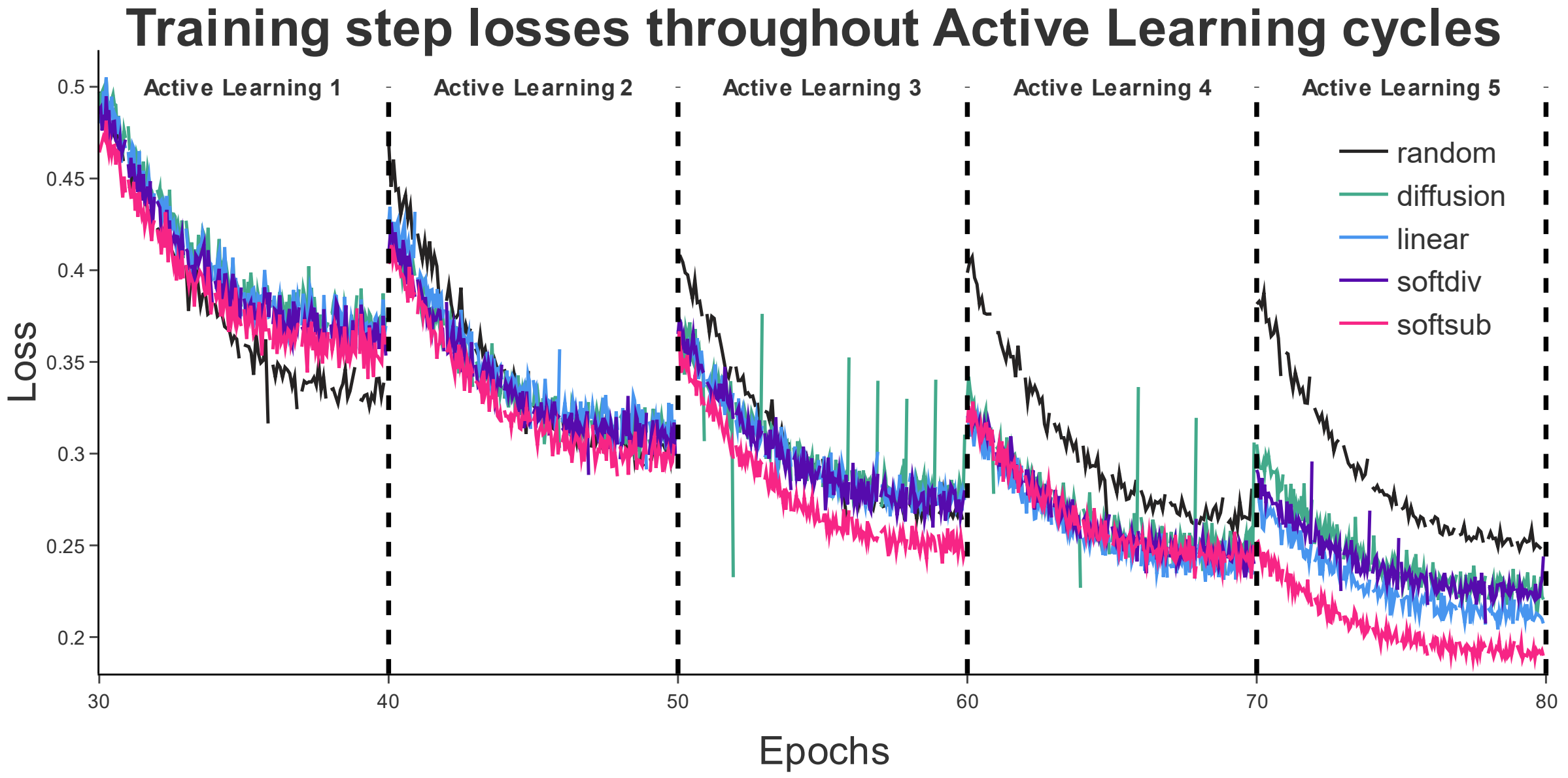

**Figure S13.2.** Training step losses (evaluated after each batch) during 5 rounds of active learning for the HNH domain of Cas9 (10 epochs each) with different conversion methods.

**Section 14:** Pretrained GPT Model Performance on the MOSES Benchmark.

**Table S14.1.** Primary results of our pretrained model on the MOSES benchmark compared to top-performing models in the field.

| Model | Validity | Unique@1K | Unique@10K | Novelty | IntDiv$_1$ | IntDiv$_2$ | Filters |
|---|---|---|---|---|---|---|---|
| ***Our Model*** | 0.996 | **1.000** | 0.999 | 0.730 | 0.856 | 0.850 | **0.998** |
| MolGPT[2] | 0.994 | N/A | **1.000** | 0.797 | 0.857 | 0.851 | N/A |
| LatentGAN[3] | 0.897 | **1.000** | 0.997 | 0.949 | 0.857 | 0.850 | 0.973 |
| JT-VAE[4] | **1.000** | **1.000** | **1.000** | 0.914 | 0.855 | 0.849 | 0.976 |
| CharRNN[5] | 0.975 | **1.000** | 0.999 | 0.842 | 0.856 | 0.850 | 0.994 |
| MolecularRNN[6] | **1.000** | N/A | 0.994 | **1.000** | 0.881 | **0.876** | N/A |
| iPPIgAN[7] | 0.989 | **1.000** | 0.999 | 0.990 | N/A | N/A | N/A |
| DNMG[8] | 0.999 | **1.000** | 0.998 | 0.936 | 0.856 | 0.850 | 0.996 |
| CogMol[9] | 0.955 | **1.000** | **1.000** | N/A | 0.857 | 0.851 | 0.989 |
| TransVAE[10] | 0.567 | NA | N/A | 0.996 | N/A | N/A | N/A |
| ShapeProb[11] | 0.969 | **1.000** | 0.995 | N/A | 0.865 | N/A | 0.865 |
| GENTRL[12] | 0.850 | N/A | N/A | N/A | N/A | N/A | N/A |
| TransAntivirus[13] | **1.000** | 0.999 | 0.999 | 0.999 | **0.895** | N/A | N/A |
| CRTmaccs[14] | **1.000** | **1.000** | **1.000** | **1.000** | N/A | N/A | N/A |
| MolGCT[15] | 0.985 | **1.000** | 0.998 | 0.814 | 0.853 | N/A | 0.996 |
| cMolGPT[16] | 0.988 | **1.000** | 0.999 | N/A | N/A | N/A | N/A |
| GraphINVENT[17] | 0.964 | **1.000** | 0.998 | N/A | 0.857 | 0.851 | 0.950 |
| cTransformer[18] | 0.988 | **1.000** | 0.999 | N/A | N/A | N/A | N/A |
| GMTransformer[19] | 0.829 | **1.000** | **1.000** | 0.883 | 0.856 | N/A | 0.980 |

[a] Validity (ratio of generated molecules deemed valid by RDKit's molecular structure parser), Unique@1K and @10K (fraction of valid generated molecules with no duplicates), Novelty (fraction of valid and unique generated molecules that are not in the training set), IntDiv$_i$ (internal diversity within the generated set for power mean $i$), and Filters (fraction of generated molecules that pass filters that check for specific fragments) are shown. See MOSES benchmark for more details on how these metrics are calculated.[20]

[b] The top value for each metric is shown in bold.

[c] Values not reported are shown as N/A.

**Table S14.2.** Additional results of our pretrained model on the MOSES benchmark compared to top-performing models in the field.

| Model | FCD/Test | FCD/TestSF | Frag/Test | Frag/TestSF | SNN/Test | SNN/TestSF | Scaff/Test | Scaff/ |
|---|---|---|---|---|---|---|---|---|
| ***Our Model*** | **0.038** | **0.450** | **1.000** | **0.999** | 0.633 | **0.585** | **0.970** | 0. |
| MolGPT[2] | 0.067 | 0.507 | N/A | N/A | N/A | N/A | N/A | N |
| LatentGAN[3] | 0.296 | 0.824 | 0.999 | 0.998 | 0.538 | 0.514 | 0.886 | 0. |
| JT-VAE[4] | 0.395 | 0.938 | 0.997 | 0.995 | 0.548 | 0.519 | 0.896 | 0. |
| CharRNN[5] | 0.073 | 0.520 | **1.000** | 0.998 | 0.601 | 0.565 | 0.924 | 0. |
| MolecularRNN[6] | N/A | N/A | N/A | N/A | N/A | N/A | N/A | N |
| iPPIgAN[7] | 5.879 | 6.171 | N/A | N/A | N/A | N/A | N/A | N |
| DNMG[8] | 0.373 | 0.631 | 0.999 | 0.998 | 0.472 | 0.579 | 0.784 | **0.** |
| CogMol[9] | 0.166 | 0.603 | 0.999 | 0.997 | 0.560 | 0. 533 | 0.905 | 0. |
| TransVAE[10] | N/A | N/A | N/A | N/A | N/A | N/A | N/A | N |
| ShapeProb[11] | 1.332 | 1.850 | 0.984 | 0.980 | 0.446 | 0.432 | 0.459 | 0. |
| GENTRL[12] | N/A | N/A | N/A | N/A | N/A | N/A | N/A | N |
| TransAntivirus[13] | 10.947 | N/A | N/A | N/A | N/A | N/A | N/A | N |
| CRTmaccs[14] | 13.565 | 13.999 | N/A | N/A | 0.334 | 0.330 | N/A | N |
| MolGCT[15] | 0.402 | 0.803 | 0.997 | 0.995 | 0.618 | 0.577 | 0.891 | 0. |
| cMolGPT[16] | N/A | N/A | **1.000** | 0.998 | 0.619 | 0.578 | N/A | N |
| GraphINVENT[17] | 0.682 | 1.223 | 0.986 | 0.986 | 0.569 | 0.539 | 0.885 | 0. |
| cTransformer[18] | N/A | N/A | **1.000** | 0.998 | 0.619 | 0.578 | N/A | N |
| GMTransformer[19] | 0.199 | 0.760 | 0.998 | 0.996 | 0.578 | 0.546 | 0.913 | 0. |

[a] FCD (Fréchet ChemNet Distance that is calculated using activation of the penultimate layer of ChemNet), Frag (compares molecular fragments between generated and training sets), SNN (average Tanimoto similarity between molecules in the generated set and the corresponding nearest molecule in the training set), and Scaff (compares molecular scaffolds between generated and training sets) are shown. Test (similarity from the training set to the test set) and TestSF (similarity from the training set to the scaffold test set) are shown for each metric. See MOSES benchmark for more details on how these metrics are calculated.[20]

[b] The top value for each metric is shown in bold.

[c] Values not reported are shown as N/A.

**Section 15:** RDKit Descriptors Used to Construct the Chemical Space Proxy.

**Table S15.1.** List of RDKit descriptors discarded.

- BCUT2D_CHGHI
- BCUT2D_CHGLO
- BCUT2D_LOGPHI
- BCUT2D_LOGPLOW
- BCUT2D_MRHI
- BCUT2D_MRLOW
- BCUT2D_MWHI
- BCUT2D_MWLOW
- Ipc
- MaxAbsPartialCharge
- MaxPartialCharge
- MinAbsPartialCharge
- MinPartialCharge

**Table S15.2.** List of RDKit descriptors included and used to construct the chemical space proxy.

- AvgIpc
- BalabanJ
- BertzCT
- Chi0
- Chi0n
- Chi0v
- Chi1
- Chi1n
- Chi1v
- Chi2n
- Chi2v
- Chi3n
- Chi3v
- Chi4n
- Chi4v
- EState_VSA1
- EState_VSA10
- EState_VSA11
- EState_VSA2
- EState_VSA3
- EState_VSA4
- EState_VSA5
- EState_VSA6
- EState_VSA7
- EState_VSA8
- EState_VSA9
- ExactMolWt
- FpDensityMorgan1
- FpDensityMorgan2
- FpDensityMorgan3
- FractionCSP3
- HallKierAlpha
- HeavyAtomCount
- HeavyAtomMolWt
- Kappa1
- Kappa2
- Kappa3
- LabuteASA
- MaxAbsEState Index
- MaxEStateIndex
- MinAbsEState Index
- MinEStateIndex
- MolLogP
- MolMR
- MolWt
- NHOHCount
- NOCount
- NumAliphatic Carbocycles
- NumAliphatic Heterocycles
- NumAliphatic Rings
- NumAromatic Carbocycles
- NumAromatic Heterocycles
- NumAromatic Rings
- NumHAcceptors
- NumHDonors
- NumHeteroatoms
- NumRadical Electrons
- NumRotatable Bonds
- NumSaturated Carbocycles
- NumSaturated Heterocycles
- NumSaturated Rings
- NumValence Electrons
- PEOE_VSA1
- PEOE_VSA10
- PEOE_VSA11
- PEOE_VSA12
- PEOE_VSA13
- PEOE_VSA14
- PEOE_VSA2
- PEOE_VSA3
- PEOE_VSA4
- PEOE_VSA5
- PEOE_VSA6
- PEOE_VSA7
- PEOE_VSA8
- PEOE_VSA9
- RingCount
- SMR_VSA1
- SMR_VSA10

- SMR_VSA2
- SMR_VSA3
- SMR_VSA4
- SMR_VSA5
- SMR_VSA6
- SMR_VSA7
- SMR_VSA8
- SMR_VSA9
- SlogP_VSA1
- SlogP_VSA10
- SlogP_VSA11
- SlogP_VSA12
- SlogP_VSA2
- SlogP_VSA3
- SlogP_VSA4
- SlogP_VSA5
- SlogP_VSA6
- SlogP_VSA7
- SlogP_VSA8
- SlogP_VSA9
- TPSA
- VSA_EState1
- VSA_EState10
- VSA_EState2
- VSA_EState3
- VSA_EState4
- VSA_EState5
- VSA_EState6
- VSA_EState7
- VSA_EState8
- VSA_EState9
- fr_Al_COO
- fr_Al_OH
- fr_Al_OH_noTert
- fr_ArN
- fr_Ar_COO
- fr_Ar_N
- fr_Ar_NH
- fr_Ar_OH
- fr_COO
- fr_COO2
- fr_C_O
- fr_C_O_noCOO
- fr_C_S
- fr_HOCCN
- fr_Imine
- fr_NH0
- fr_NH1
- fr_NH2
- fr_N_O
- fr_Ndealkylation1
- fr_Ndealkylation2
- fr_Nhpyrrole
- fr_SH
- fr_aldehyde
- fr_alkyl_carbamate
- fr_alkyl_halide
- fr_allylic_oxid
- fr_amide
- fr_amidine
- fr_aniline
- fr_aryl_methyl
- fr_azide
- fr_azo
- fr_barbitur
- fr_benzene
- fr_benzodiazepine
- fr_bicyclic
- fr_diazo
- fr_dihydropyridine
- fr_epoxide
- fr_ester
- fr_ether
- fr_furan
- fr_guanido
- fr_halogen
- fr_hdrzine
- fr_hdrzone
- fr_imidazole
- fr_imide
- fr_isocyan
- fr_isothiocyan
- fr_ketone
- fr_ketone_Topliss
- fr_lactam
- fr_lactone
- fr_methoxy
- fr_morpholine
- fr_nitrile
- fr_nitro
- fr_nitro_arom
- fr_nitro_arom_ nonortho
- fr_nitroso
- fr_oxazole
- fr_oxime
- fr_para_ hydroxylation
- fr_phenol
- fr_phenol_ noOrthoHbond
- fr_phos_acid
- fr_phos_ester
- fr_piperdine
- fr_piperzine
- fr_priamide
- fr_prisulfonamd
- fr_pyridine
- fr_quatN
- fr_sulfide
- fr_sulfonamd
- fr_sulfone
- fr_term_acetylene
- fr_tetrazole
- fr_thiazole
- fr_thiocyan
- fr_thiophene
- fr_unbrch_alkane
- fr_urea
- qed

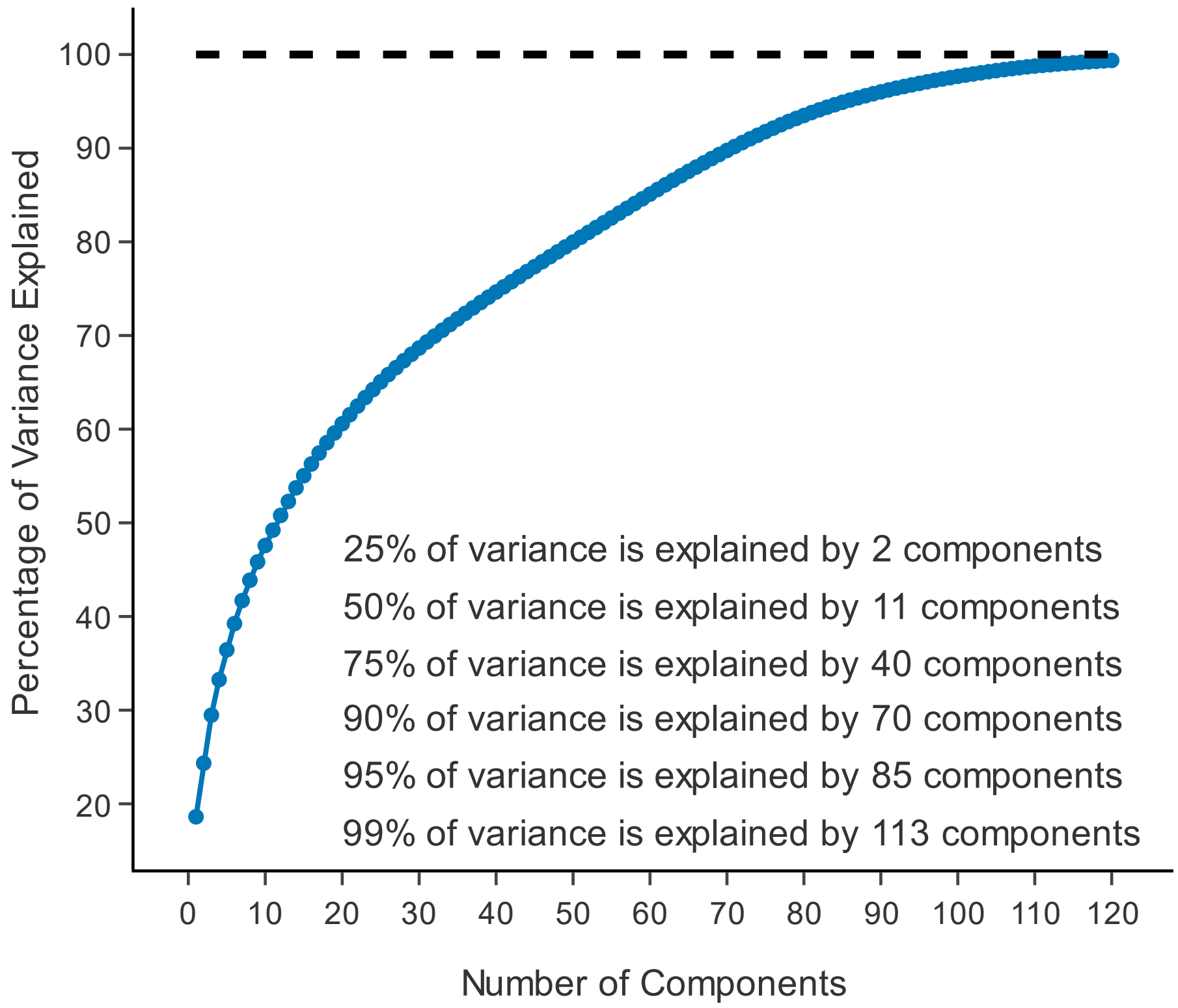

**Figure S15.3.** Cumulative fraction of variance explained by the first *N* principal components. Our chemical space proxy (i.e., the first 120 principal components) explains 99.3% of the variance in the hyperspace of 196 RDKit descriptors.

**<u>Section 16</u>:** Wall Times of Each Step in the Complete Pipeline.

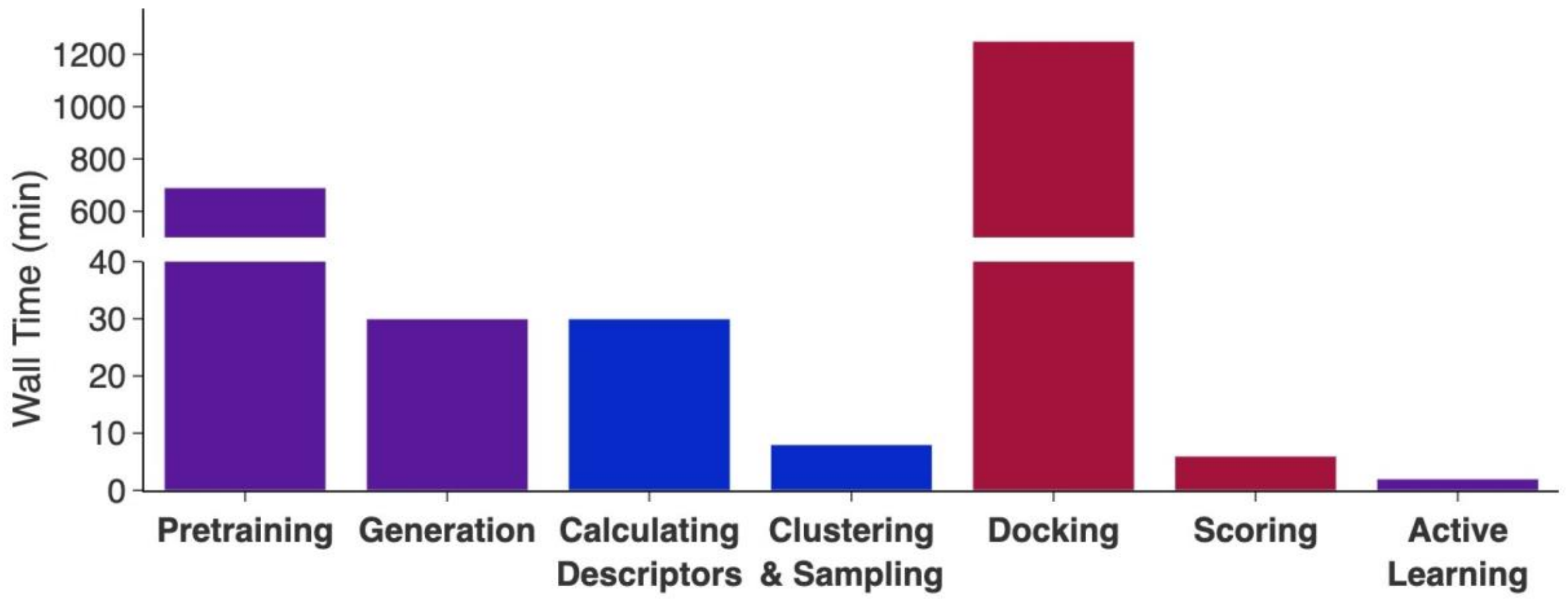

**Figure S16.1.** Wall times of each step in the complete pipeline. Steps include pretraining, generation, calculating descriptors, clustering and sampling, docking, scoring, and active learning fine-tuning. Pretraining is performed only once, while each other step is performed once per iteration.

**Section 17:** Evaluating the Methodology with Lower-Dimensional MQN Filters.

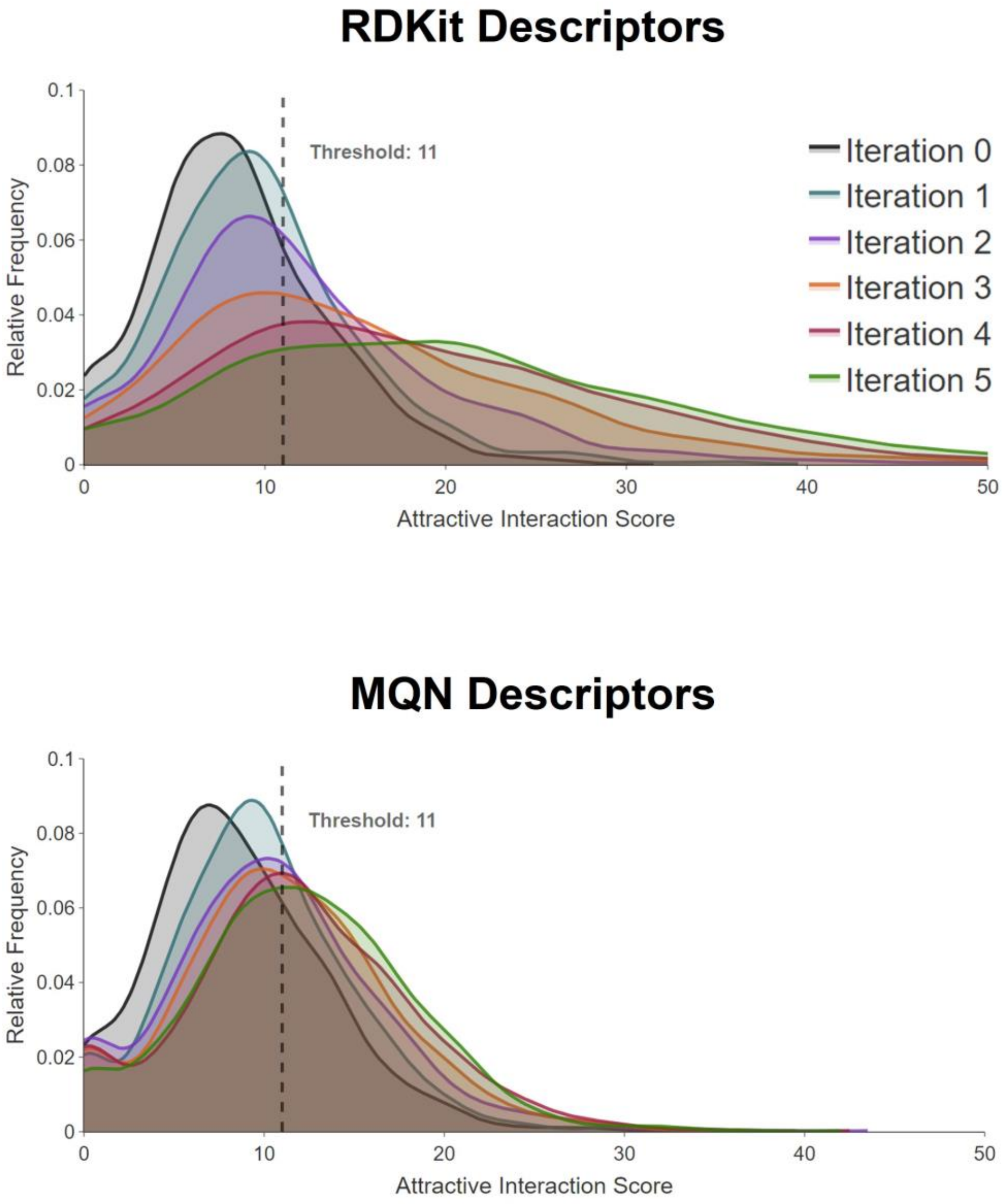

**Figure S17.1**. Attractive interaction scores of scored molecules across five iterations of active learning. Results for the methodology applied to the HNH domain of Cas9 using 196 RDKit descriptors, and using 42 MQN descriptors are shown. Iteration 0 refers to the pretraining phase, while later iterations refer to the active learning phases.

**Section 18:** Frequency as a Function of Cluster Size for Alignment to c-Abl Kinase.

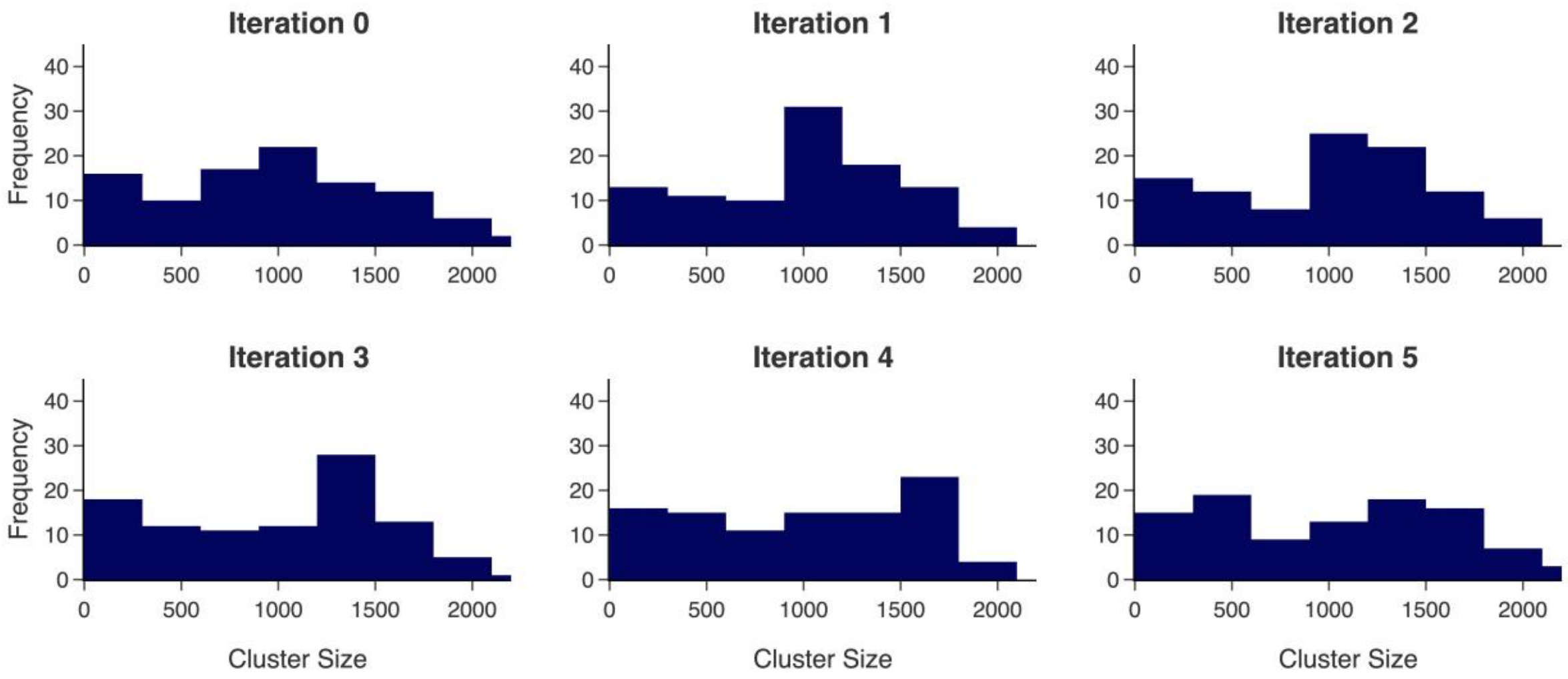

**Figure S18.1.** Frequency as a function of cluster size for each iteration of the methodology. Results are shown for the model pretrained on the combined dataset with the generations filtered based on ADMET metrics, aligned to c-Abl kinase.

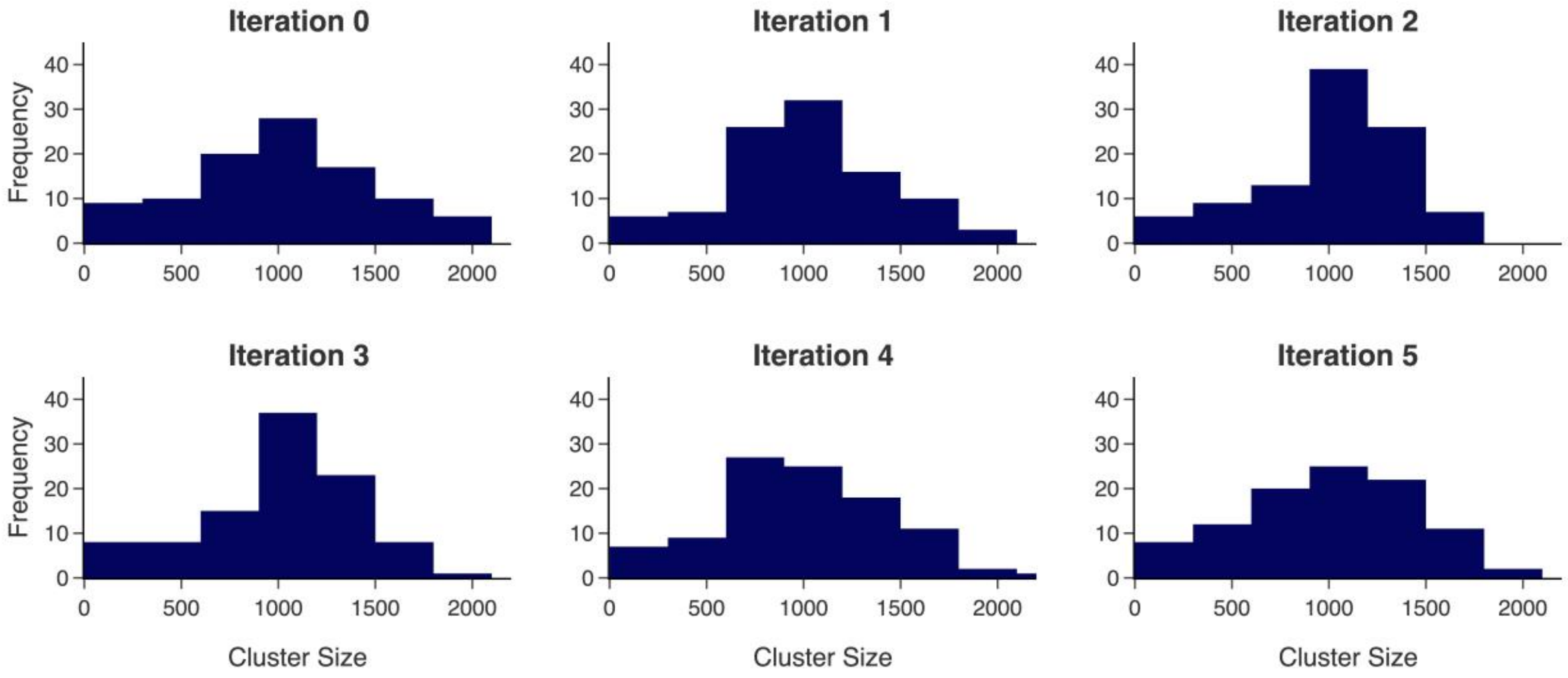

**Figure S18.2.** Frequency as a function of cluster size for each iteration of the methodology. Results are shown for the model pretrained on the combined dataset with the generations filtered based on ADMET metrics and functional group restrictions, aligned to c-Abl kinase.

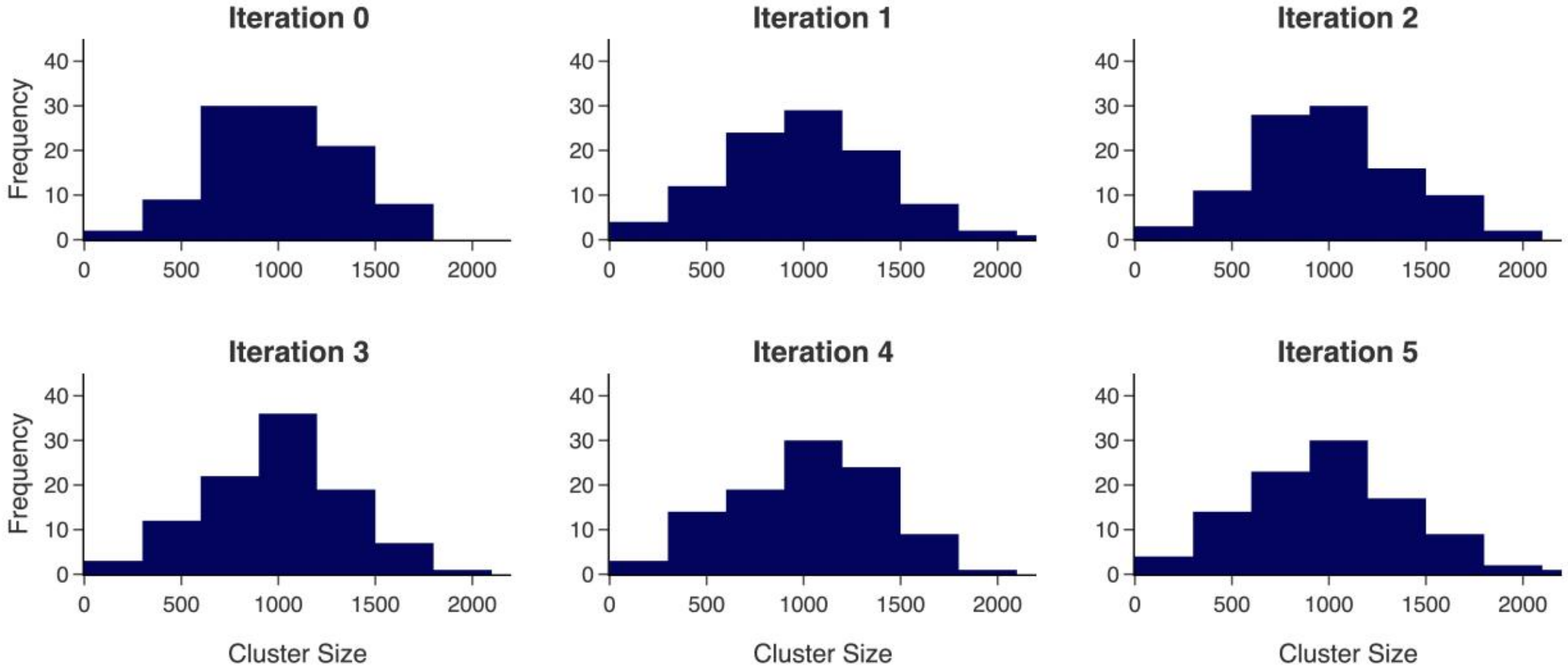

**Figure S18.3.** Frequency as a function of cluster size for each iteration of the methodology. Results are shown for the model pretrained on the MOSES dataset with the generations filtered based on ADMET metrics and functional group restrictions, aligned to c-Abl kinase.

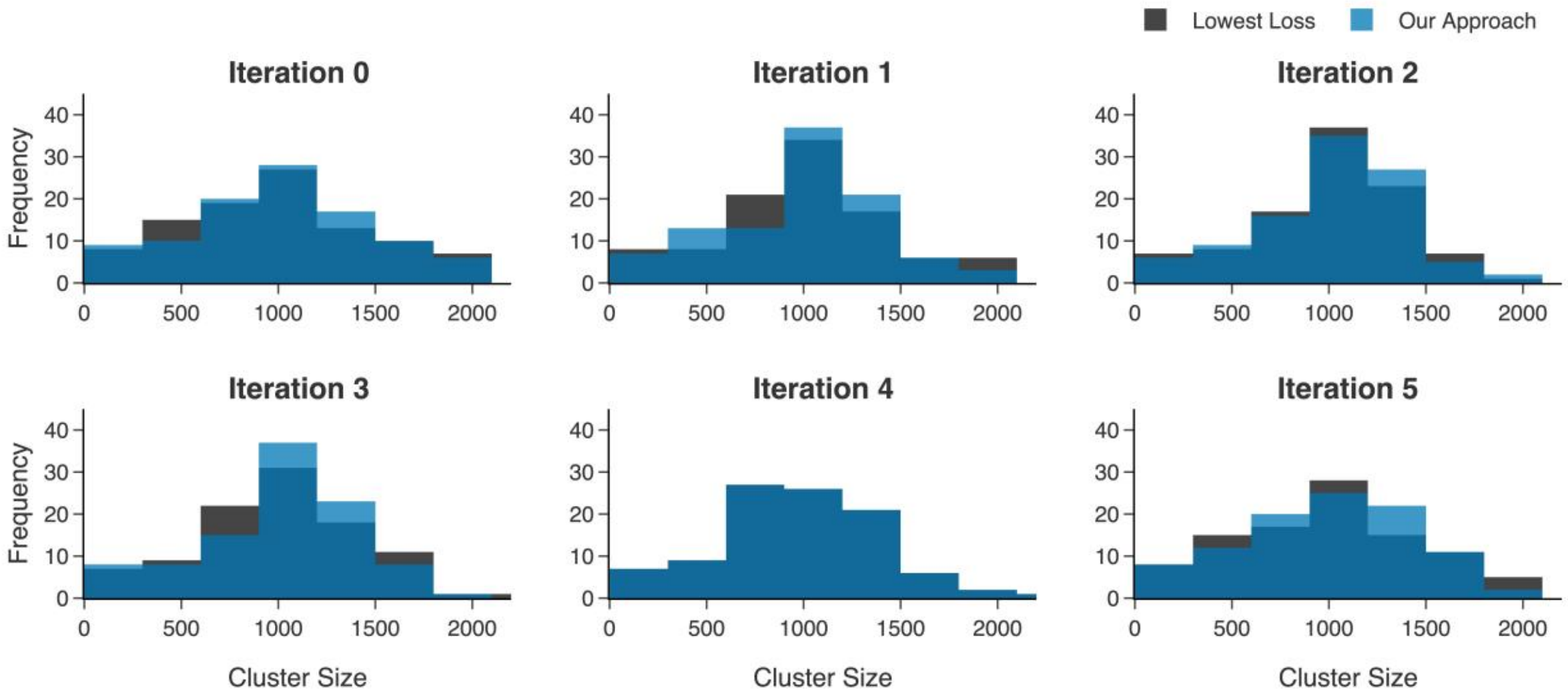

**Figure S18.4.** Frequency as a function of cluster size for each iteration of the methodology shown for the clustering that we select as well as the clustering with the lowest loss. Results are shown for the model pretrained on the combined dataset with the generations filtered based on ADMET metrics and functional group restrictions, aligned to c-Abl kinase.

**Section 19:** Evaluation of Scoring Function Compared to PDBbind v2020 Refined Set.

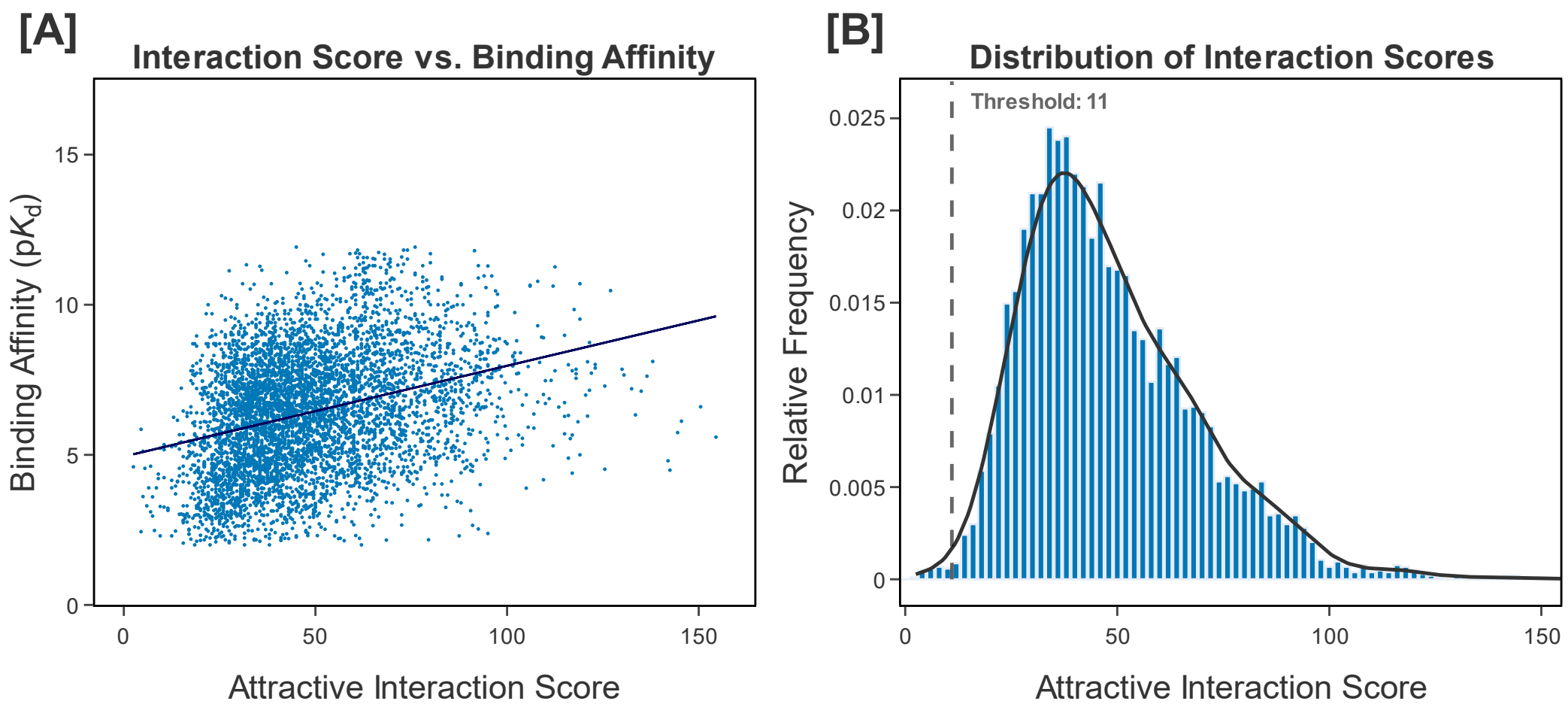

**Figure S19.1.** Evaluation of our scoring function with the protein-ligand complexes in the PDBbind v.2020 refined set. **(A)** Binding affinity (p$K_d$) plotted as a function of score. There is a corresponding Pearson correlation of 0.32. **(B)** The relative frequency of different scores is shown. 99.6% of the complexes exceed our score threshold of 11.

**Section 20:** Alternative Methods for Converting Mean Cluster Scores to Sampling Fractions.

Prior to constructing the active learning training set, we need to convert the attractive interaction scores $s_i$ obtained by using the prolif software on docked molecules into sampling fractions $f_i$, which will be used to calculate the number of molecules that we need to sample from each cluster. A simple way to do that is to normalize the sum of all scores to unity:

$$f_i^{\text{linear}}(s_i) = \frac{s_i}{\sum_i s_i}$$

We call this approach *linear* conversion. Because one could interpret sampling fractions as effective probabilities of sampling from a given cluster, it is natural to consider the use of a softmax function:

$$f_i^{\text{softmax}}(s_i) = \frac{e^{s_i}}{\sum_i e^{s_i}}$$

which, for computational stability purposes, is often implemented with the maximum value among a set of arguments subtracted from each individual argument. To contrast with a modification of a softmax function introduced later, we refer to this as *softsub* conversion. In the main text of our paper, we implement the *softsub* approach and refer to it as *softmax* because this is the common implementation of the softmax function.

$$f_i^{\text{softsub}}(s_i) = \frac{e^{s_i - s_{max}}}{\sum_i e^{s_i - s_{max}}}$$

For a pretrained model, cluster scores range from 0 to 16. Because exponential functions increase rapidly, the *softsub* approach will effectively favor the 1-5 clusters with largest scores. We conjecture that a smoother function may lead to better model behavior during active learning, and implement a *softdiv* conversion approach, in which, instead of subtracting the maximum cluster score, we divide by it:

$$f_i^{\text{softdiv}}(s_i) = \frac{e^{s_i / s_{max}}}{\sum_i e^{s_i / s_{max}}}$$

Empirically, this approach leads even to a narrower distribution of sampling fractions than that obtained with the *linear* conversion approach. We introduce a hyperparameter $divf \in (0,1]$ by which we multiply the $s_{max}$ value prior to dividing by it:

$$f_i^{\text{softdivf}}(s_i) = \frac{e^{\frac{s_i}{divf \times s_{max}}}}{\sum_i e^{\frac{s_i}{divf \times s_{max}}}}$$

By visualizing the distribution of *softdiv* values with different values of the hyperparameter (Figure S4.1), we pick $divf = 0.25$, as it maximizes the spread in sampling fractions. In what follows, the *softdiv* conversion will refer to *softdiv* with $divf = 0.25$.

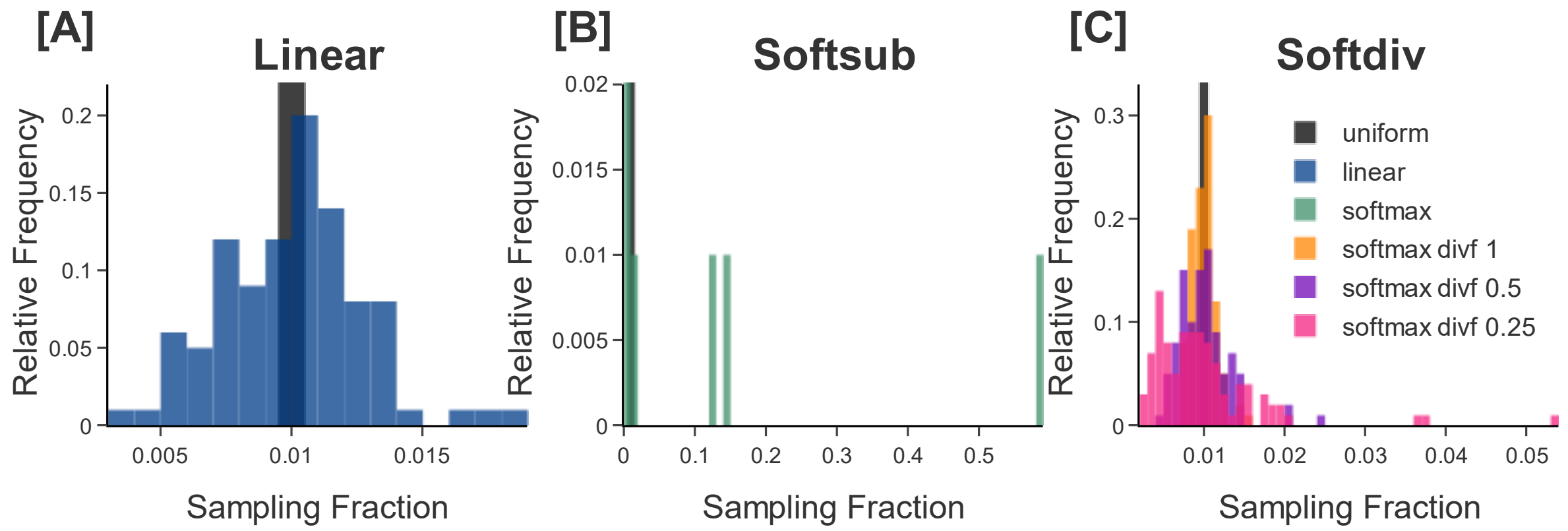

**Figure S20.1.** Distribution of sampling fractions obtained with different conversion approaches applied to cluster scores obtained from generations of the pretrained model. A bar corresponding to sampling the same number of molecules from each cluster (i.e., uniform sampling) is shown in black.

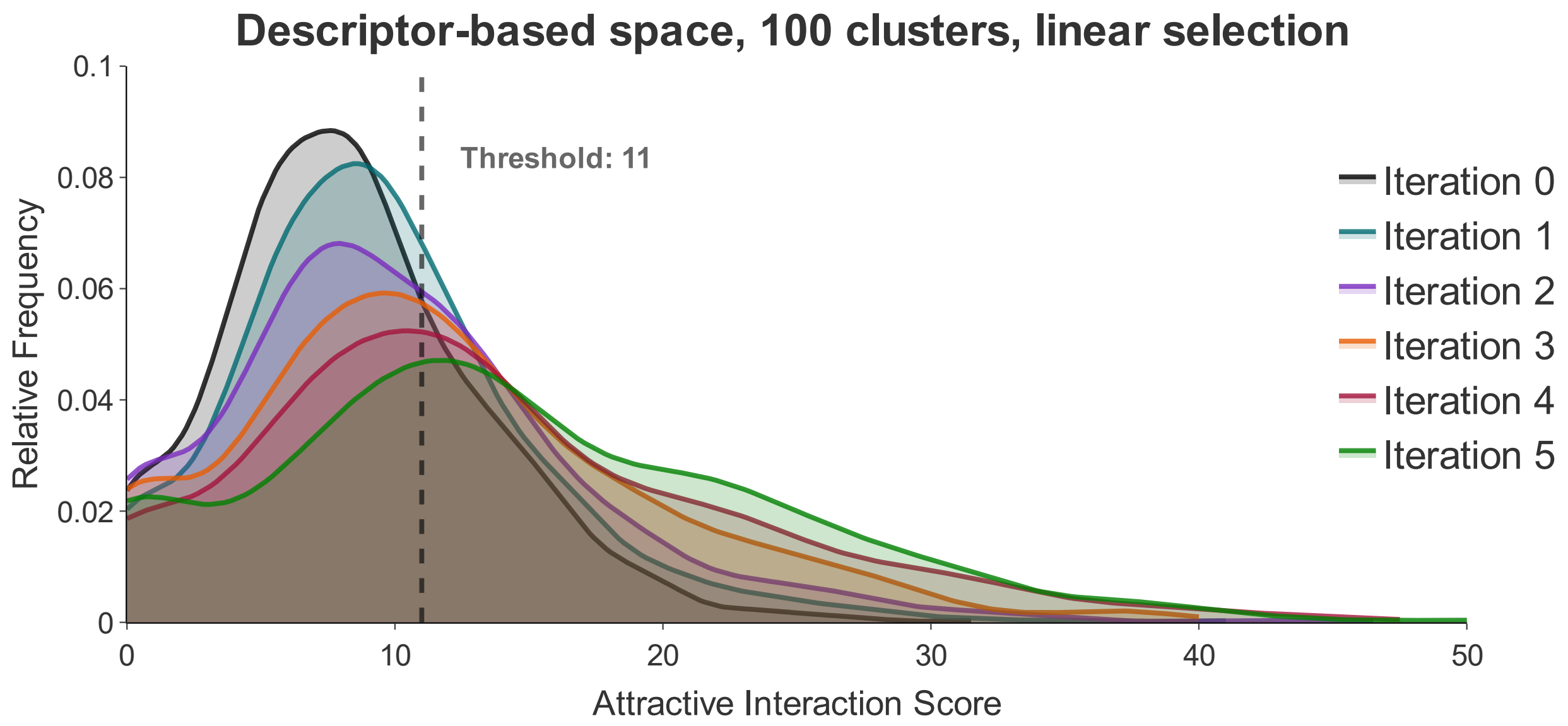

**Figure S20.2.** Attractive interaction scores for molecules generated by the pretrained model (iteration 0) and by the model after each of the five iterations of active learning where, prior to sampling for docking, molecules in the chemical space are grouped into 100 clusters, and cluster scores are converted into sampling fractions using the *linear* approach.

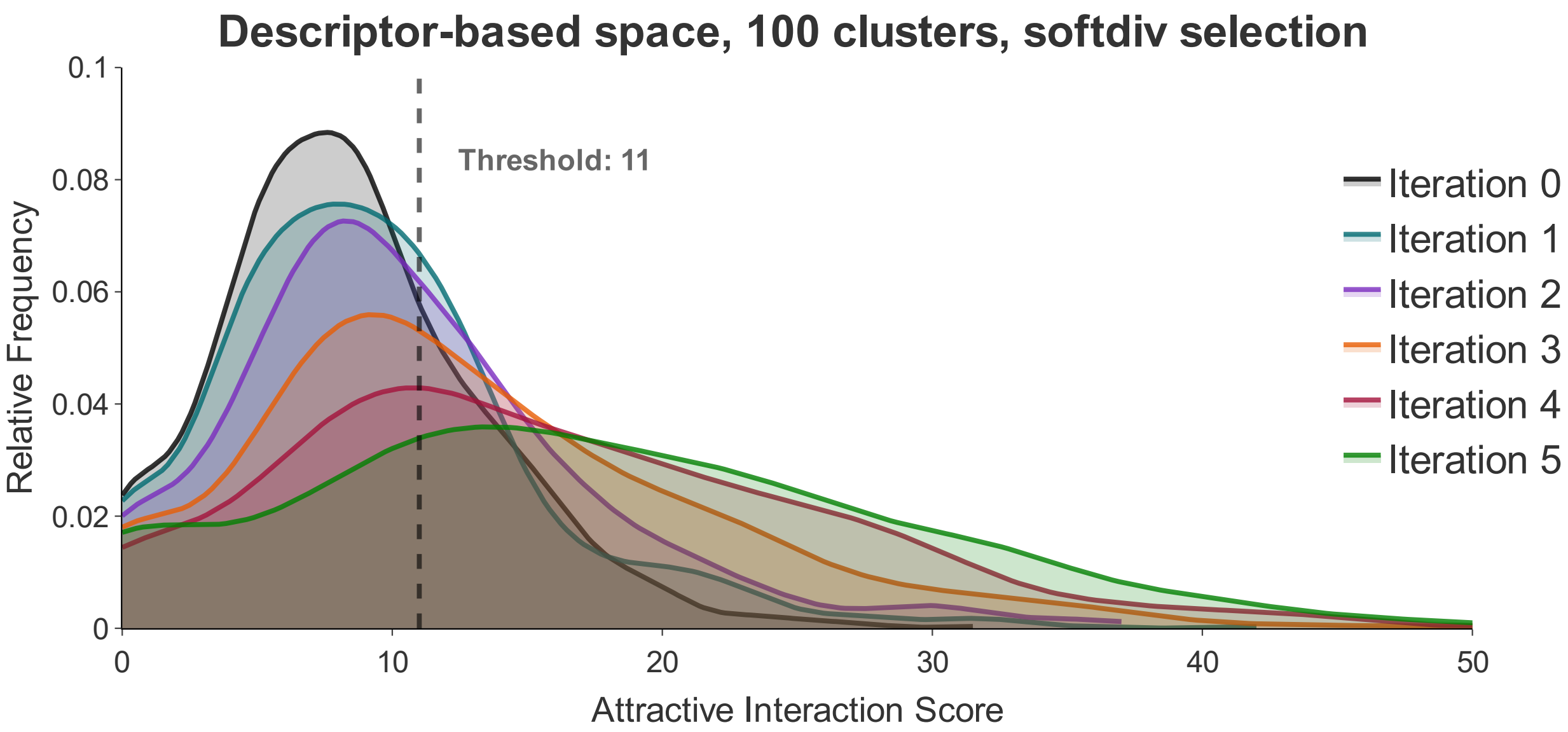

**Figure S20.3.** Attractive interaction scores for molecules generated by the pretrained model (iteration 0) and by the model after each of the five iterations of active learning where, prior to sampling for docking, molecules in the chemical space are grouped into 100 clusters, and cluster scores are converted into sampling fractions using the *softdiv* approach.

**Table S20.4.** Statistics of the distribution of attractive interaction scores, when molecules are selected randomly (naïve active learning), with no clustering.

| Iteration | Percent > 11 | Q1 | Q2 | Mean | Q3 | Max | Std |
|---|---|---|---|---|---|---|---|
| 0 | 26.20 | 5.50 | 8.00 | 8.41 | 11.00 | 33.00 | 4.58 |
| 1 | 32.40 | 5.50 | 9.00 | 9.27 | 11.50 | 35.00 | 4.79 |
| 2 | 35.00 | 6.50 | 9.00 | 9.67 | 12.50 | 33.00 | 4.89 |
| 3 | 40.00 | 6.88 | 9.50 | 10.45 | 13.50 | 37.00 | 5.31 |
| 4 | 44.80 | 7.00 | 10.00 | 11.03 | 13.63 | 42.00 | 6.16 |
| 5 | 44.20 | 7.00 | 10.00 | 11.13 | 13.50 | 38.50 | 6.05 |

[a] The percentage of generated molecules with attractive interaction scores equal to or above our score threshold is shown (Percent > 11), as well as the score at the first quartile (Q1), second quartile (Q2), Mean, third quartile (Q3), maximum (Max), and standard deviation (Std) of the distribution.

[b] Iteration 0 refers to the pretraining phase, while later iterations refer to the active learning phases.

[c] This table corresponds to the distribution in Figure 4A of main text.

**Table S20.5.** Statistics of the distribution of attractive interaction scores, when molecules are clustered into 100 groups and cluster scores are sampled *uniformly*.

| Iteration | Percent > 11 | Q1 | Q2 | Mean | Q3 | Max | Std |
|---|---|---|---|---|---|---|---|
| 0 | 12.00 | 6.68 | 8.58 | 8.43 | 9.85 | 15.80 | 2.31 |
| 1 | 24.00 | 7.75 | 9.10 | 9.17 | 10.88 | 20.80 | 3.40 |
| 2 | 30.00 | 8.17 | 9.40 | 9.96 | 11.50 | 21.50 | 3.41 |
| 3 | 35.00 | 7.86 | 10.07 | 10.23 | 11.63 | 24.00 | 3.97 |
| 4 | 50.00 | 8.88 | 10.89 | 11.28 | 13.46 | 25.45 | 4.11 |
| 5 | 50.00 | 8.80 | 10.94 | 11.84 | 14.48 | 25.80 | 4.63 |

[a] The percentage of generated molecules with attractive interaction scores equal to or above our score threshold is shown (Percent > 11), as well as the score at the first quartile (Q1), second quartile (Q2), Mean, third quartile (Q3), maximum (Max), and standard deviation (Std) of the distribution.

[b] Iteration 0 refers to the pretraining phase, while later iterations refer to the active learning phases.

[c] This table corresponds to the distribution in Figure 4B of main text.

**Table S20.6.** Statistics of the distribution of attractive interaction scores, when molecules are clustered into 100 groups and cluster scores are converted into sampling fractions using the *linear* method.

| Iteration | Percent > 11 | Q1 | Q2 | Mean | Q3 | Max | Std |
|---|---|---|---|---|---|---|---|
| 0 | 12.00 | 6.68 | 8.58 | 8.43 | 9.85 | 15.80 | 2.31 |
| 1 | 23.23 | 7.88 | 9.20 | 9.54 | 10.71 | 19.95 | 2.73 |
| 2 | 37.00 | 7.78 | 9.98 | 10.24 | 12.10 | 26.25 | 4.40 |
| 3 | 47.00 | 8.29 | 10.55 | 11.60 | 14.51 | 23.90 | 5.09 |
| 4 | 62.00 | 10.19 | 13.27 | 13.19 | 15.84 | 31.41 | 5.09 |
| 5 | 71.00 | 10.50 | 13.43 | 14.00 | 16.93 | 29.50 | 5.50 |

[a] The percentage of generated molecules with attractive interaction scores equal to or above our score threshold is shown (Percent > 11), as well as the score at the first quartile (Q1), second quartile (Q2), Mean, third quartile (Q3), maximum (Max), and standard deviation (Std) of the distribution.
[b] Iteration 0 refers to the pretraining phase, while later iterations refer to the active learning phases.
[c] This table corresponds to the distribution in Figure S4.1.

**Table S20.7.** Statistics of the distribution of attractive interaction scores, when molecules are clustered into 100 groups and cluster scores are converted into sampling fractions using the *softdiv* method.

| Iteration | Percent > 11 | Q1 | Q2 | Mean | Q3 | Max | Std |
|---|---|---|---|---|---|---|---|
| 0 | 28.10 | 5.50 | 8.00 | 8.46 | 11.50 | 31.50 | 4.89 |
| 1 | 34.70 | 5.50 | 8.50 | 9.29 | 12.50 | 42.00 | 5.89 |
| 2 | 42.20 | 6.50 | 9.50 | 10.54 | 13.50 | 37.00 | 6.64 |
| 3 | 54.20 | 7.50 | 11.50 | 13.07 | 18.00 | 55.00 | 8.64 |
| 4 | 65.90 | 8.50 | 14.25 | 15.82 | 22.00 | 56.50 | 9.99 |
| 5 | 71.20 | 9.50 | 16.00 | 17.32 | 24.50 | 51.00 | 10.90 |

[a] The percentage of generated molecules with attractive interaction scores equal to or above our score threshold is shown (Percent > 11), as well as the score at the first quartile (Q1), second quartile (Q2), Mean, third quartile (Q3), maximum (Max), and standard deviation (Std) of the distribution.
[b] Iteration 0 refers to the pretraining phase, while later iterations refer to the active learning phases.
[c] This table corresponds to the distribution in Figure S4.2.

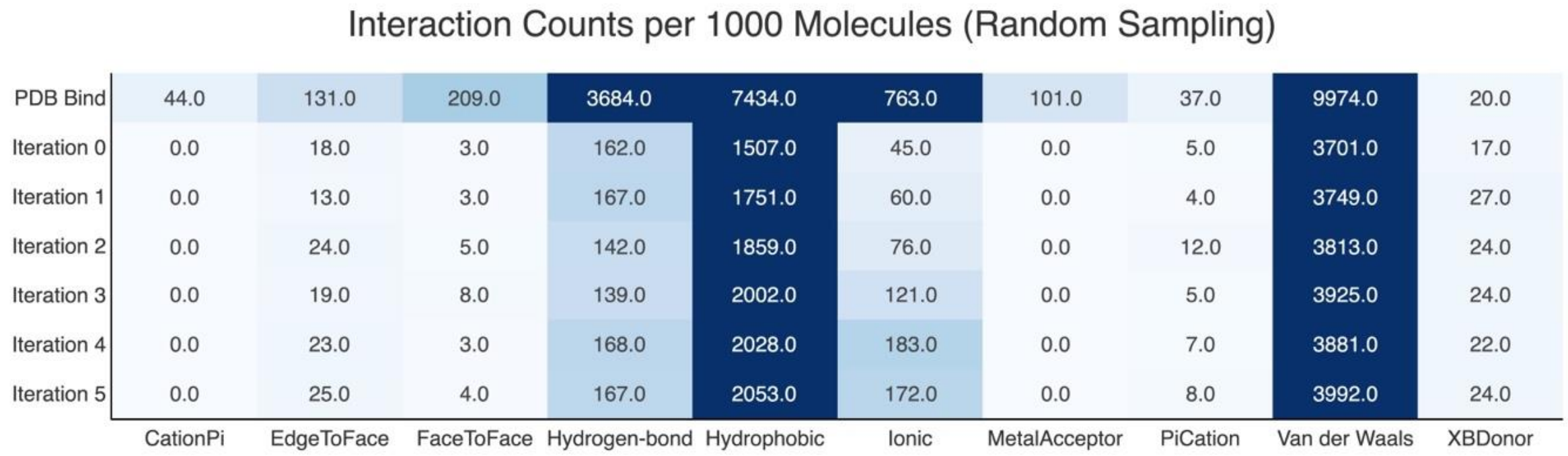

**Figure S20.8.** Counts of interactions of each type for 1000 scored molecules generated by the pretrained model (iteration 0) and by the model after each of the five rounds of naïve active learning with *random sampling*. A count of interactions from 1000 protein-ligand complexes randomly sampled from the refined set of PDBbind v2020 is included for comparison. These counts correspond to the score distribution in Figure 4A of main text.

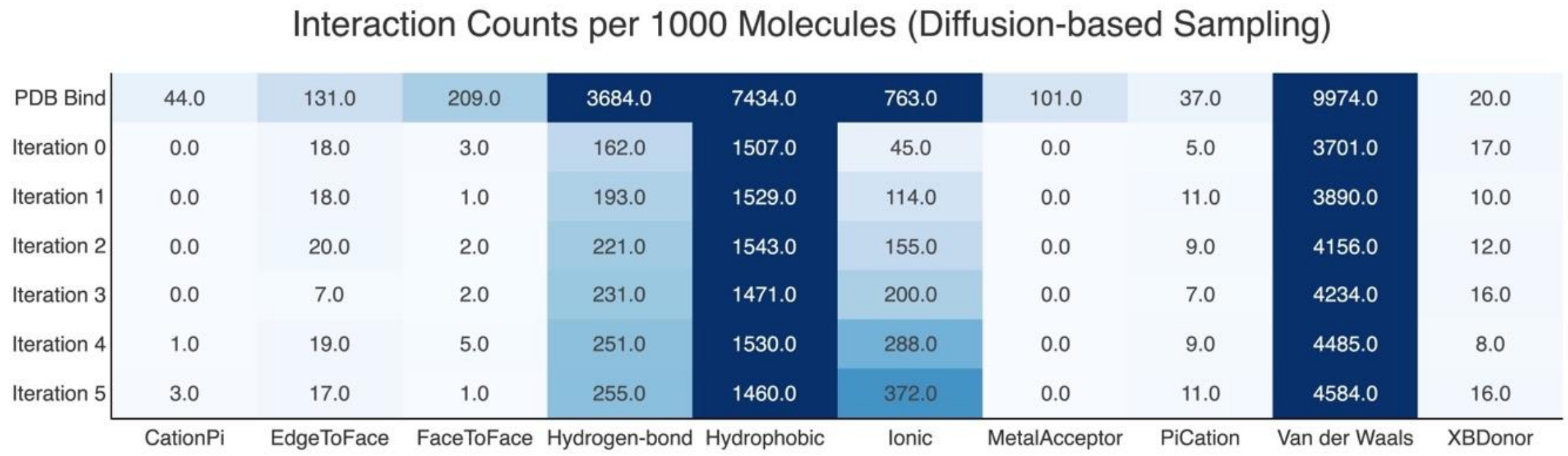

**Figure S20.9.** Counts of interactions of each type for 1000 molecules generated by the pretrained model (iteration 0) and by the model after each of the five rounds of active learning with clustering into 100 groups and *uniform* selection from each cluster. A count of interactions from 1000 protein-ligand complexes randomly sampled from the refined set of PDBbind v2020 is included for comparison. These counts correspond to the score distribution in Figure 4B of main text.

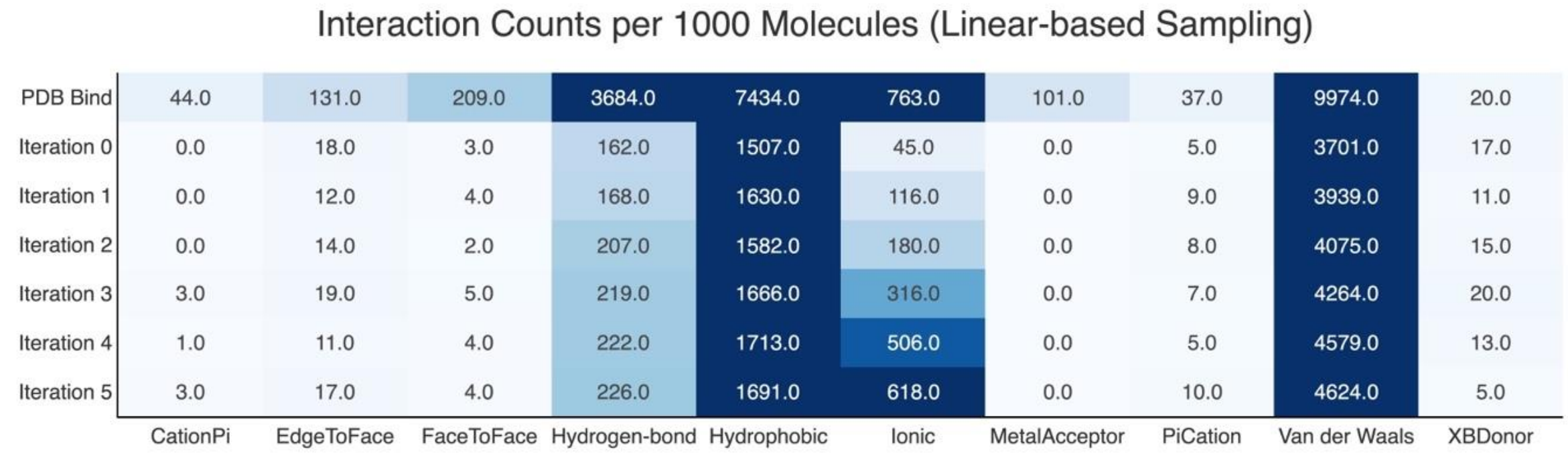

**Figure S20.10.** Counts of interactions of each type for 1000 scored molecules generated by the pretrained model (iteration 0) and by the model after each of the five rounds of active learning with clustering into 100 groups and conversion of cluster scores into sampling fractions using the *linear* method. A count of interactions from 1000 protein-ligand complexes randomly sampled from the refined set of PDBbind v2020 is included for comparison. These counts correspond to the score distribution in Figure S4.1.

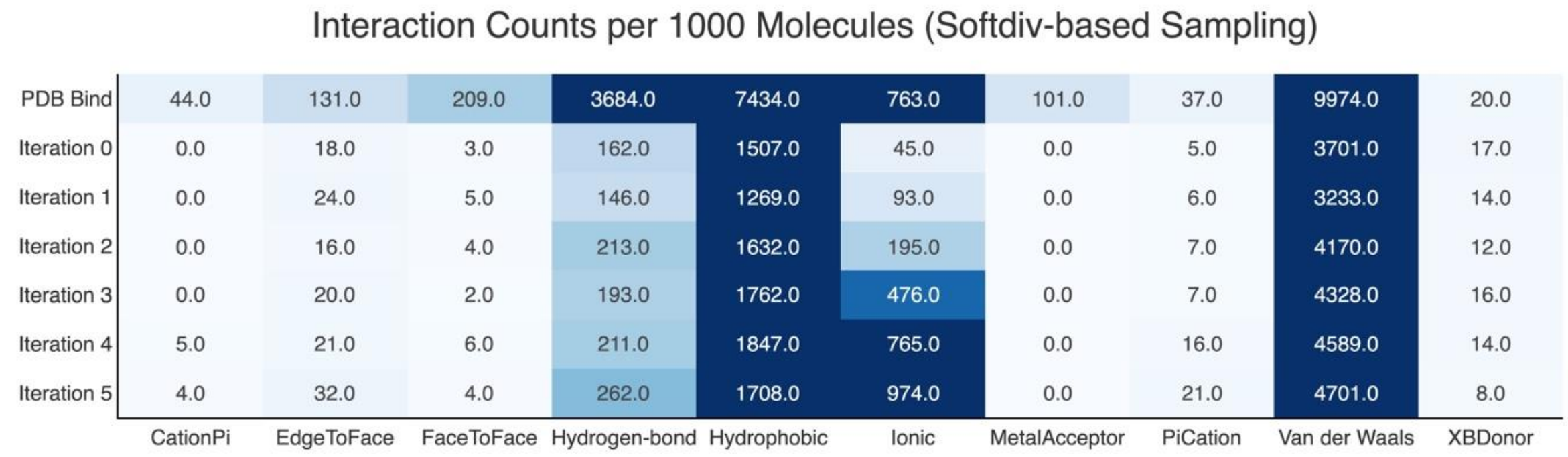

**Figure S20.11.** Counts of interactions of each type for 1000 scored molecules generated by the pretrained model (iteration 0) and by the model after each of the five rounds of active learning with clustering into 100 groups and conversion of cluster scores into sampling fractions using the *softdiv* method. A count of interactions from 1000 protein-ligand complexes randomly sampled from the refined set of PDBbind v2020 is included for comparison. These counts correspond to the score distribution in Figure S4.2.

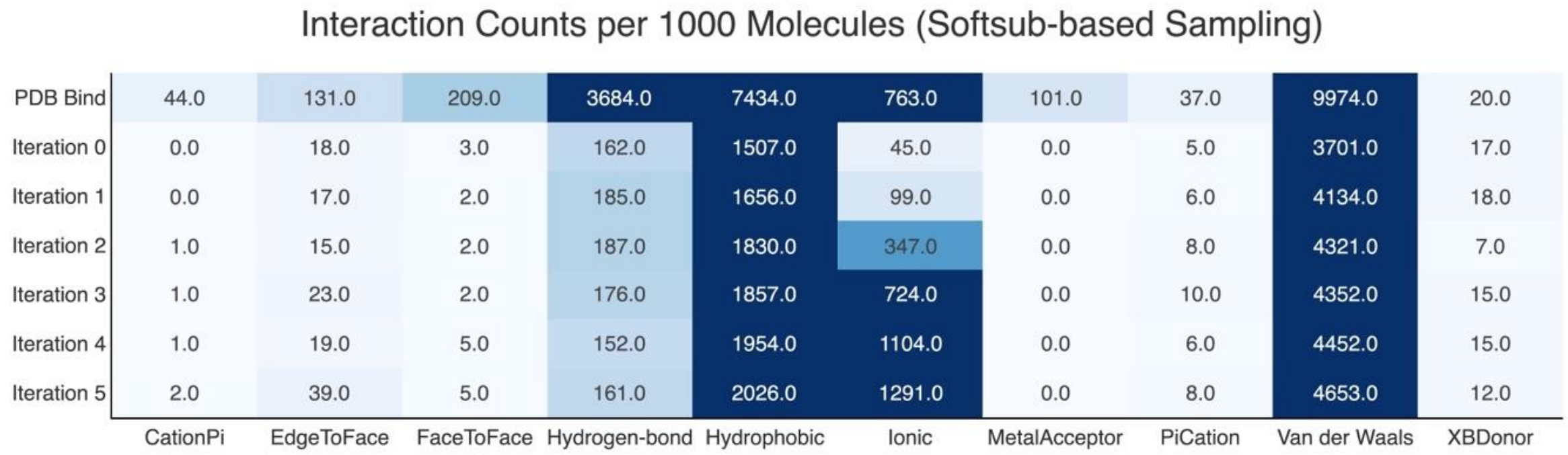

Interaction Counts per 1000 Molecules (Softsub-based Sampling)

| | CationPi | EdgeToFace | FaceToFace | Hydrogen-bond | Hydrophobic | Ionic | MetalAcceptor | PiCation | Van der Waals | XBDonor |
|---|---|---|---|---|---|---|---|---|---|---|
| PDB Bind | 44.0 | 131.0 | 209.0 | 3684.0 | 7434.0 | 763.0 | 101.0 | 37.0 | 9974.0 | 20.0 |
| Iteration 0 | 0.0 | 18.0 | 3.0 | 162.0 | 1507.0 | 45.0 | 0.0 | 5.0 | 3701.0 | 17.0 |
| Iteration 1 | 0.0 | 17.0 | 2.0 | 185.0 | 1656.0 | 99.0 | 0.0 | 6.0 | 4134.0 | 18.0 |
| Iteration 2 | 1.0 | 15.0 | 2.0 | 187.0 | 1830.0 | 347.0 | 0.0 | 8.0 | 4321.0 | 7.0 |
| Iteration 3 | 1.0 | 23.0 | 2.0 | 176.0 | 1857.0 | 724.0 | 0.0 | 10.0 | 4352.0 | 15.0 |
| Iteration 4 | 1.0 | 19.0 | 5.0 | 152.0 | 1954.0 | 1104.0 | 0.0 | 6.0 | 4452.0 | 15.0 |
| Iteration 5 | 2.0 | 39.0 | 5.0 | 161.0 | 2026.0 | 1291.0 | 0.0 | 8.0 | 4653.0 | 12.0 |

**Figure S20.12.** Counts of interactions of each type for 1000 molecules generated by the pretrained model (iteration 0) and by the model after each of the five rounds of active learning with clustering into 100 groups and conversion of cluster scores into sampling fractions using the *softsub* method. A count of interactions from 1000 protein-ligand complexes randomly sampled from the refined set of PDBbind v2020 is included for comparison. These counts correspond to the score distribution in Figure 4C of the main text.

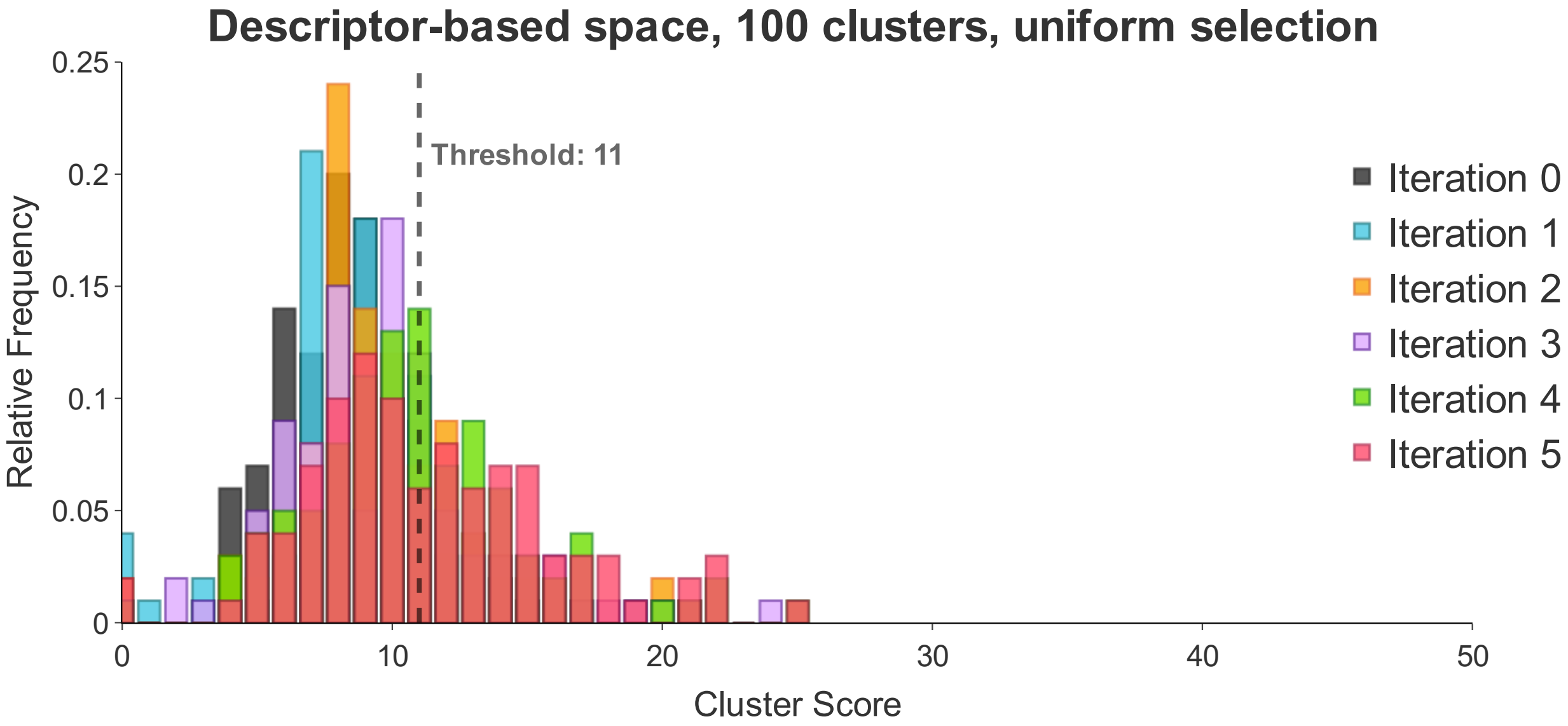

**Figure S20.13.** Cluster scores (obtained as an average of attractive interaction scores for molecules in the cluster) for molecules generated by the pretrained model (iteration 0) and by the model after each of the five iterations of active learning where, prior to sampling for docking, molecules in the chemical space are grouped into 100 clusters, and molecules are sampled from each cluster *uniformly*. These cluster scores correspond to score distribution in Figure 4B of the main text.

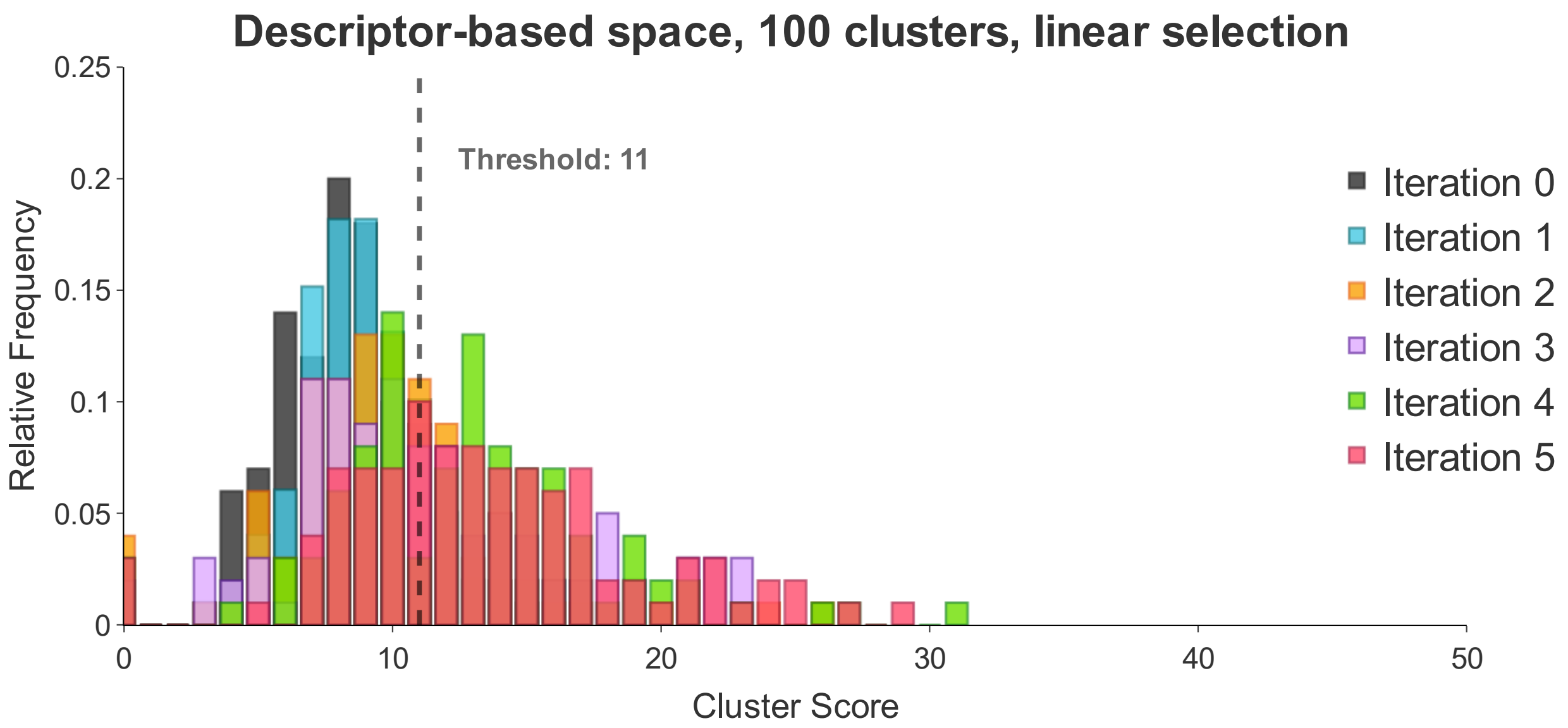

**Figure S20.14.** Cluster scores (obtained as an average of attractive interaction scores for molecules in the cluster) for molecules generated by the pretrained model (iteration 0) and by the model after each of the five iterations of active learning where, prior to sampling for docking, molecules in the chemical space are grouped into 100 clusters, and molecules are sampled from each cluster using the *linear* method. These cluster scores correspond to score distribution in Figure SI4.1.

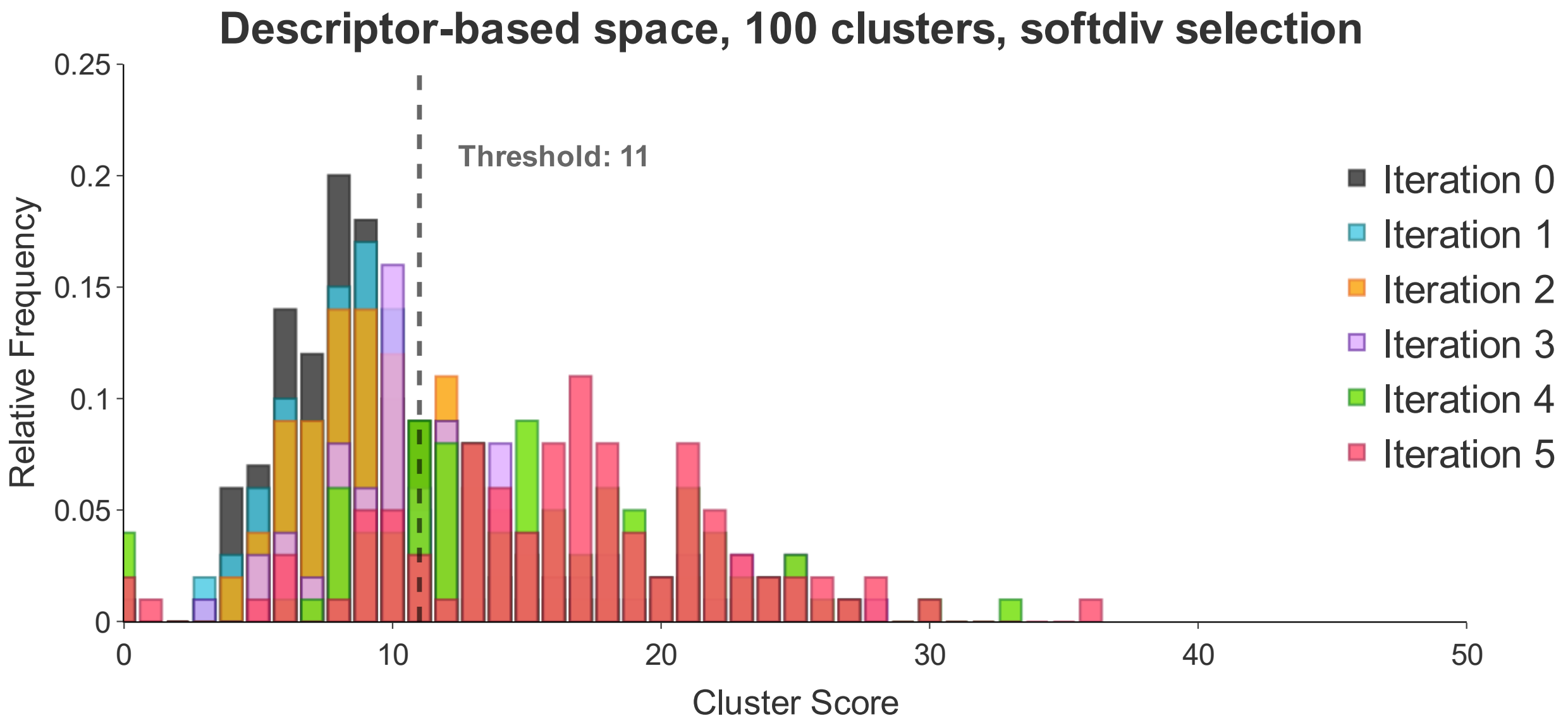

**Figure S20.15.** Cluster scores (obtained as an average of attractive interaction scores for molecules in the cluster) for molecules generated by the pretrained model (iteration 0) and by the model after each of the five iterations of active learning where, prior to sampling for docking, molecules in the chemical space are grouped into 100 clusters, and molecules are sampled from each cluster using the *softdiv* method. These cluster scores correspond to score distribution in Figure SI4.2.

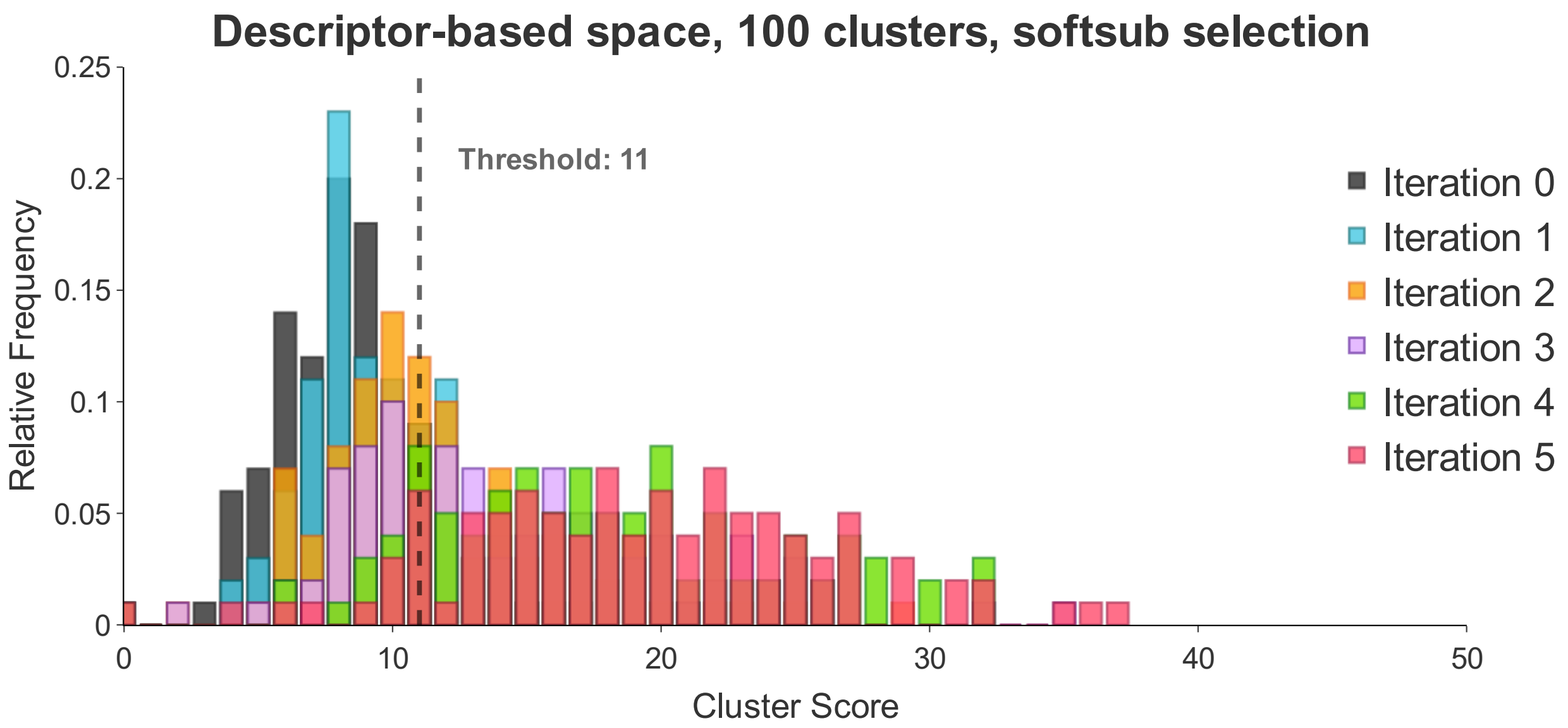

**Figure S20.16.** Cluster scores (obtained as an average of attractive interaction scores for molecules in the cluster) for molecules generated by the pretrained model (iteration 0) and by the model after each of the five iterations of active learning where, prior to sampling for docking, molecules in the chemical space are grouped into 100 clusters, and molecules are sampled from each cluster using the *softsub* method. These cluster scores correspond to score distribution in Figure 4C of the main text.

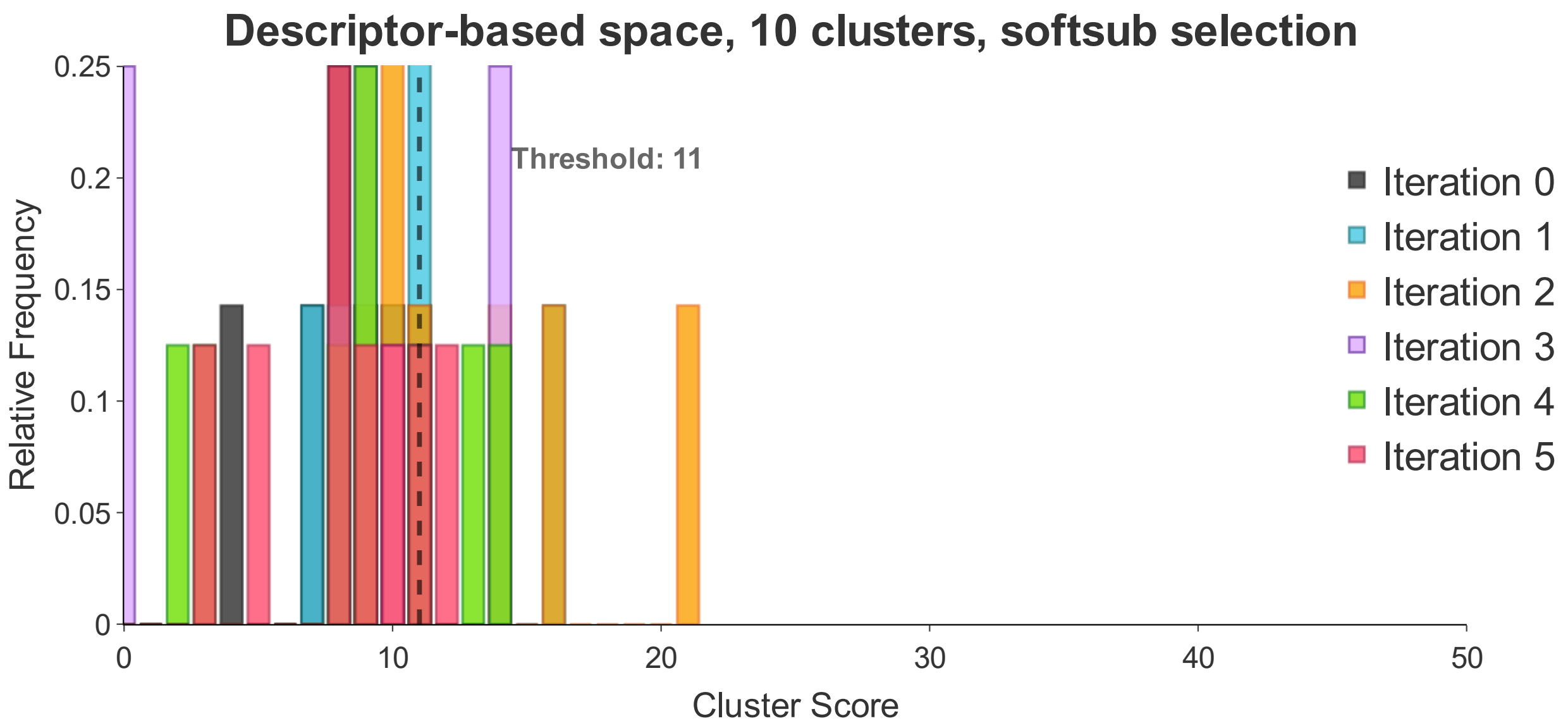

**Figure S20.17.** Cluster scores (obtained as an average of attractive interaction scores for molecules in the cluster) for molecules generated by the pretrained model (iteration 0) and by the model after each of the five iterations of active learning where, prior to sampling for docking, molecules in the chemical space are grouped into 10 clusters, and molecules are sampled from each cluster using the *softsub* method. These cluster scores correspond to score distribution in Figure S10.1 of the main text.

**Section 21:** Distributions of Mean and Median Cluster Scores.

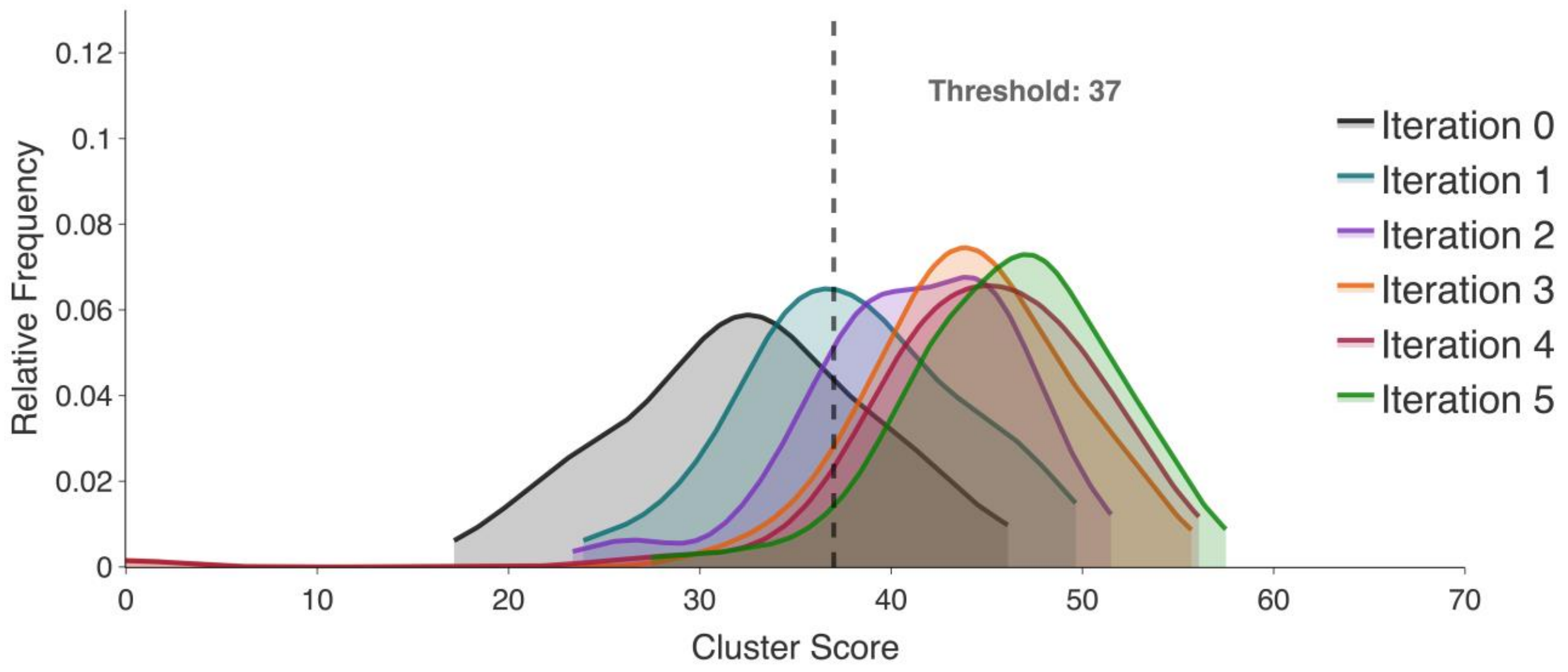

**Figure S21.1**. Mean cluster attractive interaction scores of scored molecules across five iterations of active learning for c-Abl kinase. The distribution for the model pretrained on the combined dataset with generation conditioned on ADMET filters are shown. Iteration 0 refers to the pretraining phase, while later iterations refer to the active learning phases.

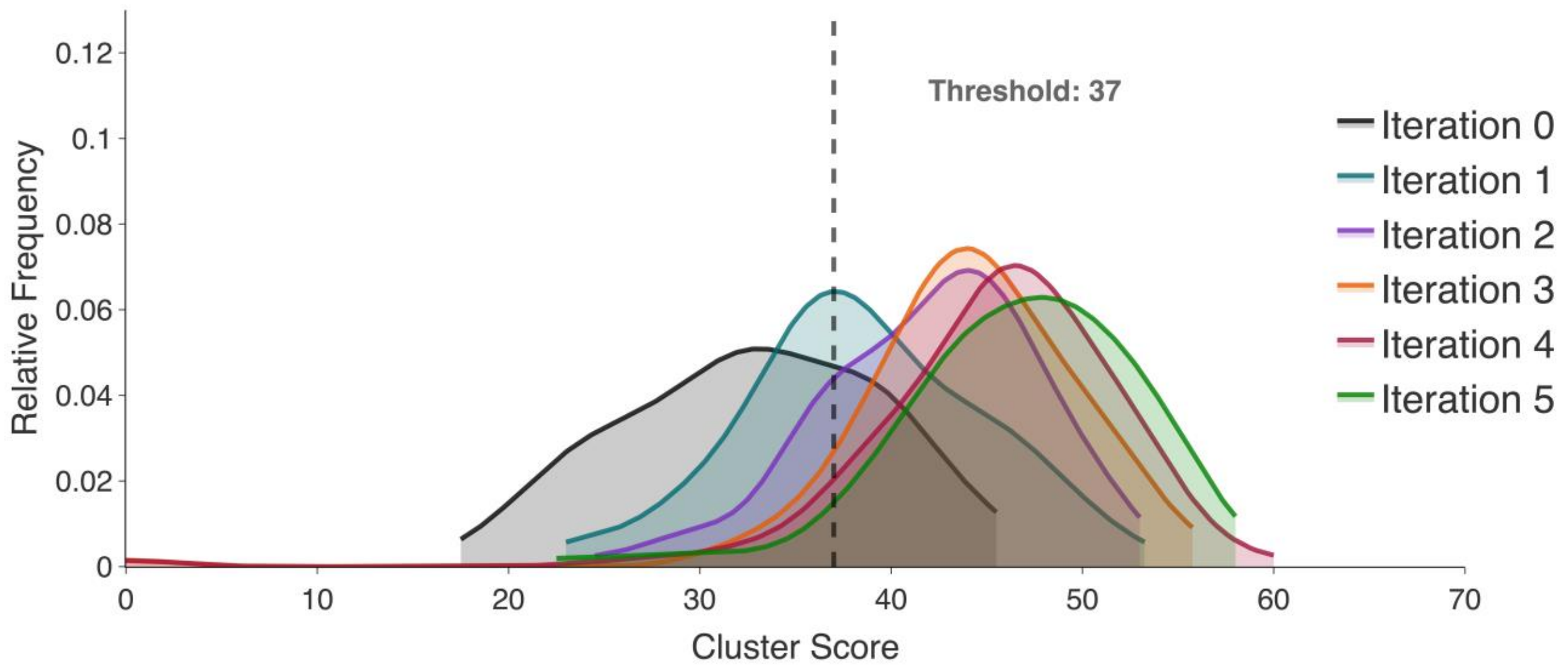

**Figure S21.2**. Median cluster attractive interaction scores of scored molecules across five iterations of active learning for c-Abl kinase. The distribution for the model pretrained on the combined dataset with generation conditioned on ADMET filters are shown. Iteration 0 refers to the pretraining phase, while later iterations refer to the active learning phases.

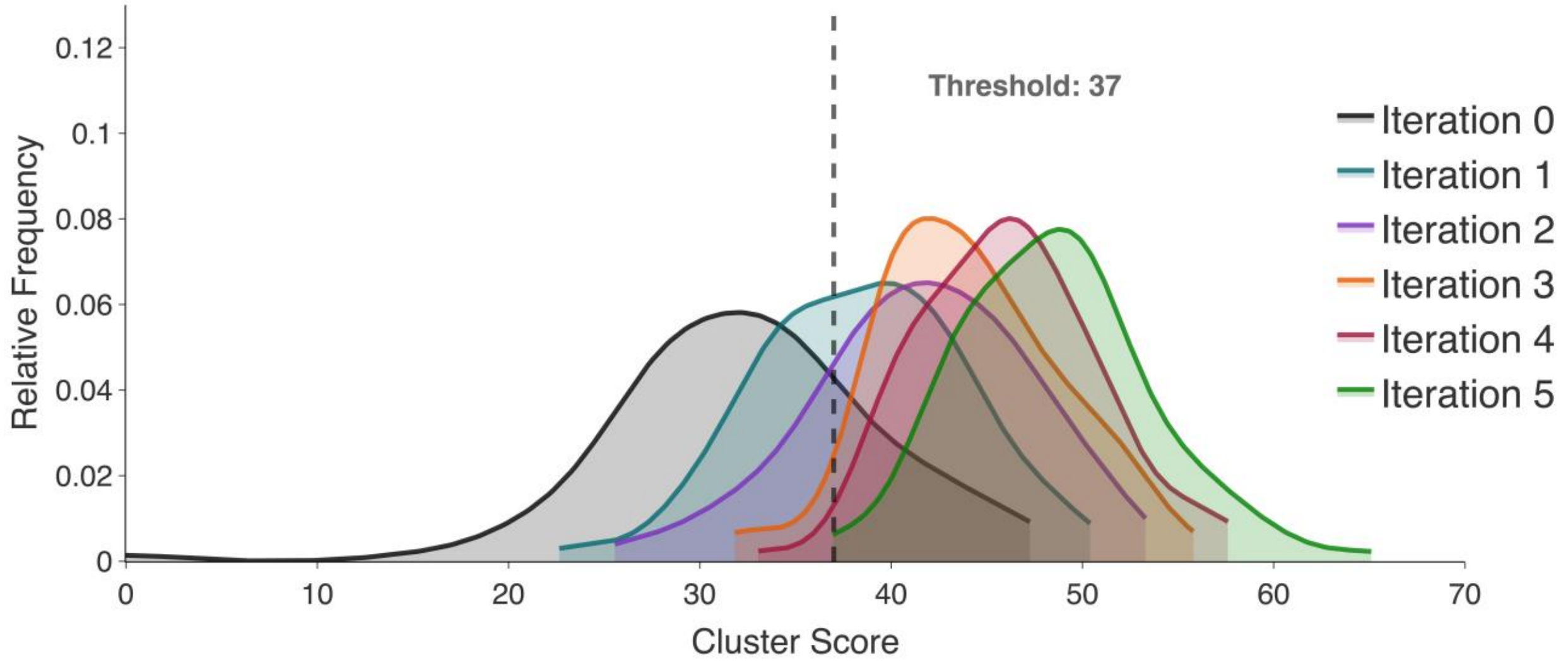

**Figure S21.3**. Mean cluster attractive interaction scores of scored molecules across five iterations of active learning for c-Abl kinase. The distribution for the model pretrained on the combined dataset with generation conditioned on ADMET and functional group filters are shown. Iteration 0 refers to the pretraining phase, while later iterations refer to the active learning phases.

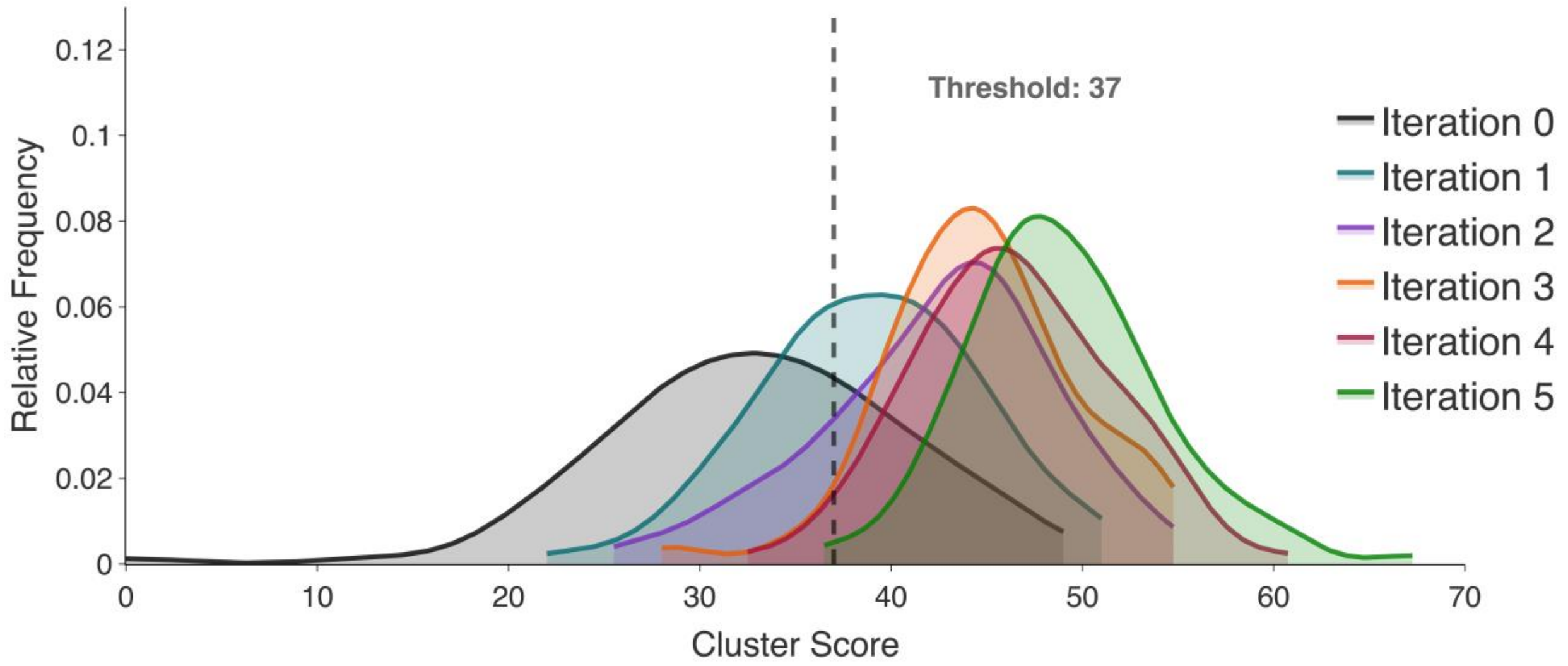

**Figure S21.4**. Median cluster attractive interaction scores of scored molecules across five iterations of active learning for c-Abl kinase. The distribution for the model pretrained on the combined dataset with generation conditioned on ADMET and functional group filters are shown. Iteration 0 refers to the pretraining phase, while later iterations refer to the active learning phases.

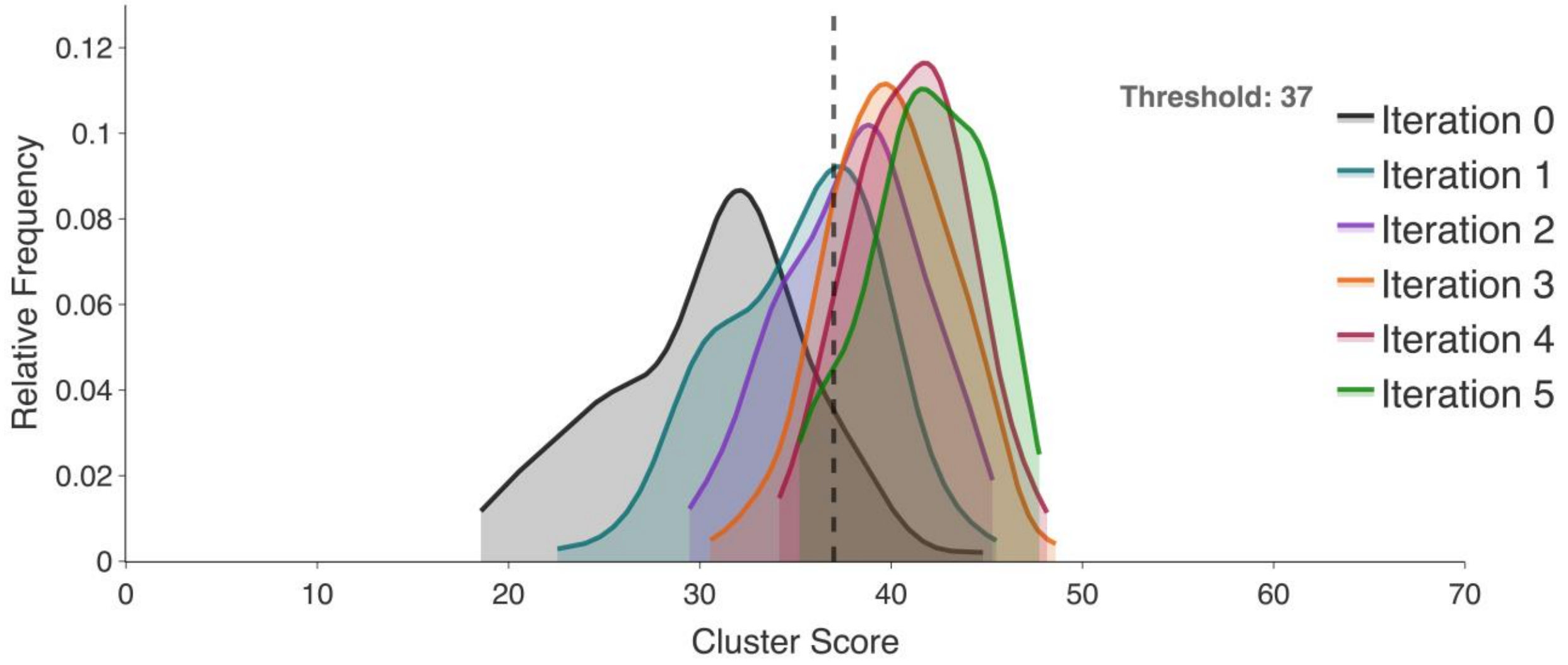

**Figure S21.5**. Mean cluster attractive interaction scores of scored molecules across five iterations of active learning for c-Abl kinase. The distribution for the model pretrained on the MOSES dataset with generation conditioned on ADMET and functional group filters are shown. Iteration 0 refers to the pretraining phase, while later iterations refer to the active learning phases.

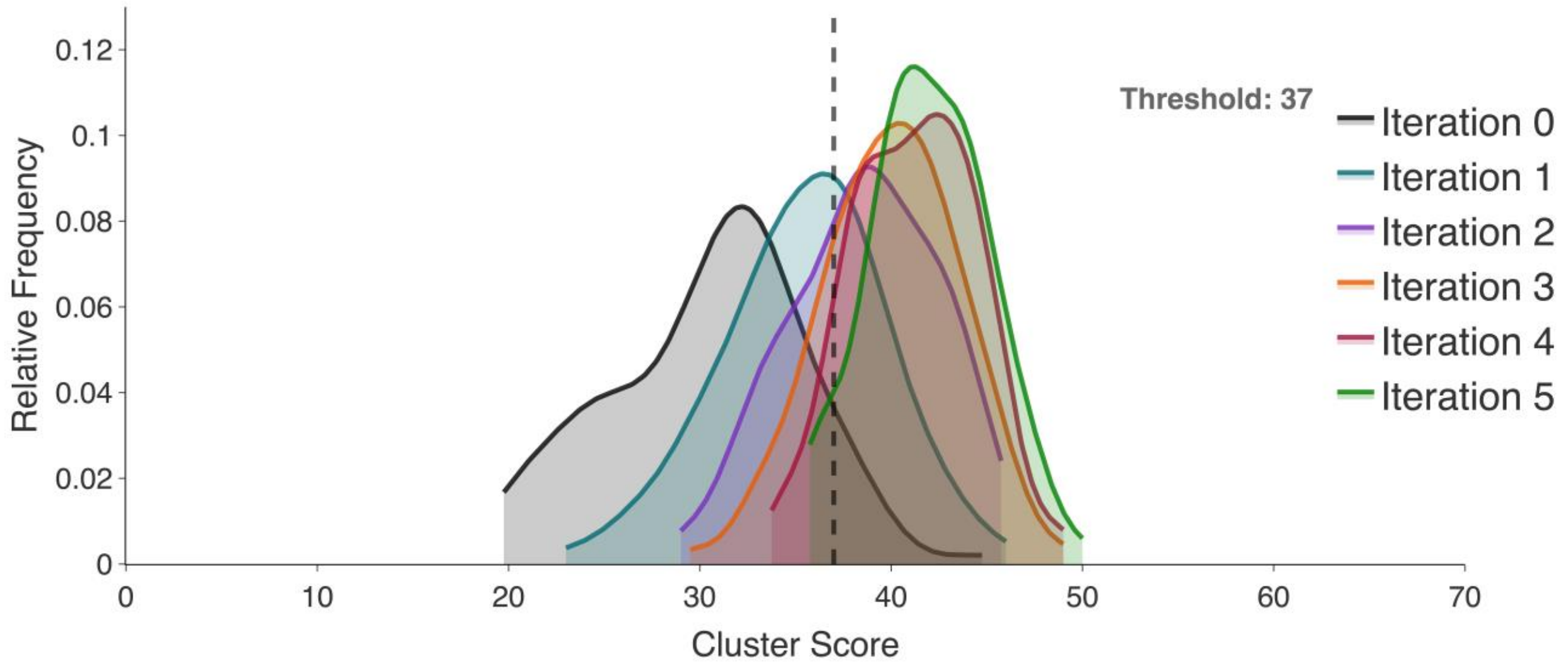

**Figure S21.6**. Median cluster attractive interaction scores of scored molecules across five iterations of active learning for c-Abl kinase. The distribution for the model pretrained on the MOSES dataset with generation conditioned on ADMET and functional group filters are shown. Iteration 0 refers to the pretraining phase, while later iterations refer to the active learning phases.

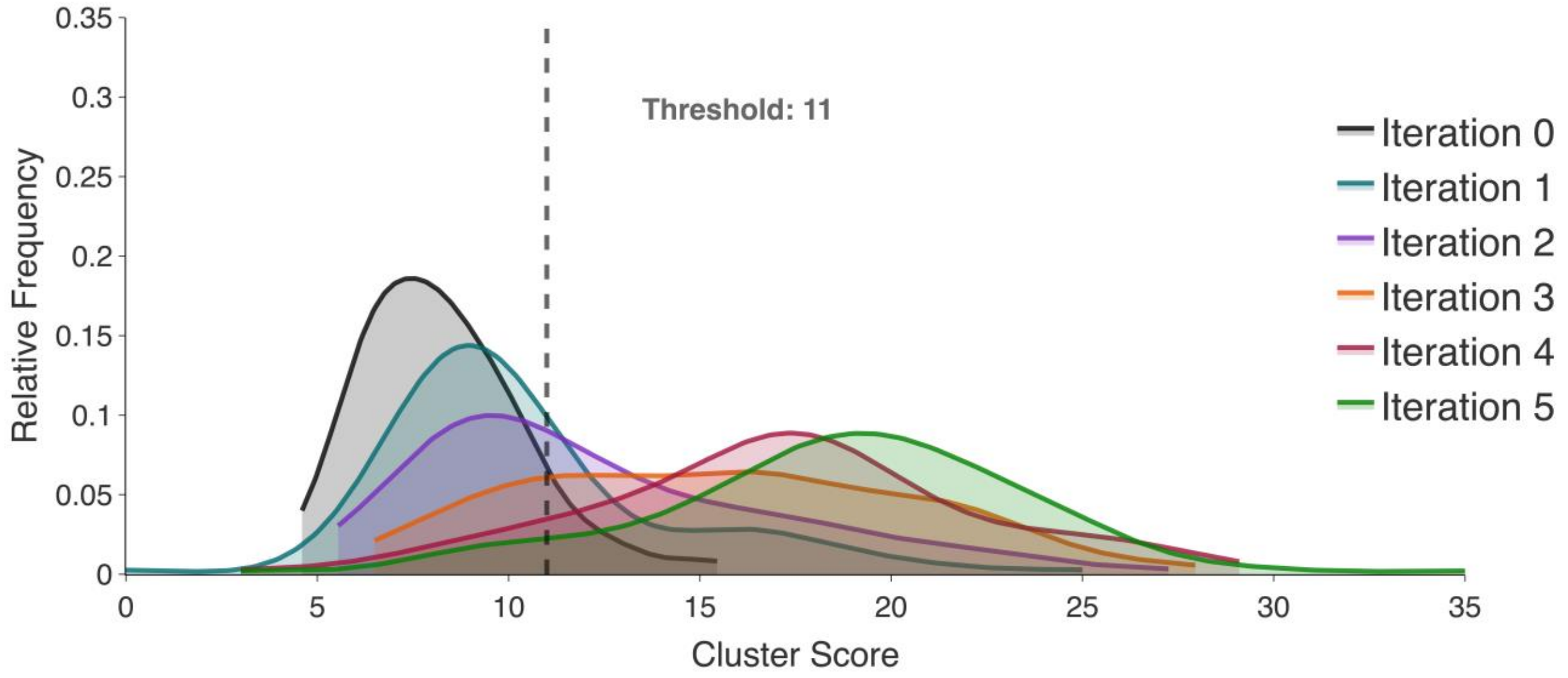

**Figure S21.7**. Mean cluster attractive interaction scores of scored molecules across five iterations of active learning for the HNH domain of Cas9. The distribution for the model pretrained on the combined dataset with generation conditioned on ADMET filters are shown. Iteration 0 refers to the pretraining phase, while later iterations refer to the active learning phases.

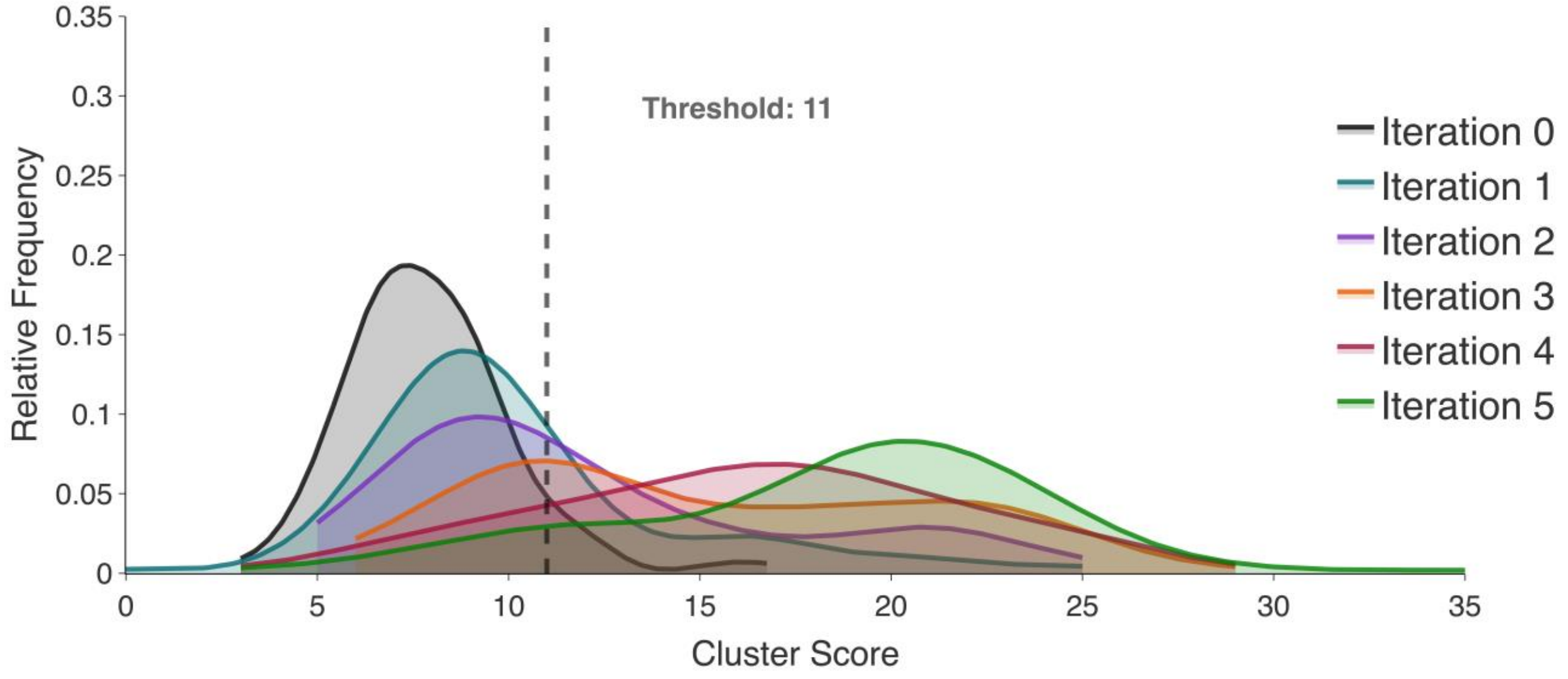

**Figure S21.8**. Median cluster attractive interaction scores of scored molecules across five iterations of active learning for the HNH domain of Cas9. The distribution for the model pretrained on the combined dataset with generation conditioned on ADMET filters are shown. Iteration 0 refers to the pretraining phase, while later iterations refer to the active learning phases.

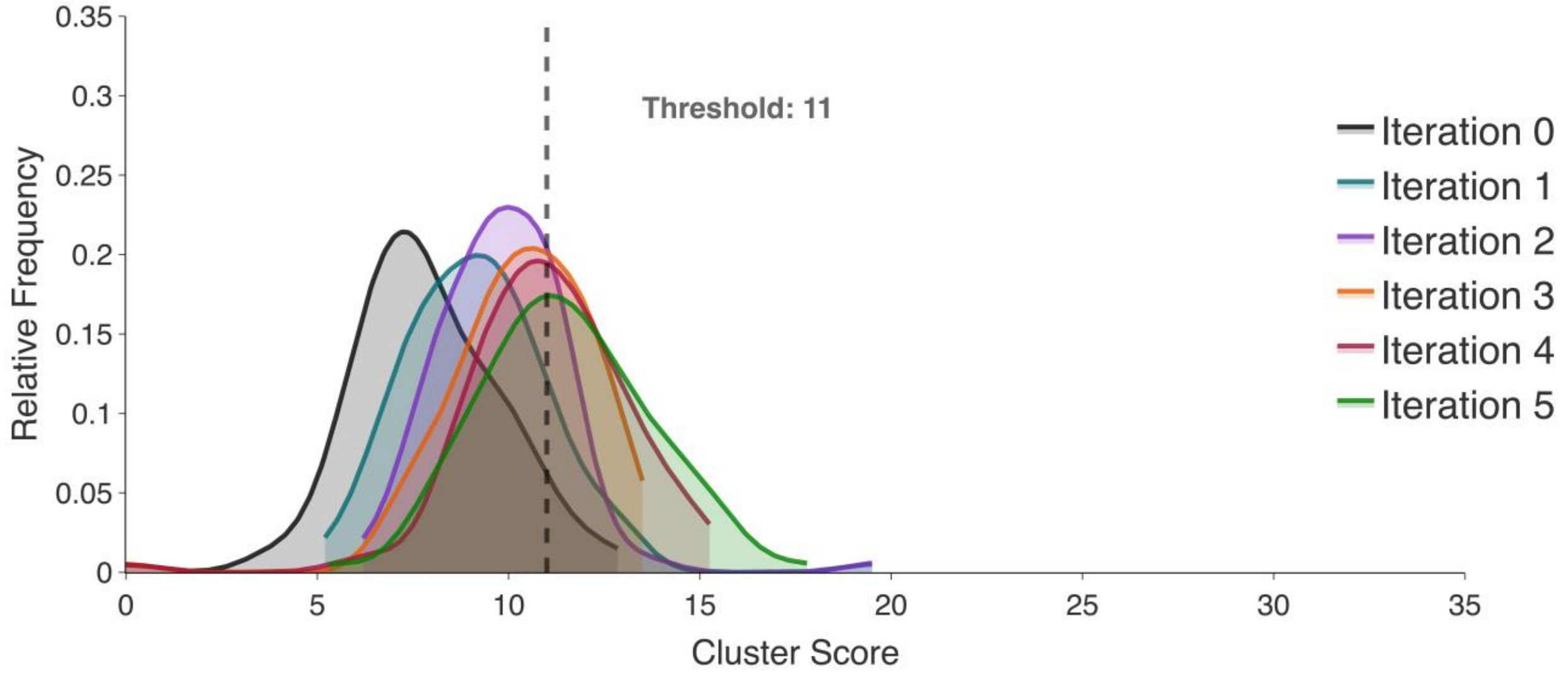

**Figure S21.9.** Mean cluster attractive interaction scores of scored molecules across five iterations of active learning for the HNH domain of Cas9. The distribution for the model pretrained on the combined dataset with generation conditioned on ADMET and functional group filters are shown. Iteration 0 refers to the pretraining phase, while later iterations refer to the active learning phases.

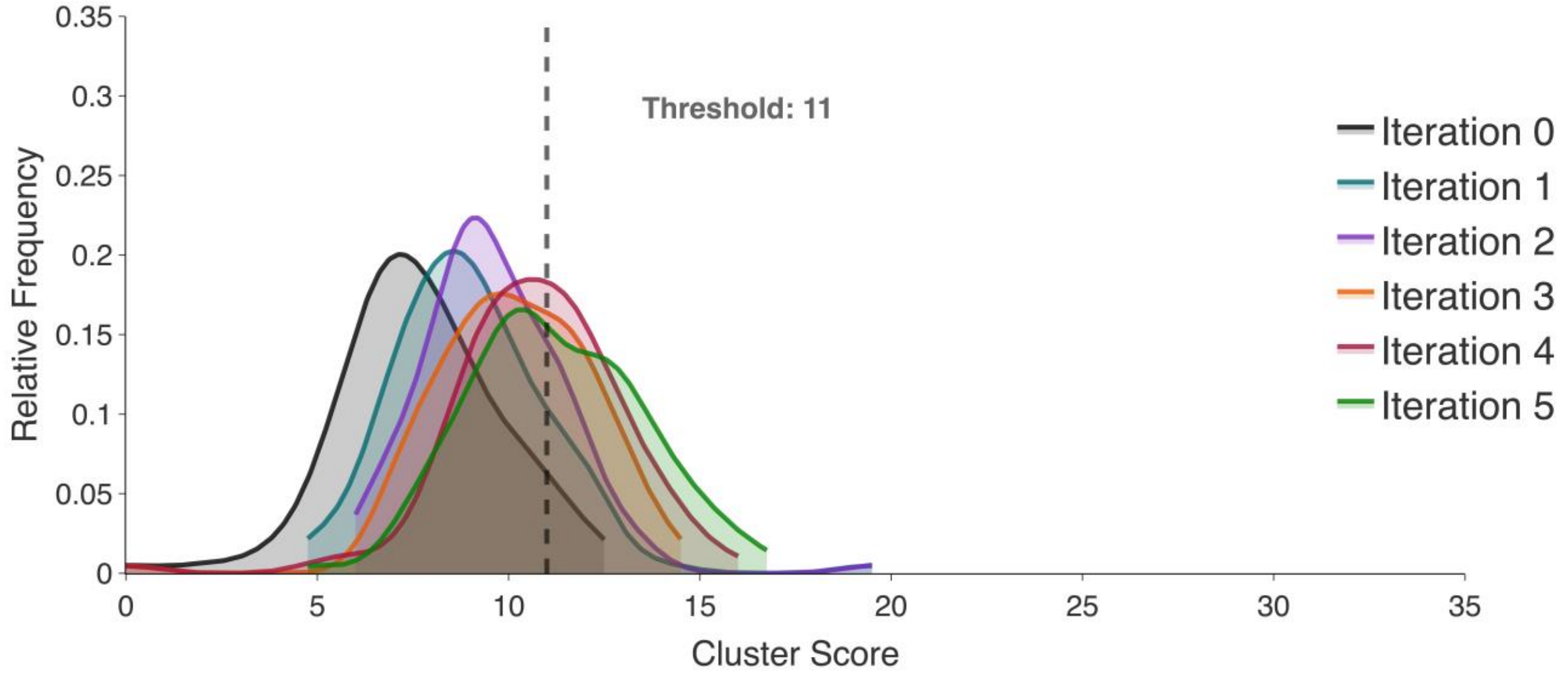

**Figure S21.10.** Median cluster attractive interaction scores of scored molecules across five iterations of active learning for the HNH domain of Cas9. The distribution for the model pretrained on the combined dataset with generation conditioned on ADMET and functional group filters are shown. Iteration 0 refers to the pretraining phase, while later iterations refer to the active learning phases.

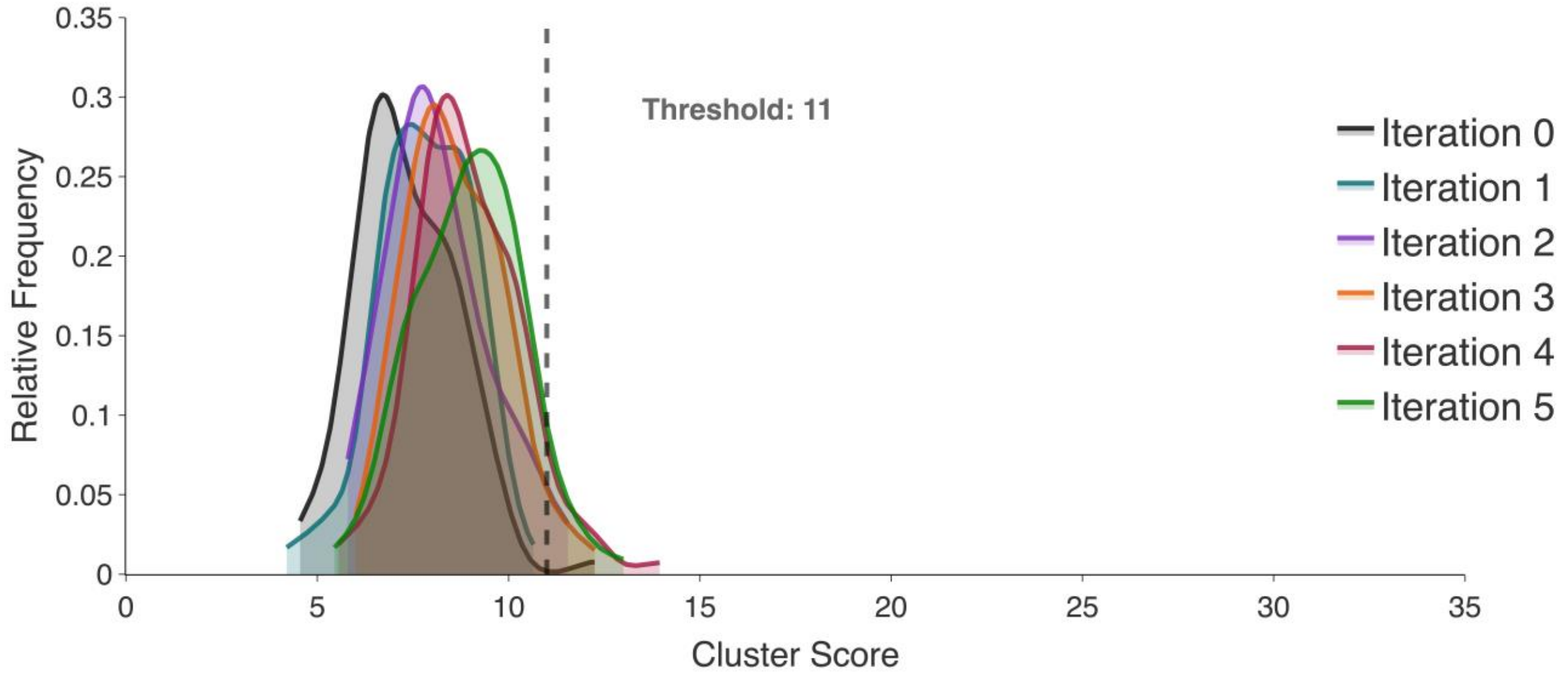

**Figure S21.11.** Mean cluster attractive interaction scores of scored molecules across five iterations of active learning for the HNH domain of Cas9. The distribution for the model pretrained on the MOSES dataset with generation conditioned on ADMET and functional group filters are shown. Iteration 0 refers to the pretraining phase, while later iterations refer to the active learning phases.

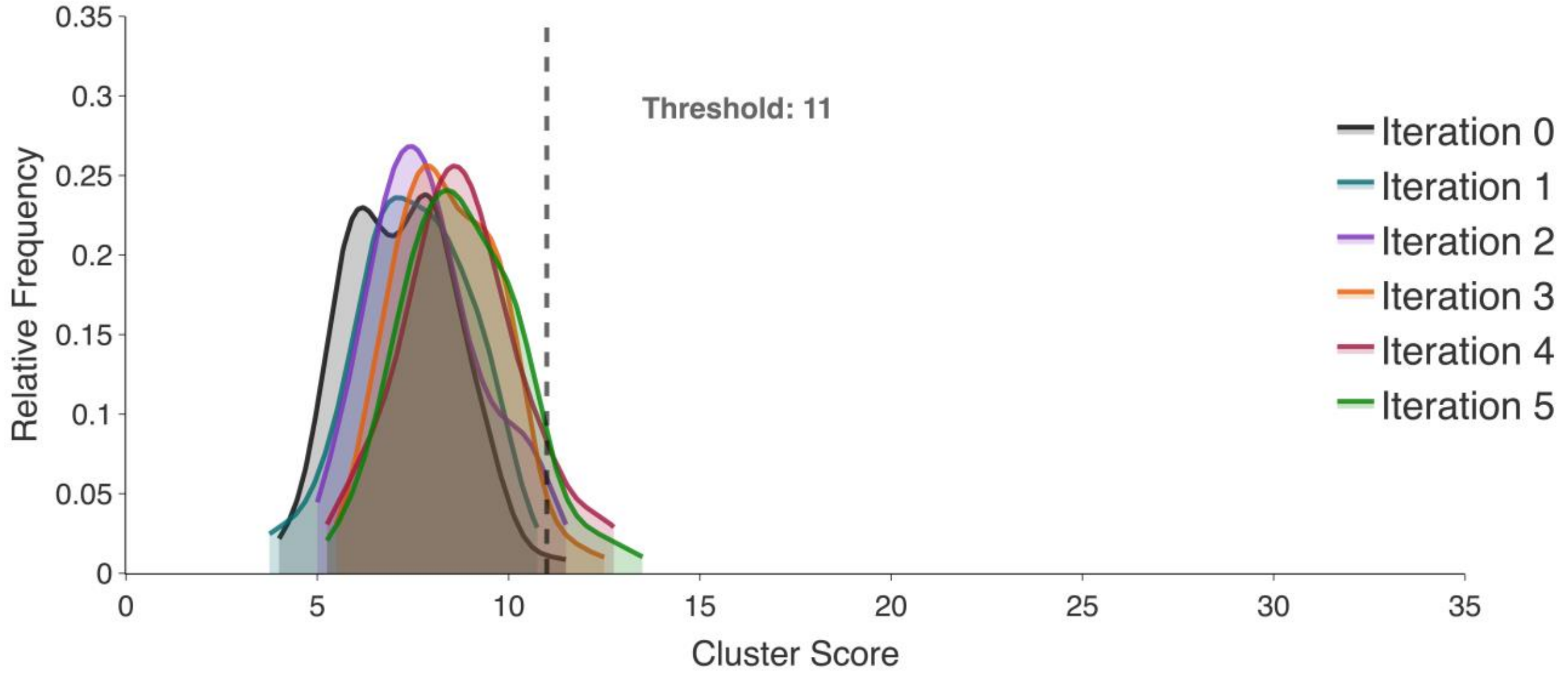

**Figure S21.12.** Median cluster attractive interaction scores of scored molecules across five iterations of active learning for the HNH domain of Cas9. The distribution for the model pretrained on the MOSES dataset with generation conditioned on ADMET and functional group filters are shown. Iteration 0 refers to the pretraining phase, while later iterations refer to the active learning phases.

**Section 22:** Additional Evaluation of Generations across Active Learning Iterations.

Results in this section are regarding the model pretrained on the combined dataset with no filters applied to the generations, for alignment to HNH. It should be noted in the comparisons that poor values for the generations from the model aligned with sets curated with random sampling are likely due to memorization, since we only utilize replicas of scored molecules in this scenario.

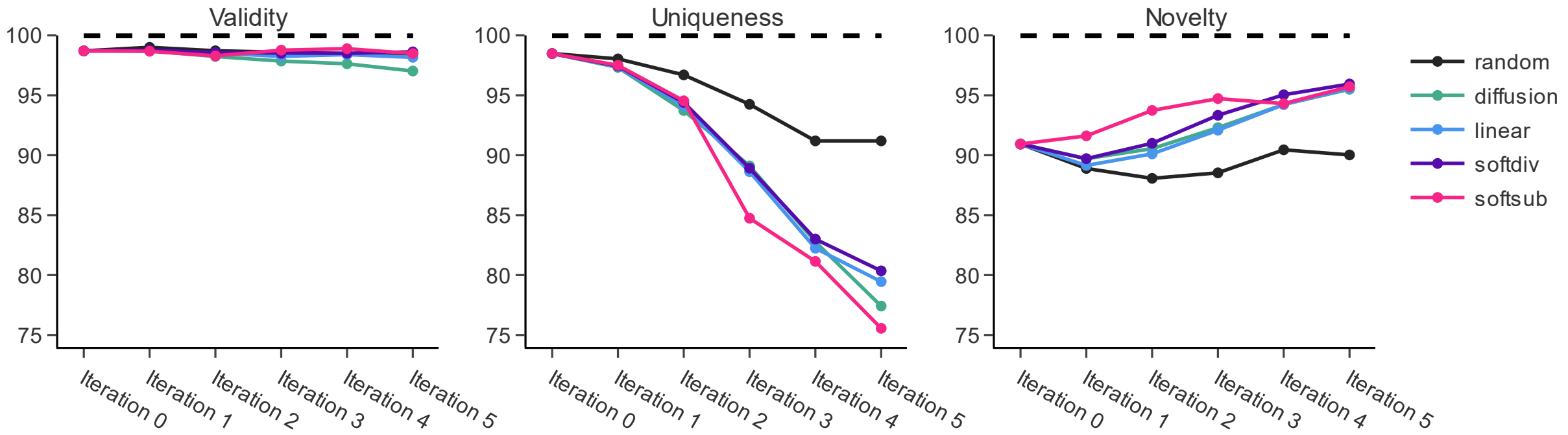

**Figure S22.1.** Percentage of molecules generated by our model that are valid, unique, or novel after pretraining (iteration 0) and five rounds of active learning. Data are shown for different sampling/conversion schemes.

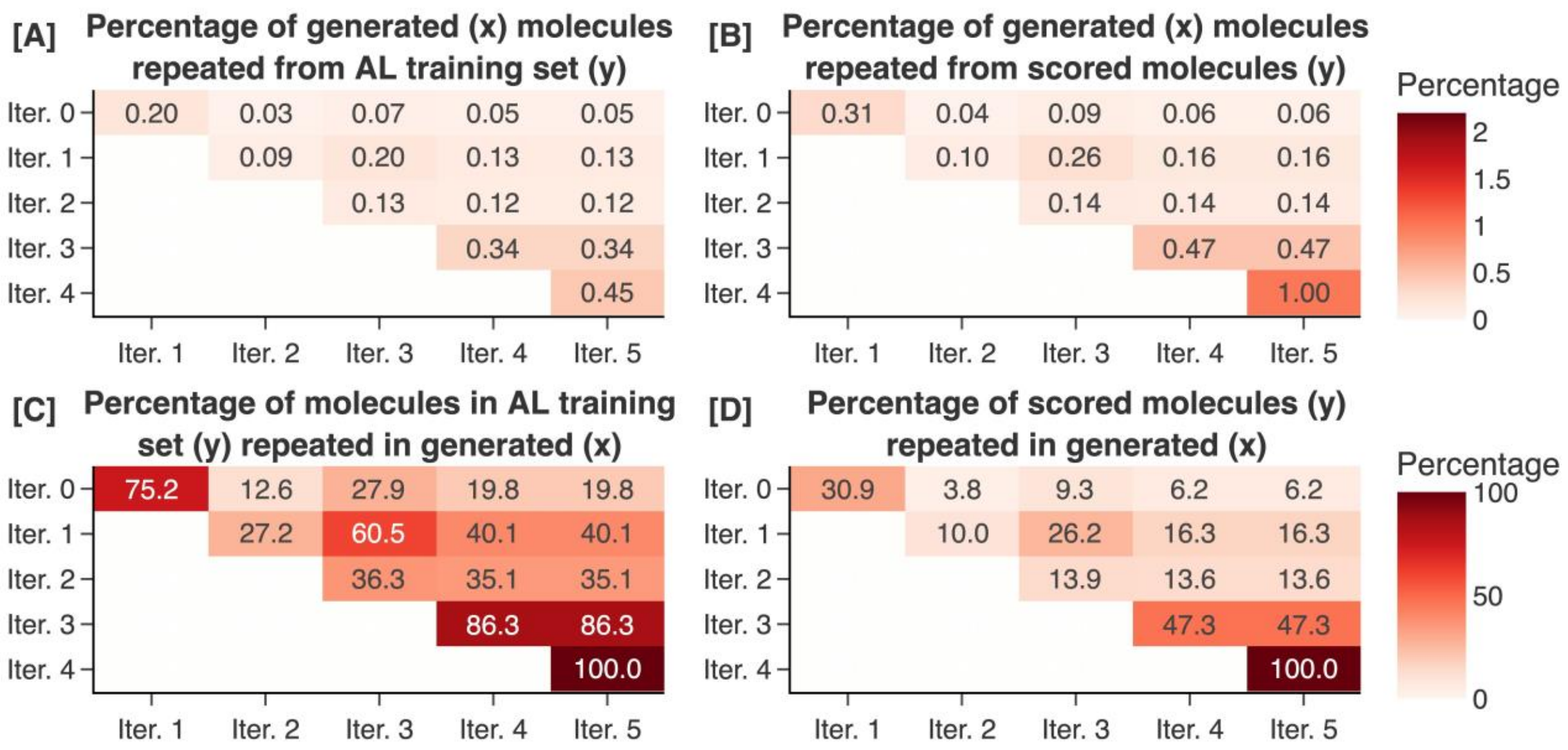

**Figure S22.2.** Memorization of training set by our model over five rounds of naïve active learning with random selection. (A) The percentage of molecules in a set of 100,000 generated at iteration *i* that occur in the training set at iteration *i*-1. (B) The percentage of molecules in a set of 100,000 generated at iteration *i* that occur in the set of scored molecules at iteration *i*-1. (C) The number of molecules from the active training set at iteration *i*-1 that occurs in generations at iteration *i* divided by the size of the active learning training set at iteration *i*-1 multiplied by 100. (D) the number of scored molecules at iteration *i*-1 that occur in generations at iteration *i* divided by number of scored molecules at iteration *i*-1 (i.e., 1000) multiplied by 100.

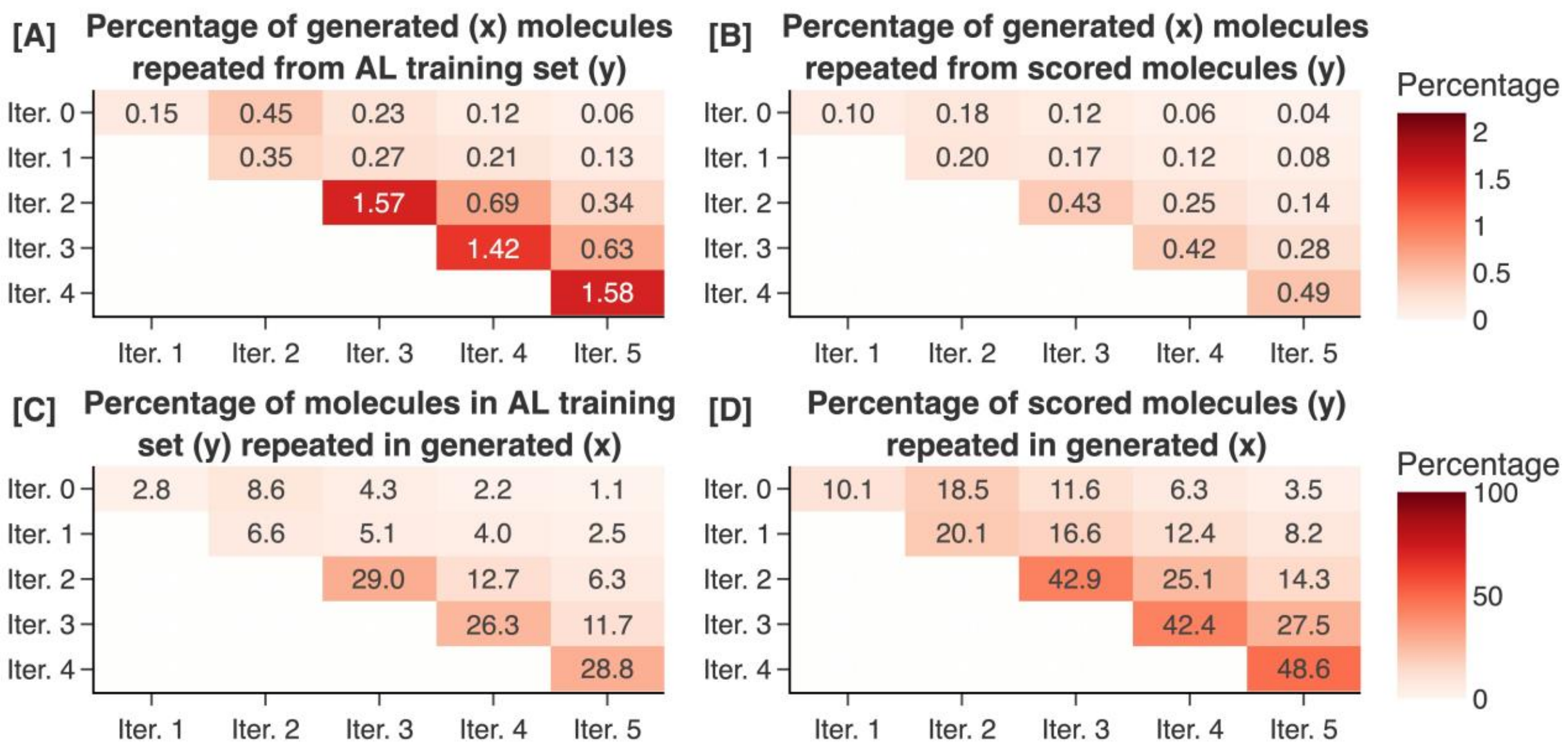

**Figure S22.3.** Memorization of training set by our model over five rounds of active learning with 100 clusters and uniform selection. (A) The percentage of molecules in a set of 100,000 generated at iteration *I* that occur in the training set at iteration *i*-1. (B) The percentage of molecules in a set of 100,000 generated at iteration *i* that occur in the set of scored molecules at iteration *i*-1. (C) The number of molecules from the active training set at iteration *i*-1 that occurs in generations at iteration *i* divided by the size of the active learning training set at iteration *i*-1 multiplied by 100. (D) the number of scored molecules at iteration *i*-1 that occur in generations at iteration *i* divided by number of scored molecules at iteration *i*-1 (i.e., 1000) multiplied by 100.

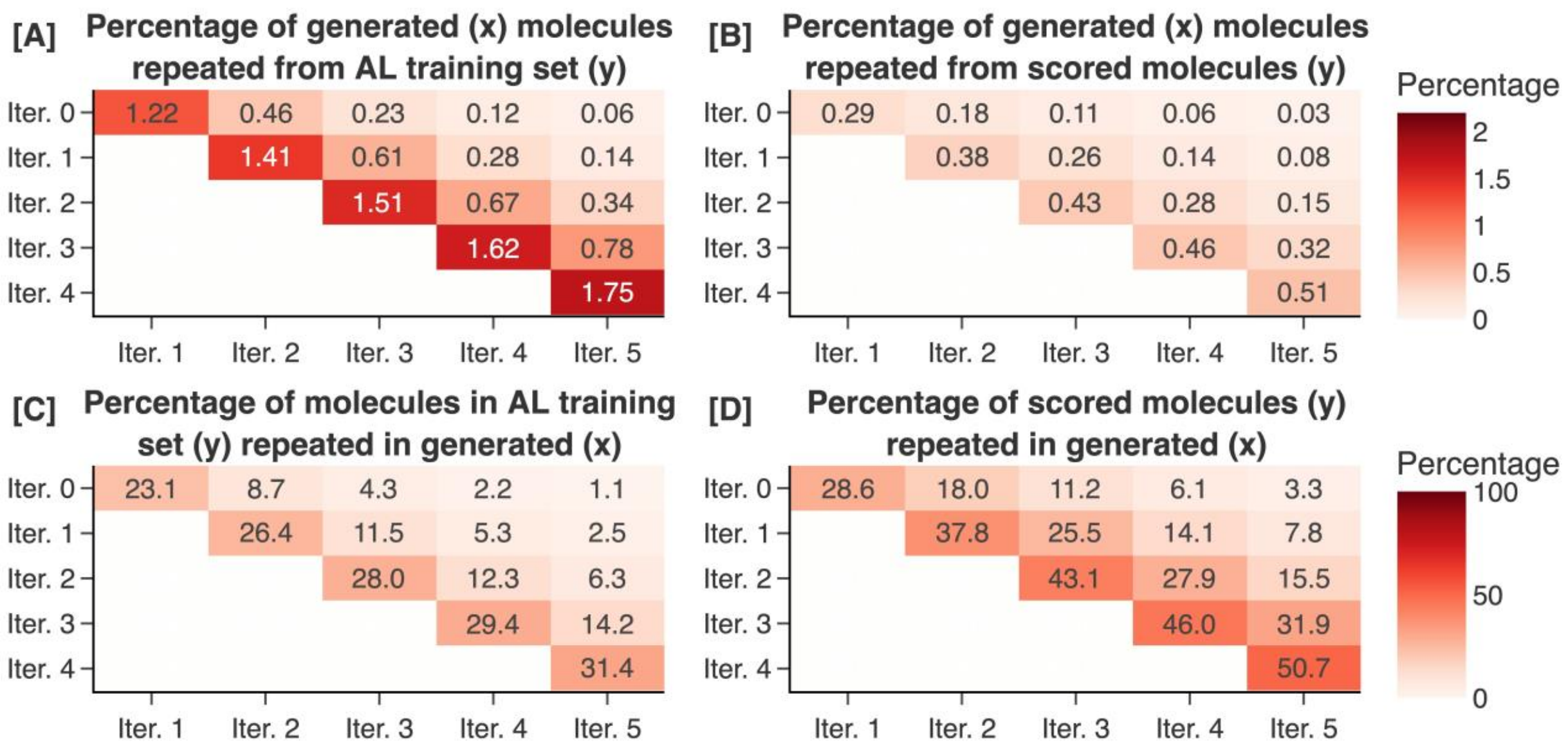

**Figure S22.4.** Memorization of training set by our model over five rounds of active learning with 100 clusters and linear selection. (A) The percentage of molecules in a set of 100,000 generated at iteration *i* that occur in the training set at iteration *i*-1. (B) The percentage of molecules in a set of 100,000 generated at iteration *i* that occur in the set of scored molecules at iteration *i*-1. (C) The number of molecules from the active training set at iteration *i*-1 that occurs in generations at iteration *i* divided by the size of the active learning training set at iteration *i*-1 multiplied by 100. (D) the number of scored molecules at iteration *i*-1 that occur in generations at iteration *i* divided by number of scored molecules at iteration *i*-1 (i.e., 1000) multiplied by 100.

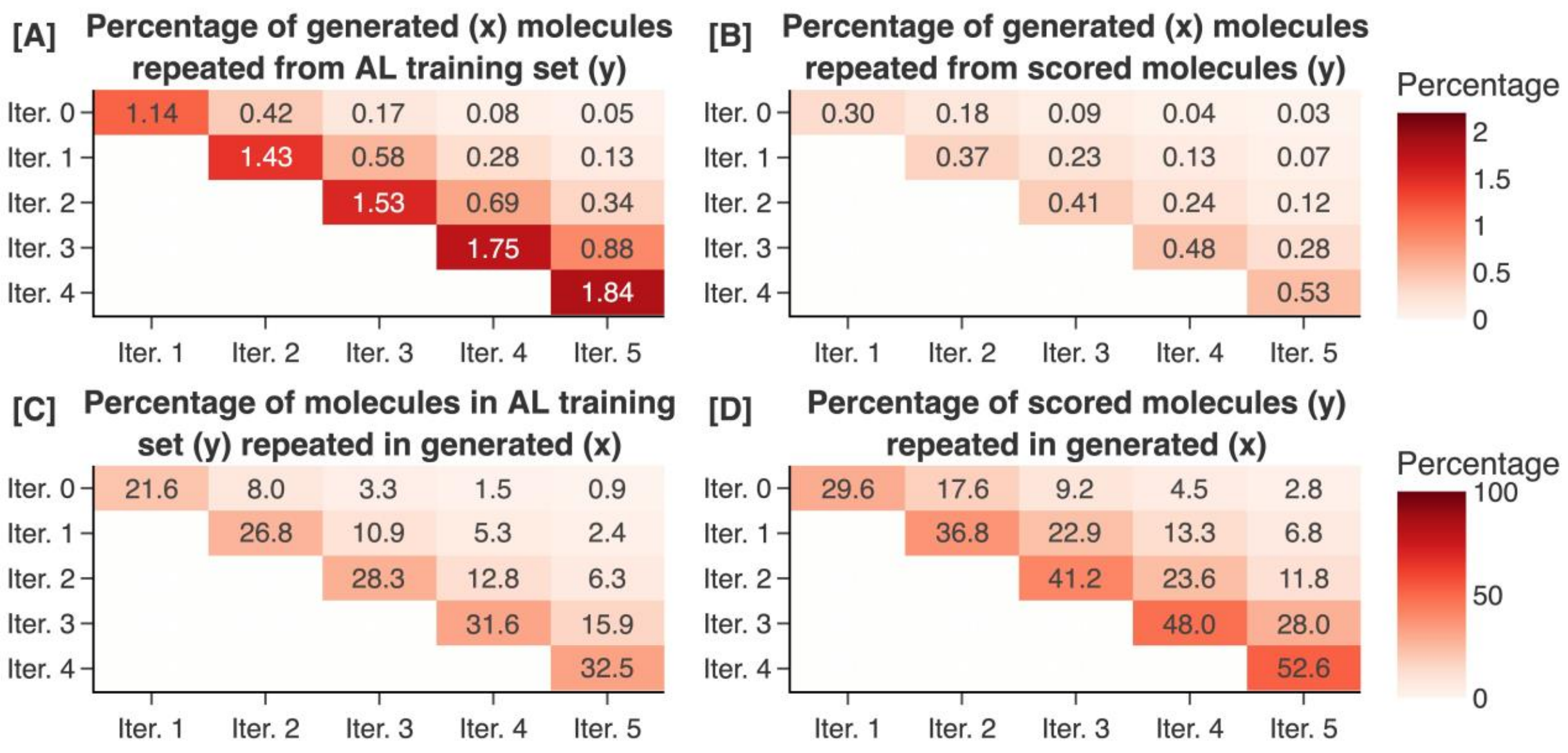

**Figure S22.5.** Memorization of training set by our model over five rounds of active learning with 100 clusters and softdiv selection. (A) The percentage of molecules in a set of 100,000 generated at iteration *i* that occur in the training set at iteration *i*-1. (B) The percentage of molecules in a set of 100,000 generated at iteration *i* that occur in the set of scored molecules at iteration *i*-1. (C) The number of molecules from the active training set at iteration *i*-1 that occurs in generations at iteration *i* divided by the size of the active learning training set at iteration *i*-1 multiplied by 100. (D) the number of scored molecules at iteration *i*-1 that occur in generations at iteration *i* divided by number of scored molecules at iteration *i*-1 (i.e., 1000) multiplied by 100.

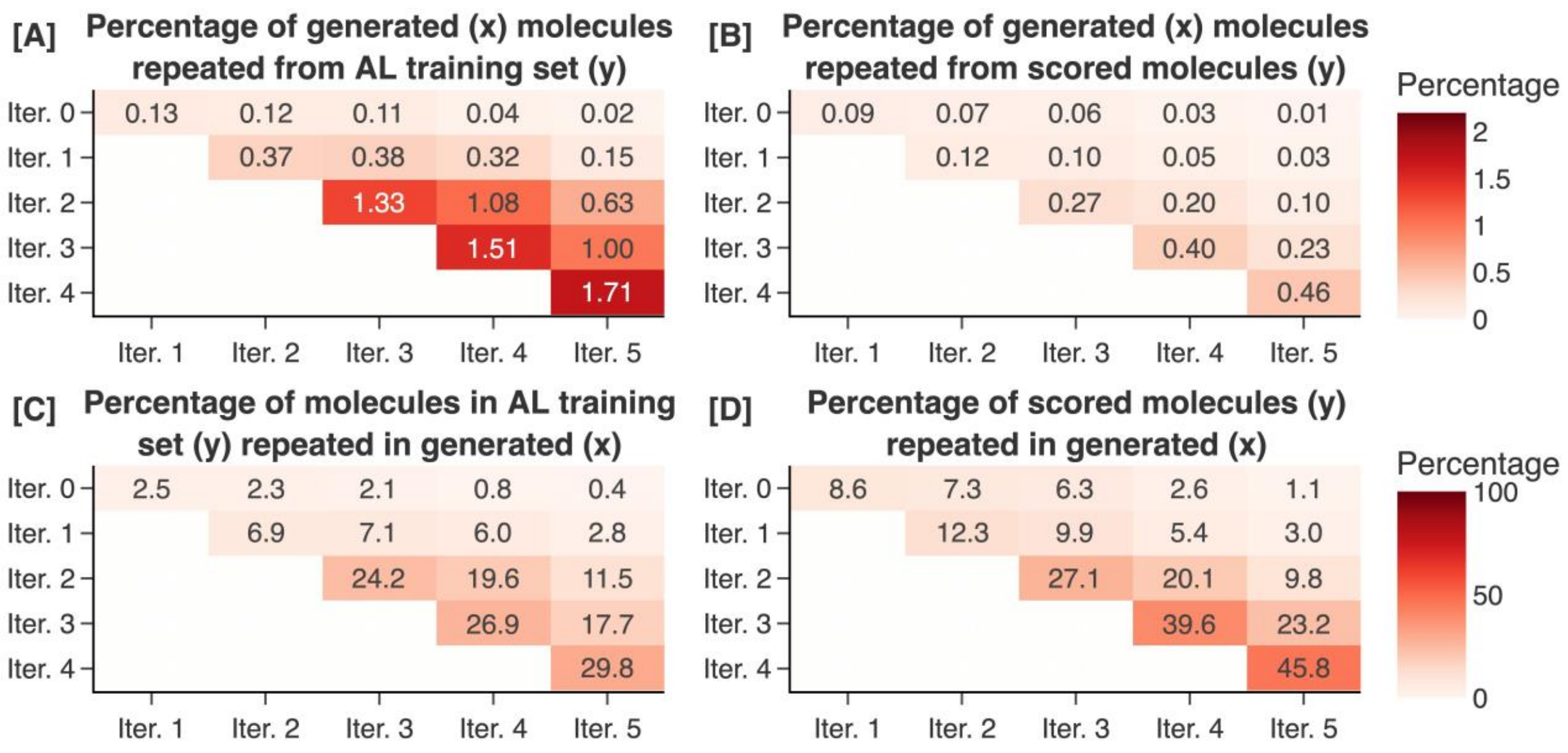

**Figure S22.6.** Memorization of training set by our model over five rounds of active learning with 100 clusters and softsub selection. (A) The percentage of molecules in a set of 100,000 generated at iteration *i* that occur in the training set at iteration *i*-1. (B) The percentage of molecules in a set of 100,000 generated at iteration *i* that occur in the set of scored molecules at iteration *i*-1. (C) The number of molecules from the active training set at iteration *i*-1 that occurs in generations at iteration *i* divided by the size of the active learning training set at iteration *i*-1 multiplied by 100. (D) the number of scored molecules at iteration *i*-1 that occur in generations at iteration *i* divided by number of scored molecules at iteration *i*-1 (i.e., 1000) multiplied by 100.

# For Table of Contents Use Only

ChemSpaceAL: An Efficient Active Learning Methodology Applied to Protein-Specific Molecular Generation

Gregory W. Kyro, Anton Morgunov, Rafael I. Brent, Victor S. Batista*

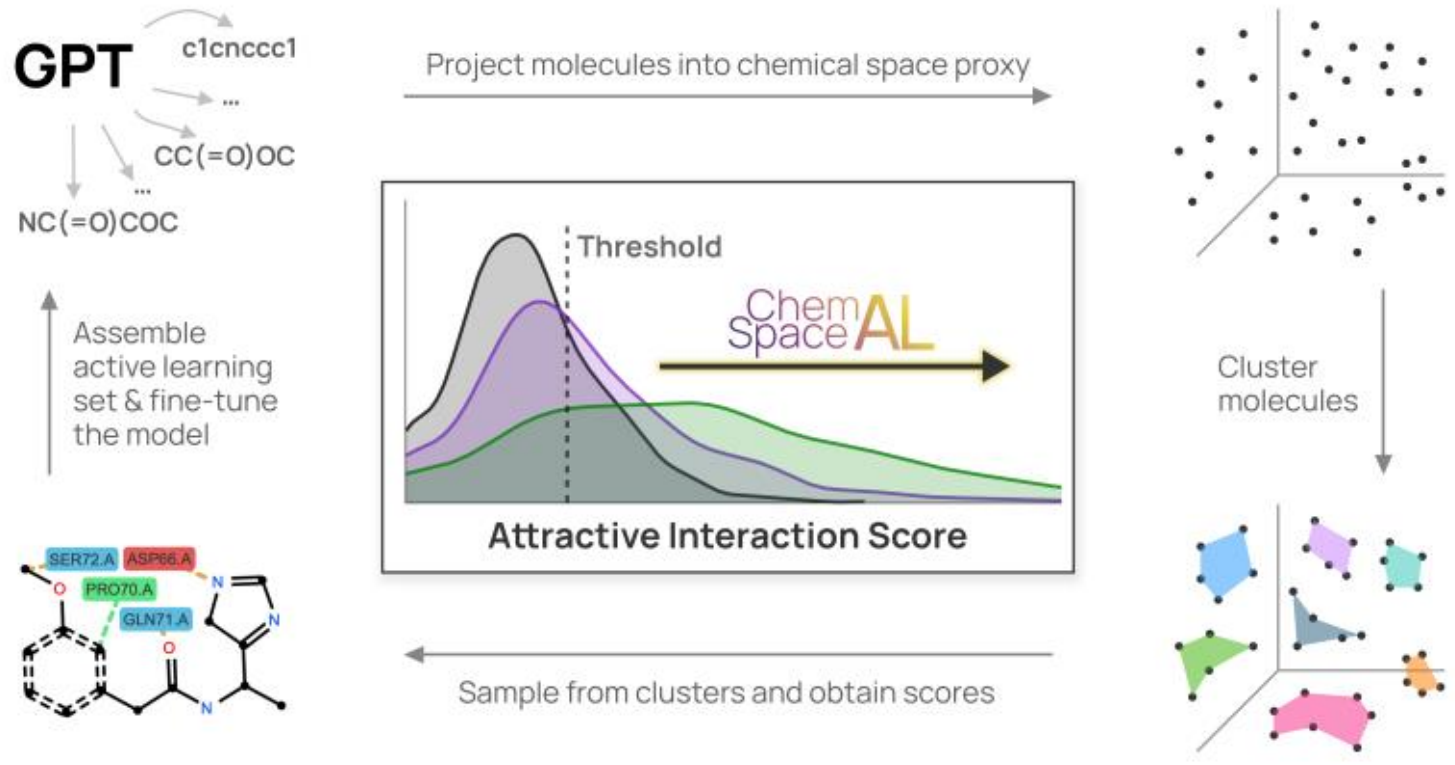